%% file: main.tex
\documentclass[11pt,a4paper]{article}
\usepackage[utf8]{inputenc}
\usepackage[margin=1in]{geometry}

\usepackage{natbib}
 \bibpunct[, ]{(}{)}{,}{a}{}{,}%
 \def\BIBand{and}%

\usepackage[hidelinks]{hyperref} 
\usepackage{subcaption}
\usepackage[table,dvipsnames]{xcolor}
\usepackage{flafter} 
\usepackage{booktabs}
\usepackage{multirow}
\usepackage{arydshln} 
\usepackage{amsmath,amssymb,amstext}

\usepackage{graphicx}

\usepackage[linesnumbered,algoruled,boxed,lined]{algorithm2e}

\usepackage{setspace}
\DeclareMathOperator{\Tr}{tr}

\newcommand{\setCandidateLocations}{\mathcal{V}}
\newcommand{\setSelected}{\mathcal{S}}
\newcommand{\setTargets}{\setCandidateLocations}
\newcommand{\setAll}{\mathcal{N}}
\newcommand{\setDepotStart}{\setAll_S}
\newcommand{\setDepotEnd}{\setAll_E}
\newcommand{\setVehicles}{\mathcal{M}}

\newcommand{\informativeness}{\mathcal{I}}
\newcommand{\coveringNH}{\mathcal{C}}

\newcommand{\yCoord}{p^y}
\newcommand{\xCoord}{p^x}
\newcommand{\solution}{\omega}
\newcommand{\setOpen}{\setAll^{o}}
\newcommand{\setIncluded}{\setAll^{p}}
\newcommand{\setNonIncluded}{\setAll^{u}}
\newcommand{\removalStrat}{\delta^{\textit{removal}}}

\newcommand{\insertStrat}{\delta^{\textit{insert}}}

\newcommand{\insertRandFactor}{p^{\textit{rand}}}
\newcommand{\remainingUtility}{u^r}
\newcommand{\marginalUtility}{u^m}
\newcommand{\currentObjLB}{\underline{\informativeness}}
\newcommand{\currentObjUB}{\overline{\informativeness}}
\newcommand{\objVal}{\informativeness}
\newcommand{\detour}{\textit{detour}}

\newcommand{\temperature}{\Psi}
\newcommand{\heating}{\psi}
\newcommand{\dplabel}{\mathcal{L}}

\begin{document}

\title{Mission planning for emergency rapid mapping with drones}
\author{Katharina Glock, Anne Meyer}
\author{ Katharina Glock\footnote{FZI Research Center for Information Technology, Haid-und-Neu-Str. 10-14, D-76131 Karlsruhe, kglock@fzi.de, ORCID: \url{https://orcid.org/0000-0003-0670-6269}}, Anne Meyer\footnote{TU Dortmund University, Leonhard-Euler-Straße 5, D-44227 Dortmund, anne2.meyer@tu-dortmund.de, ORCID: \url{https://orcid.org/0000-0001-6380-1348}}}
\date{}

\maketitle

\begin{abstract}

We introduce a mission planning concept for routing unmanned aerial vehicles (UAVs) through a set of sampling locations in the immediate aftermath of an incident such as a fire or chemical accident.
Using interpolation methods that account for the spatial interdependencies inherent in the surveyed phenomenon, these samples allow predicting the distribution of hazardous substances across the affected area.
We define the generalized correlated team orienteering problem (GCorTOP) for selecting {informative} samples considering spatial correlations between observed and unobserved locations as well as priorities in the surveyed area. 
To quickly provide high-quality solutions in time-sensitive situations, we propose a two-phase multi-start adaptive large neighborhood search (2MLS).
We show the competitiveness of the solution approach using benchmark instances for the team orienteering problem and investigate the performance of the proposed models and solution approach in an extensive study based on newly introduced benchmark instances for the mission planning problem.

\end{abstract}


\input{content.tex}

\subsection*{Acknowledgement}

This work was supported by the Federal Ministry of Education
and Research (BMBF) under grant number 01IS14012.

\bibliographystyle{informs2014trsc}

\input{main.bbl}

\end{document}

%% file: content.tex
\section{Introduction}

In the immediate aftermath of an incident, the first step towards an effective emergency response is to assess the nature and scale of the situation at hand in order to coordinate emergency response teams.
For major disasters such as landslides, earthquakes, or floods, dedicated systems have been proposed for acquiring this information using satellite-based remote sensing technologies, for example, the Copernicus Rapid Mapping Service \citep{Copernicus}.
Meanwhile, the vast majority of emergencies faced by first responders do not reach the magnitude of natural disasters.
Nonetheless, reconnaissance is the first action to be taken in every operation (see, e.g., the command system of the German federal fire service regulations \citep{Feuerwehr1999}).
In this work, we focus on operations dealing with large fires and chemical accidents that lead to the release of hazardous substances.
The German Federal Office of Civil Protection and Disaster Assistance (BBK) reports that a total of 99 severe incidents and an estimated 1000 incidents involving transports of dangerous goods occurred within three years \citep{BBK2018}.

The objective of the reconnaissance phase after an incident is to obtain information about the spatial distribution of smoke or toxic substances.
The BBK suggests the prioritization of locations depending on the size of the population that may be affected so as to obtain reliable estimates in areas where intervention is most likely to be necessary \citep{BBK2016}.
To this date, fire services largely depend on ground-based surveillance in these situations, which is slow and carries the risk of exposing response teams to hazardous substances.
Rotary wing UAVs equipped with optical remote sensing systems offer an alternative: they can be deployed quickly, can start and land vertically, require less personnel than conventional surveillance methods, and are able to provide information about areas that are inaccessible or are dangerous to access for ground-based reconnaissance units.
Nonetheless, due to the limited flight time of the UAV, it is often impossible to survey an area completely. 
In this case, observations can instead be made at a number of sampling locations, taking advantage of the fact that the distribution of gases and contaminants is usually positively spatially autocorrelated, i.e., similar values can be observed at locations close to each other \citep{Stachniss2009}.
Using geostatistical interpolation approaches, these samples yield an estimation of the extent and severity of the contamination.

The purpose of this paper is to derive models and solution methods that allow the determination of ``informative'' missions for several UAVs within a reasonable time for practical applications.
We refer to this setting as \emph{emergency rapid mapping}.
As it is the case today in the ``manual'' ground-based reconnaissance process, we investigate the emergency rapid mapping problem in a static setting. 
In particular, we assume that priorities and the distribution of gases do not change during the mission. 
These assumptions can be argued, as response teams consider the distribution of the contaminants as well as the distribution of the population as more or less stable during the considerably shorter time required for the UAV missions compared to ground-based reconnaissance.

The contribution of this paper can be summarized as follows: 
\begin{enumerate}
\item We derive new models for jointly considering spatial correlations and priorities for planning informative UAV missions. 
\item We evaluate the impact of the proposed models based on a set of new benchmark instances for the UAV mission planning problem.
\item We propose an exact solution procedure for benchmark purposes, introduce an adaptive large neighborhood search, and demonstrate its performance in experimental studies.
\end{enumerate}

The remainder of this paper is organized as follows: 
In Section \ref{sec:use_case}, we introduce the use case and planning requirements.
Section \ref{sec:spatial_models} provides an overview of statistical models for describing the distribution of airborne substances, which are the foundation of the proposed mission planning approach.
We introduce the central planning problem in Section \ref{sec:problem_statement} and review the literature related to this planning problem in Section \ref{sec:related_literature}.
In Section \ref{sec:models}, we derive models for planning informative UAV missions. 
An exact algorithm and a metaheuristic solution approach are proposed in Section \ref{sec:solution_approach}. 
In Section \ref{sec:cortop_study}, we evaluate the proposed approach on existing and new benchmark instances.
We summarize the main findings of this paper and give an outline of future research in Section \ref{sec:conclusion}.

\section{Mission planning for emergency rapid mapping}\label{sec:use_case}

The solution approach proposed here has been developed within the BigGIS research project\footnote{\url{http://biggis-project.eu/}, accessed 14.05.2019}.
This section provides an overview of the technologies studied within the course of this project and outlines the core concepts of emergency rapid mapping applications.
In Section \ref{sec:uav_system}, we give an overview of UAV systems and sensor technologies based on the experiences in this project.
We summarize sampling guidelines and relevant mission parameters in Sections \ref{sec:sampling_guidelines} and \ref{sec:mission_parameters}, respectively.
Section \ref{sec:use_case_example} gives an illustrative example demonstrating different mission concepts.

\subsection{UAV platform and sensor system}\label{sec:uav_system}

UAV systems can be grouped into two categories: fixed-wing systems and rotary-wing UAVs. 
Fixed-wing UAVs can fly at higher speeds but need to maintain constant forward motion. 
In contrast to fixed-wing UAVs, rotary-wing UAVs have a lower maximum speed and flight duration, but can start and land vertically and are able to stay stationary during the flight. 
This flexibility makes them attractive for emergency surveillance \citep{Boccardo2015}.

For this reason, we focus on rotary-wing UAVs with optical remote sensor systems for surveying an area after an emergency.
These UAVs can carry sensors and equipment with a total weight of up to 3~kg. 
Some systems are able to fly up to around 40 minutes at a maximum horizontal cruise speed of around 50~km/h above ground, depending on the total payload and environmental influences.
In practice, these factors can have a major impact on realistic flight times and cruise speed. 
Optical remote sensors allow the provisioning of information while avoiding direct contact with a substance, even if these substances are invisible to the human eye. 
Commercially available systems include, for example, infrared (IR) camera systems that are able to detect substances such as methane.
The BBK operates, for example, the IR remote reconnaissance device SIGIS~2 for detecting chemical and biological threats \citep{Harig2011}.
In recent years, the emergence of lightweight hyperspectral cameras has allowed their use onboard of UAVs \citep{Aasen2015}.
Similar to earlier approaches for the remote sensing of chemical agents \citep{Flanigan1996,Mayfield2000}, prototypical systems for detecting a wider range of substances using thermal and hyperspectral imaging systems have been developed in the BigGIS project.
Proofs of concepts have successfully been carried out for detecting chlorophyll within smoke clouds.

\subsection{Sampling guidelines}\label{sec:sampling_guidelines}

Because the selection of sampling locations is crucial for estimating the extent and severity of an airborne substance, the BBK has published guidelines and recommendations for sampling in case of chemical, biological and radioactive hazards \citep{BBK2016}.
These guidelines suggest a basic sampling strategy for first responders, consisting of: 
\begin{enumerate}
\item Defining the area of interest, i.e., the area suspected of contamination, 
\item Identifying potential sampling points where a hazard may be present,
\item Prioritizing these candidate sampling locations,
\item Taking samples at a selected set of locations in accordance with these priorities,
\item Evaluating the samples, and
\item Decontaminating personnel and equipment, if necessary.
\end{enumerate}
The priorities referred to in step 3 depend on the likelihood of contamination and on the size of the potentially affected population in an area.
The purpose is to ensure that samples are taken at critical locations, i.e., at regions with a high likelihood of contamination where the civil population may need assistance. 

\subsection{Mission parameters}\label{sec:mission_parameters}

Prior to planning the UAV mission, the operators of the system determine the target area using a rectangular bounding box that encloses the affected region.
Moreover, they specify the number and characteristics of the UAVs, including their maximum flight range and their starting and ending locations.
The target area is discretized into a grid of evenly distributed target locations, each one representing the center of the area that can be covered with one image taken during the flight.
These centers make up the set of candidate sampling locations for the UAVs.
Priorities are assigned to each location based on population data.
If additional data sources such as updated population maps based on mobile phone data are available, they can be used to update UAV missions while they are in progress based on the currently available data and previous observations.

\subsection{Illustrative example}\label{sec:use_case_example}

\begin{figure}[tb]
    \centering
    \begin{subfigure}[b]{0.32\textwidth}
        \includegraphics[width=\textwidth]{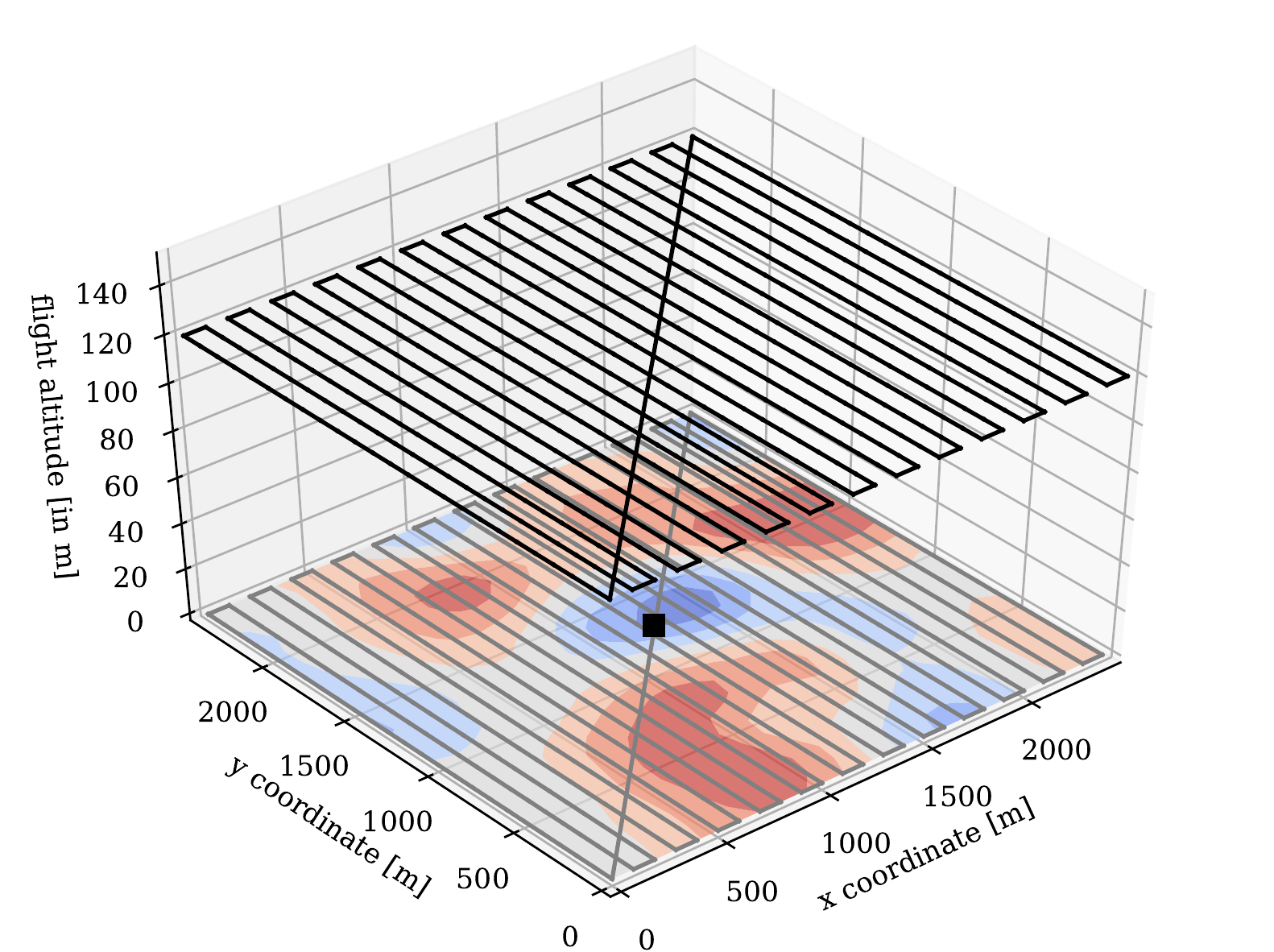}
        \caption{Full coverage \newline($T_r$ of approx. 12500 s)}
        \label{fig:full_coverage}
    \end{subfigure}
    \begin{subfigure}[b]{0.32\textwidth}
        \includegraphics[width=\textwidth]{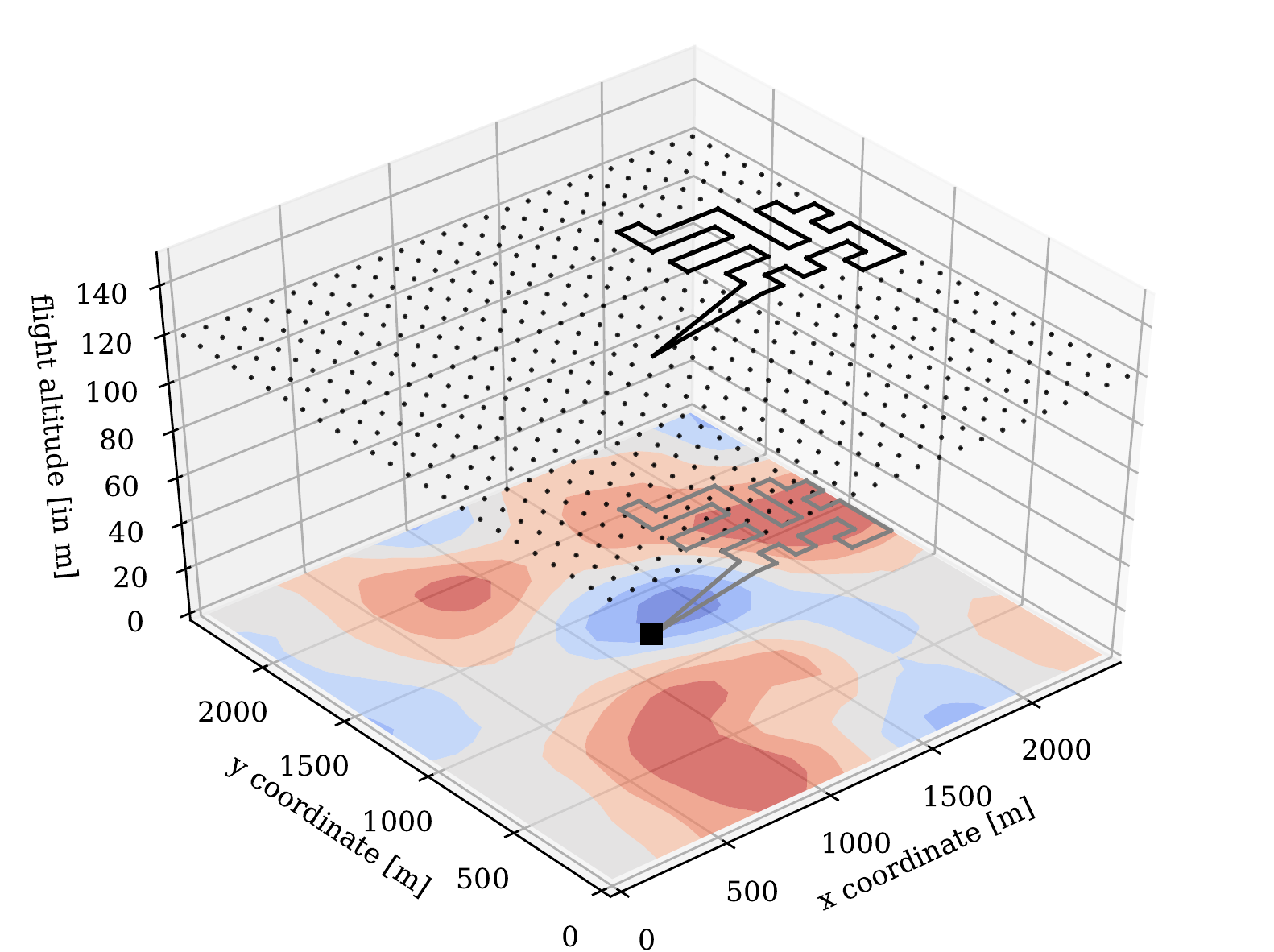}
        \caption{Maximization of priorities \newline($T_r=1200$ s)}
        \label{fig:op_solution}
    \end{subfigure}
    \begin{subfigure}[b]{0.34\textwidth} 
        \includegraphics[width=\textwidth]{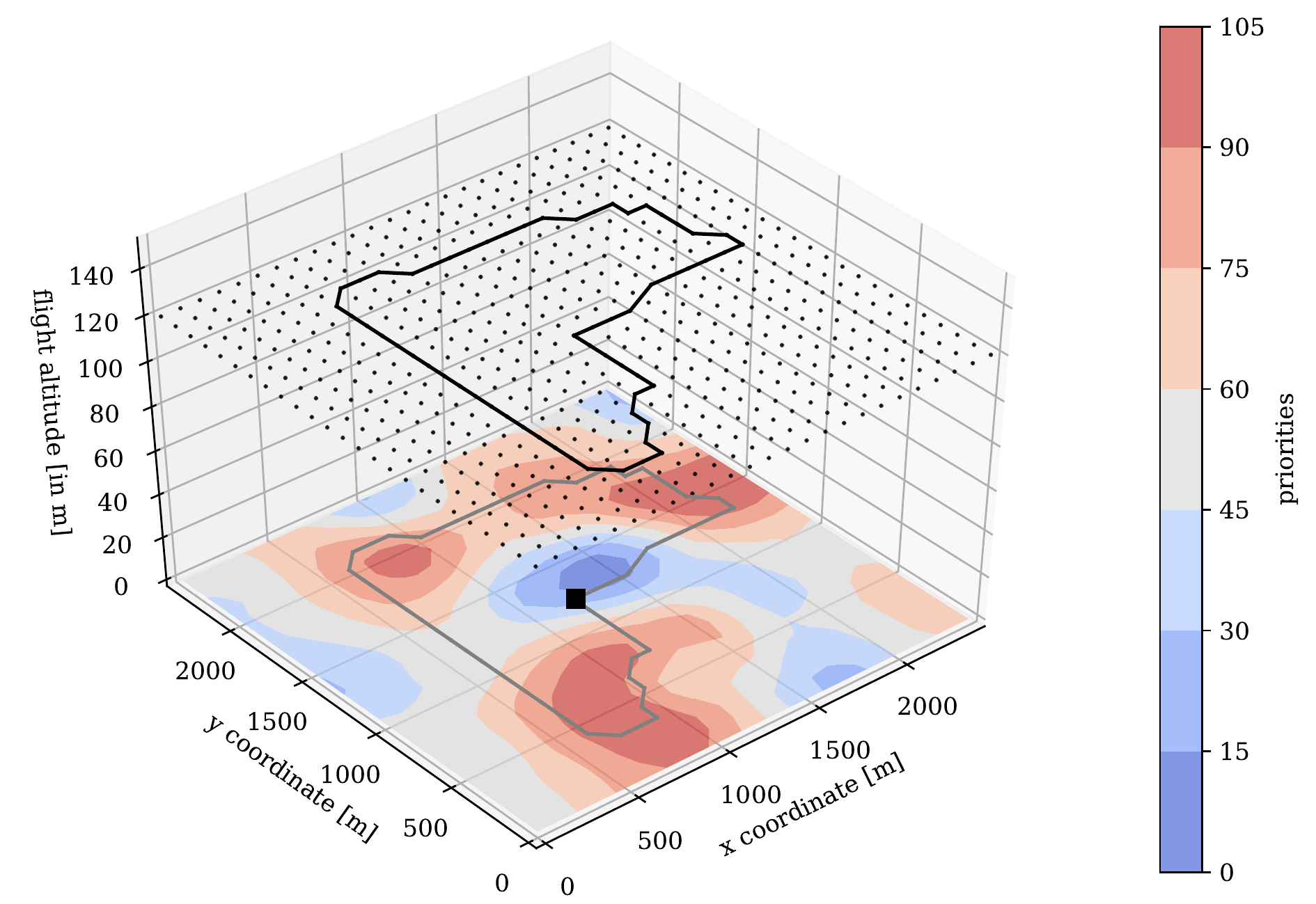}
        \caption{Emergency rapid mapping \newline($T_r=1200$ s)}
        \label{fig:ipp_solution}
    \end{subfigure}
    \caption{Illustration of different mission planning concepts}\label{fig:example_routes}
\end{figure}

In Figure \ref{fig:example_routes}, we illustrate how priorities in the target area and spatial coverage can be taken into account for planning informative UAV missions.
The images depict a scenario with a target area of approximately $2.5 \times 2.5~\text{km}^{2}$.
The colors indicate the priorities, with red representing highly relevant locations, and blue representing lower-ranked ones.
Candidate waypoints for the UAV are indicated at the flight altitude of 120~m, while the route is depicted as a solid line.

Figures \ref{fig:full_coverage} to \ref{fig:ipp_solution} show three concepts for surveying this area.
In all cases, the UAV departs from and returns to a location indicated in the middle of the target region.
Figure \ref{fig:full_coverage} shows a mission pattern that ensures full ground coverage, a concept that is often used in environmental sensing \citep[see e.g.][]{Stachniss2009}.
While providing complete information about the target area, this UAV mission is not feasible in practice due to the long flight time.
In contrast, Figure \ref{fig:op_solution} represents the mission resulting from maximizing the sum of priorities of the surveyed locations given a realistic flight time limitation.
In this solution, the UAV quickly travels to the area with the highest priorities and stays there.
Finally, in Figure \ref{fig:ipp_solution}, we depict a mission plan that combines priorities and spatial coverage.
This plan offers two advantages: Firstly, it includes all highly prioritized regions. Secondly, as we will see in the next section, distributing the sample locations across the area allows a more accurate prediction of the distribution of hazardous substances.

\section{Models for spatial processes}\label{sec:spatial_models}

The comparison of the two UAV missions depicted in Figures \ref{fig:op_solution} and \ref{fig:ipp_solution} illustrates the impact that considering the spatial correlation in the surveyed field can have on the UAV mission.
Existing approaches for sampling design use formalized models for spatial phenomena account for these interdependencies \citep{Krause2008}.
Likewise, these concepts provide the theoretical foundation for our own models.
In this section, we provide an overview of the central ideas and approaches for modeling spatial processes in a stochastic fashion.

\subsection{Expressing spatial structure}

Environmental processes such as the distribution of airborne contaminants over a given spatial area follow physical laws, but are too complex to be easily described or even understood.
A possibility to deal with the complexity is to consider these processes as random and to employ statistical methods to describe and analyze them \citep{Cressie2011}. 
Representing the affected area via a set of Cartesian coordinates $\mathcal{V}\subset \mathbb{R}^2$, a physical phenomenon can be described as a stochastic process $\lbrace Z(s)\rbrace_{s \in \mathcal{V}}$ over the locations in $\mathcal{V}$, i.e., as a collection of random variables defined over a two-dimensional field (ibid.).
The spatial structure of a given phenomenon is expressed by its spatial covariance
$\text{Cov}(Z(s), Z(s'))$,
which describes the relationship between pairs of variables $Z(s)$ and $Z(s')$ at different locations $s, s' \in \setTargets$.
The covariance of a stochastic process is usually modeled using a covariance function $k(s, s') := \text{Cov}(Z(s), Z(s'))$, often called a kernel \citep{Rasmussen2006}.
It depends on the distance between these locations and generally decreases with growing distance \citep{Krause2008}.
The pairwise covariances are combined in a covariance matrix $\Sigma_{\setTargets, \setTargets}$ with each element describing the covariance $\text{Cov}(Z(s), Z(s'))$ for $s, s' \in \mathcal{V}$.
Therefore, this matrix comprises the fundamental information and assumptions about the relationship between locations, and therefore about the spatial process itself. 

\subsection{Spatial interpolation}\label{sec:spatial_interpolation}

The main purpose of modeling spatial processes is to allow inferences about unsurveyed locations based on a finite number of known observations.
A popular way to do so is to model the spatial phenomenon as a Gaussian process (GP), denoted as $Z_{\mathcal{V}}:=\lbrace Z(s)\rbrace_{s \in \mathcal{V}}$, i.e., as ``a collection of random variables, any finite number of which have a multivariate normal distribution'' \citep{Rasmussen2006}.
This means that for any finite subset $\setTargets' \subset \setTargets$, the random vector $Z_{\mathcal{V}'}$ is normally distributed.
A GP is completely defined by its covariance $k(s, s')$ and mean $\mu(s) = \text{E}[Z(s)]$, which gives the expected value for all $s \in \setTargets$.

The GP without taking sample observations into account is referred to as a Gaussian \emph{prior}.
This prior encapsulates all assumptions or information that one could possibly have about the phenomenon without having performed observations, in particular, assumptions about the kernel function and smoothness of the distribution. 
In short, we denote the prior as
\begin{align}
Z_{\setTargets} \backsim \mathcal{GP}(\mu_{\setTargets}, \Sigma_{\setTargets, \setTargets})\label{f:gp_prior}
\end{align} 
with the prior mean $\mu_\mathcal{V}$ representing a column vector of expected means with one entry for each $s \in \mathcal{V}$.

If samples are available for a set of locations $\setSelected\subset \setTargets$, the prior belief is updated taking these measurements into account. 
The result is a Gaussian \textit{posterior}, represented by a posterior mean $\mu_{\setTargets|\setSelected}$ and a covariance matrix $\Sigma_{\setTargets,\setTargets |\setSelected }$ conditioned on a set $\setSelected$.
We write this as
\begin{align}
Z_{\setTargets|\setSelected} \backsim \mathcal{GP}(\mu_{\setTargets|\setSelected}, \Sigma_{\setTargets, \setTargets|\setSelected}).\label{f:gp_posterior}
\end{align} 
This posterior incorporates the knowledge obtained using the sampled locations in combination with the fundamental characteristics of the process that are given by the prior.
To derive the posterior, we define $\Sigma_{\setTargets,\setSelected}$ as the covariance matrix between all $s, s'$ with $s \in \setTargets, s' \in \setSelected$, and $\Sigma_{\setSelected,\setSelected}$ as the covariance matrix between all $s,  s'\in \setSelected$. 
Furthermore, $\mu_\setSelected$ designates a column vector of prior mean values $\mu(s)$ for $s \in \setSelected$.
The observed values at locations $\setSelected$ are represented as a column vector $u_\setSelected$.
We can then determine the mean vector and covariance matrix of the GP posterior as follows:
\begin{align}
\mu_{\setTargets|\setSelected} & = \mu_\setTargets + \Sigma_{\setTargets,\setSelected}\Sigma_{\setSelected,\setSelected}^{-1} (u_\setSelected - \mu_\setSelected) \label{f:GP_mean}\\
\Sigma_{\setTargets,\setTargets|\setSelected} & = \Sigma_{\setTargets,\setTargets} - \Sigma_{\setTargets,\setSelected}\Sigma_{\setSelected,\setSelected}^{-1}\Sigma_{\setSelected,\setTargets}.\label{f:GP_cov}
\end{align}

The posterior mean determined in Equation (\ref{f:GP_mean}) gives the predicted values $\mu_{\setTargets|\setSelected}$ for all $s \in \setTargets$.
The posterior covariance in Equation (\ref{f:GP_cov}) indicates the uncertainty associated with these predictions. 

\subsection{Designing sampling strategies}\label{sec:sampling_strategies}

Gaussian processes can be used to design sampling strategies, i.e., strategies for selecting sampling locations that yield as much information as possible about a spatial process \citep{Curran1998}.
This is made possible by the fact that a GP provides knowledge about the variability that remains in the process, represented in form of the posterior covariance $\Sigma_{\setTargets,\setTargets|\setSelected}$ (Equation (\ref{f:GP_cov})).
This covariance is independent of the observed \emph{values} $u_\setSelected$ and only depends on the sampled \emph{locations} $\mathcal{S}$, which makes it possible to estimate the quality of a set of sampling locations $\setSelected$ prior to actually making the observations. 

One possible measure for the variability in the process is the trace $\Tr(\Sigma)$ of a covariance matrix, which is defined as the sum of the diagonal elements $k(s,s), s \in \mathcal{V}$.
Another measure of uncertainty is the so-called ``entropy'' $\mathcal{H}(Z_{\setTargets})$ of a Gaussian process $Z_{\setTargets}$ in $|\setTargets|=n$ dimensions \citep{Rasmussen2006}:
\begin{align*}
\mathcal{H}(Z_{\setTargets}) = 0.5 \cdot \text{log}((2\pi e)^n |\Sigma_{\setTargets,\setTargets}|),
\end{align*}
where $|\Sigma_{\setTargets,\setTargets}|$ denotes the determinant of the covariance matrix $\Sigma_{\setTargets,\setTargets}$.
The lower these measures are, the better is the corresponding prediction.

Based on the trace and the entropy, additional variability measures for estimating the information gain yielded by a set $\setSelected$ have been proposed in the literature on Gaussian processes.
We discuss two measures often used in literature:
The first criterion for measuring the quality of a set $\mathcal{S}$ is the \emph{average reduction in variance (ARV)} that is achieved by the selected sampling locations \citep{Das2008,Krause2008b}.
The measure is formally defined as
\begin{align}
ARV(\setSelected) = \frac{1}{|\setTargets|}(\Tr(\Sigma_{\setTargets,\setTargets}) - \Tr(\Sigma_{\setTargets,\setTargets|\setSelected})),\label{f:arv}
\end{align}
with $\Tr(\Sigma_{\setTargets,\setTargets})$ and $\Tr(\Sigma_{\setTargets,\setTargets|\setSelected})$ measuring the variability of the GP prior and posterior, respectively.
Another criterion is the \emph{mutual information (MI)} between the selected sensing locations and the interpolated locations $\setTargets \setminus \setSelected$ \citep{Caselton1984}.
With $Z_{\setTargets\setminus\setSelected}$ as the Gaussian prior over all interpolated locations and $Z_{\setTargets\setminus\setSelected|\setSelected}$ as the corresponding Gaussian posterior, MI is defined as follows:
\begin{align}
MI(\setSelected) = \mathcal{H}(Z_{\setTargets\setminus\setSelected}) - \mathcal{H}(Z_{\setTargets\setminus\setSelected|\setSelected}).\label{f:mi}
\end{align}

Using these criteria, sensing locations $\setSelected$ can be determined such that ARV or MI is maximal, which means that the variability of the posterior is minimal.
Note that the basic complexity of all of these measures is $\mathcal{O}(n^3)$, because computing the GP posterior requires inverting an $n \times n$ matrix with $n = |\setSelected|$ \citep{Rasmussen2006}.

\section{The mission planning problem for emergency surveillance}\label{sec:problem_statement}

Based on foundations laid in the previous two sections, we define the mission planning problem for emergency surveillance (MPPES): 
The MPPES consists of determining tours for UAVs through a selected subset of candidate sensing locations such that first responders can be provided with a reliable overview of the distribution of hazardous substances.
This planning problem consists of three simultaneous decisions:
\begin{enumerate}
\item The selection of appropriate sensing locations,
\item The assignment of the selected locations to UAVs, and
\item The determination of routes through the assigned target locations for each UAV.
\end{enumerate}
The routes have to comply with the flying time constraints of the UAVs.
They are furthermore restricted by the specified take-off and landing locations, which may lay outside the specified target area.

The second and third decisions arise in nearly all types of vehicle routing problems \citep{Irnich2014}.
The first decision is crucial for planning informative missions and mainly characterizes the problem at hand.
To increase the informativeness of the missions, this selection has to take into account that the distribution of hazardous substances within a region exhibits positive spatial correlations.
Hence, samples in close proximity yield less information about the overall distribution of a contaminant than samples taken further apart at locations not correlated with one another.
This interrelationship has to be adequately considered when planning UAV missions.

The priority of sample locations depends on the risk of contamination and the size of the population that may be affected.
This ensures that the focus of the UAV mission lies in areas where an intervention of the response personnel is most likely to be necessary.
Jointly considering spatial interdependencies and the target locations' priorities in MPPES is therefore essential to achieve both accurate information in highly prioritized areas and a reliable overview of the distribution of contaminants throughout the entire region.
\section{Related literature}\label{sec:related_literature}

The MPPES introduced in the previous section shares similarities with problems studied in different disciplines: 
the variants of the covering tour problem and the team orienteering problem discussed in the operations research literature on vehicle routing, and the informative path planning addressed in the field of robotics.

\subsection{Covering tour problem}\label{sec:ctp_literature}

Covering tour problems deal with the determination of cost-minimal (e.g., assignment, travel, and/or service cost) routes such that every node that is not included in a vehicle tour is within a given maximum distance to a node directly served by a vehicle. 
The covering salesman problem (CSP) was first introduced and formulated by \cite{Current1989} as a variant of the traveling salesman problem (TSP).
\cite{Gendreau1997} and \cite{Golden2012} proposed generalizations of this problem variant, referred to as covering tour problems (CTP), which consider modified versions of covering concepts (e.g., mandatory nodes, nodes that must be covered but cannot be visited directly and nodes that must be served or covered multiple times).
All authors provide heuristics for this single-vehicle version that combine approaches for solving a set covering problem with VRP heuristics.
\cite{Gendreau1997} additionally proposed an exact branch-and-cut approach. 

\cite{Hachicha2000}, \cite{Naji-Azimi2012}, \cite{Ha2013}, and \cite{Allahyari2015} consider different variants of the multi-vehicle case denoted the $m$-CTP. 
\cite{Ha2013} propose a heuristic approach and an exact branch-and-cut algorithm, while the other authors describe heuristic approaches, notably local search variants and metaheuristics. 

\subsection{Combining coverage and profit maximization}

Coverage aspects in the context of profit maximization have been addressed more frequently in recent years.
\cite{Yu2014} introduce the correlated team orienteering problem (CorTOP) with the explicit objective of integrating information about spatial correlations in the model.
Visited nodes partially cover unvisited nodes nearby.
This coverage is cumulative, i.e., additional stops within covering distance provide an additional benefit. 
When all covering nodes are included in a vehicle tour, the full reward of an unvisited target node is provided. 
The authors propose a mixed integer quadratic programming (MIQP) formulation for solving the problem using a commercial solver. 
The largest instances solved involve 49 candidate locations in the single-vehicle case and 36 for the multi-vehicle case.
Still, the combination of coverage aspects with profit maximization seems promising for the planning problem addressed in this work.
More recently, \cite{Ozbaygin2016} proposed a branch-and-cut approach for a single-vehicle problem called the maximal covering salesman problem (MaxCSP) with the objective of maximizing covered demand.
The authors consider unvisited nodes as covered as long as they are within a given distance to a visited node. 
However, only a given percentage of a node's demand can be covered that way.
This percentage is independent of the number of nodes that provide coverage.

A related problem is the set orienteering problem (SOP) introduced by \cite{Archetti2018}, where customers are grouped into clusters and visiting one customer within a cluster allows collecting the full profit associated with this cluster.
Clusters can, but do not have to, represent spatial relations.
Additional visits within one cluster do not provide additional benefits.
Both heuristic and exact solution approaches have been proposed \citep{Pvenivcka2019}.
However, only the single-vehicle case has been considered to this date.

\subsection{Orienteering and team orienteering}

The orienteering problem (OP) and the team orienteering problem (TOP) are variants of the TSP and VRP, respectively, in which not all nodes can be visited due to scarce resources.
The optimization objective is profit maximization rather than the minimization of resource consumption.
This problem has been widely studied in VRP literature.
A detailed overview is provided in recent surveys by \cite{Vansteenwegen2011} and \cite{Gunawan2016}.
In this review, we focus on the approaches that are state-of-the-art in terms of solution quality or computational performance on benchmark instances.

\cite{Souffriau2010} propose a fast path relinking approach based on a greedy randomized adaptive search procedure.
\cite{Dang2013pso} introduced a particle swarm optimization-inspired algorithm (PSOiA) combining destruction and recreation operators, TSP search moves, and crossover operators for merging promising partial tours.
\cite{Vidal2015} propose a genetic search including a new neighborhood concept: Based on solutions where resource constraints are relaxed, feasible TOP solutions are determined by solving a resource-constrained shortest path problem for each vehicle.
For generating faster solutions, \cite{Vidal2015} also provided a multi-start local search relying on a similar neighborhood concept.
\cite{Ke2016} propose a Pareto mimic algorithm that maintains a population of solutions based on Pareto dominance.
The so-called mimic operator is used to build new solutions based on an incumbent one in a similar fashion as path relinking \citep{Souffriau2010}.

\subsection{Informative path planning}\label{sec:lit_ipp}

Informative path planning (IPP) is concerned with planning vehicle trajectories for monitoring a phenomenon that varies in time and space while respecting the vehicles' maximum range.
IPP approaches seek to determine sensing locations such that the uncertainty remaining in the process is minimal.
To this end, they use the variability measures discussed in Section \ref{sec:sampling_strategies} for determining the information gain achieved by the vehicles with respect to the observed phenomenon.

Most of the solution approaches discussed in this line of research have focused on optimal algorithms or approximation algorithms providing performance guarantees in view of high sensing costs. 
\cite{Singh2007,Singh2009} propose a recursive greedy algorithm for solving a single-vehicle problem variant.
They furthermore address the multi-vehicle case by sequentially applying this algorithm to a series of single-vehicle problems.
\cite{Binney2010} use a version of this algorithm to solve an IPP variant with time windows for accessing certain areas.
A branch-and-bound algorithm for the single-vehicle IPP has been introduced by \cite{Binney2012}.
Despite applying acceleration techniques, problems remain computationally intractable for vehicle routes comprising more than around 15 locations.
\cite{Hollinger2014} discuss a rapidly-exploring information gathering algorithm, which iteratively assigns random sampling locations to vehicle routes and expands vehicle paths towards these nodes.
However, all of these approaches scale poorly in case of more than a few dozen target nodes, despite the integration of acceleration techniques.

\subsection{Comparison of models and contribution of this paper}

\begin{table}
\centering      
\scriptsize    
\renewcommand{\arraystretch}{1.2}
\begin{tabular}{ p{0.15\textwidth} p{0.14\textwidth}  p{0.07\textwidth}  p{0.07\textwidth}  p{0.15\textwidth}  p{0.12\textwidth}  p{0.14\textwidth} }
\hline
{Problem variant}     &    {First reference}     &    Vehicles &  Priorities & Spatial interdep. & Objective & {Solution approach}\\
\hline
Orienteering problem (OP)  & \cite{Tsiligirides1984}    & 1 &  Yes & Not considered &     Max. sum of priorities    &    Heuristics,exact approaches    \\ 
\hdashline
Covering salesman problem (CSP)    &    \cite{Current1989}    & 1 &    No &  Complete coverage of nearby nodes & Min. distance & Heuristics, exact approaches    \\
\hdashline
Team orienteering problem (TOP)    & \cite{Chao1996top} &    $m$    &    Yes & Not considered &     Max. sum of priorities    &    Heuristics,exact approaches    \\ 
\hdashline
Covering tour problem (CTP)    &    \cite{Gendreau1997}    &    1    &    No    &    Complete coverage of nearby nodes &  Min. distance    & Heuristics, branch-and-cut \\
\hdashline
$m$-covering tour problem($m$-CTP)    &    \cite{Hachicha2000}    &    $m$    &    No    &    Complete coverage of nearby nodes &     Min. distance    & Heuristics, branch-and-cut    \\
\hdashline
Informative path planning (IPP)    &    \cite{Singh2007}    &    1,$m^{*}$ & No    &    Implicitly defined by GP models &     Min. prediction variability & Exact and approximative approaches \\
\hdashline
Correlated team orienteering problem (CorTOP)    &    \cite{Yu2014}    &    $m$    & Yes &    Partial and cumulative coverage of nearby nodes & Max. sum of direct and covered priorities    &Branch-and-bound   \\
\hdashline
Maximal covering salesman problem (MaxCSP)    &    \cite{Ozbaygin2016}    &    1    &    Yes  &    Partial coverage of nearby nodes & Max. sum of direct and covered priorities    &Branch-and-cut    \\
\hdashline
Set orienteering problem (SOP)    &    \cite{Archetti2018}    & 1 &    Yes &  Complete coverage of nodes in same cluster & Max. sum of direct and covered priorities & Heuristics, exact approaches    \\
\hline
This contribution &        &    $m$    &    Yes    &    Partial, distance-dependent, cumulative coverage of nearby nodes & Max. sum of direct and covered priorities    & Dynamic programming and heuristic approach \\
\hline\noalign{\smallskip} \multicolumn{7}{l}{\textsuperscript{(*)} Multi-vehicle cases are typically solved by sequentially planning single-vehicle tours}\\
\end{tabular}
\caption{Summary of problem variants with profit maximization or coverage constraints}
\label{table:related_work}
\end{table}

The problem variants related to the MPPES are summarized in Table \ref{table:related_work}, focusing in particular on the respective planning objective and the consideration of spatial interdependencies between nodes in the various models.
Comparing these characteristics with the requirements for planning UAV missions discussed in Section \ref{sec:problem_statement}, we can identify three major aspects that need to be addressed for solving this planning problem successfully:

\paragraph{Joint consideration of profits and spatial interdependencies}
A significant body of work either focus on profit maximization or the minimization of resource utilization subject to aspects of spatial coverage or spatial interdependencies.
To the best of our knowledge, only the models of \cite{Yu2014} and \cite{Ozbaygin2016} combine profit maximization with coarse models for spatial correlations.

\paragraph{Approximate models for spatial interdependencies}
Most models proposed in the literature considering spatial interdependencies in route planning use either very coarse approximations (CSP, CTP, CorTOP, MaxCSP) or apply Gaussian process models (IPP). 
The latter provide sophisticated models for spatial interdependencies, but are computationally expensive ($\mathcal{O}(n^3)$ in the number of sampled locations): 
The impact of local changes to a solution cannot be computed efficiently -- a prerequisite for many successful VRP approaches such as local search based heuristics.
Models that are sufficiently accurate but can be evaluated efficiently are missing, even if the CorTOP is a first promising step.

\paragraph{Efficient solution approaches for the mission planning problem}
No efficient solution approach exists that is applicable to the mission planning problem.
Efficient heuristics designed for TOP do not account for spatial interdependencies between nodes, while those proposed for CTP exploit the fact that all nodes need to be covered, which is not applicable in our use case. IPP approaches are limited in the size of the problem instances they can solve.
For MaxCSP and CorTOP, only exact solution approaches have been proposed, which are not scalable.

We address this gap by proposing generalized models combining priorities and approximation of spatial interdependencies that are more accurate than the simple coverage models and require less computational effort than GP models (Section \ref{sec:models}).
These models can be seen as a generalization of the models with partial coverage introduced by \cite{Yu2014} and \cite{Ozbaygin2016}. 
To find good solutions quickly, we propose a two-phase multi-start adaptive large neighborhood search that incorporates knowledge about spatial interdependencies for the construction of vehicle routes (Section \ref{sec:solution_approach}).
For benchmarking purposes, we introduce an exact solution approach.
In an extensive computational study, we evaluate the impact of different modeling variants on the solutions and the performance of the proposed approaches (Section \ref{sec:cortop_study}).

\section{Models for planning informative UAV missions}\label{sec:models}

This section formally defines the MPPES and discusses different modeling variants that differ in how they quantify ``informativeness''.
Because this measure determines the selection of locations to be surveyed, it is crucial for the accuracy of any spatial interpolation approach based on these samples.
Based on a basic problem formulation, we introduce stochastic modeling variants.
We derive approximations for spatial interdependencies and show how to integrate priorities.

\subsection{Basic problem formulation} \label{sec:ipp_base_model}

Consistent with the notation introduced in Section \ref{sec:spatial_models}, we denote the set of locations within the two-dimensional target area as $\setTargets$ and the set of sensing locations as $\setSelected \subseteq \setTargets$.
When planning informative tours, the selection of the sensing locations $\setSelected$ constitutes the main decision.
Each UAV $m$ in the set of available UAVs $\setVehicles$ is associated with a starting location $s_m$ and ending location $e_m$.
In the following, $\setDepotStart$ and 
$\setDepotEnd$ designate the sets comprising UAV starting and ending locations, respectively.
We refer to the set of all locations as $\setAll$, with $\setAll = \setTargets \cup \setDepotStart \cup \setDepotEnd$.
Target locations $i \in \setTargets$ are associated with priorities $u_i \geq 0$ that specify their relevance to the response units.
As discussed in Section \ref{sec:mission_parameters}, these priorities can, for example, indicate the size of the affected population.
A survey at a location $i$ requires a fixed sensing time $\tau_i$ in order to take and process images at standstill.
Each UAV mission is limited by a maximum duration of $T_m^{max}$.
The distance between two locations $i$ and $j$ is denoted as $d_{ij}$.
Traveling from a location $i$ to a location $j$ with $i,j \in \setAll$ requires a nonnegative travel time $\tau_{ij}$ that includes the time necessary for acceleration and deceleration. 
To simplify the formulation, this also includes the sensing time $\tau_j$ at the destination location.

We define the set of all feasible routes $\Omega = \bigcup_{m \in \setVehicles} \Omega_m$.
Each route $r \in \Omega_m$ consists of a sequence of locations $r = (s_m, i_0, i_1, \ldots, i_n, e_m)$ such that the total travel time of this route $T_r = \tau_{s_m i_0} + \sum_{c = 0}^{n - 1}\tau_{i_c i_{c+1}} + \tau_{i_n e_m} \leq T_m^{max}$.
The set of included locations $\setSelected_r = \{i_0, i_1, \ldots, i_n\} \subset \setTargets$ corresponds to the sensing locations selected for the route $r$.
We denote a feasible solution to this problem as $\solution$, where $\solution = (r_1, \ldots, r_m) \in \Omega_1 \times \ldots \times \Omega_m$ represents one element in the Cartesian product of routes, i.e., one combination of routes with one feasible route per vehicle.
Finally, we consider a general measure of informativeness $\informativeness$ to determine the quality of a solution.
We can now model the mission planning problem using binary decision variables $y_r$ with
\begin{align*}
y_r = 
\begin{cases}
1,~~\text{if route } r \in \Omega \text{ is selected},\\
0,~~\text{otherwise.}
\end{cases}
\end{align*}
Then, the set of sampling locations is $\setSelected := \bigcup_{r: y_r = 1} \setSelected_r$.
This notation can consistently be used both for the stochastic informativeness measures of Section \ref{sec:ipp_probabilistic_models} and for the models introduced in this work (Section \ref{sec:ipp_discretized_models}).
The basic problem of planning informative UAV missions can then be stated as follows:
\begin{align}
\text{(MPPES)~~~} 
& \max ~~ \informativeness(\setSelected) \label{f:ipp_objective}\\
\text{~~~ s.t.~~~} 
& \sum_{r \in \Omega_m}y_{r} = 1 & m \in \setVehicles \label{f:ipp_01}\\
& y_r \in \{0, 1\} & r \in \Omega \label{f:ipp_02}
\end{align}
Objective \eqref{f:ipp_objective} maximizes total informativeness, while constraints \eqref{f:ipp_01} ensure that exactly one route is selected per vehicle. 
Constraints \eqref{f:ipp_02} set the variable domains.

\subsection{Stochastic informativeness measures}\label{sec:ipp_probabilistic_models}

As discussed in Section \ref{sec:lit_ipp}, IPP models use the Gaussian process models discussed in Section \ref{sec:spatial_models}.  
\cite{Binney2012} use the average reduction in variance measure as defined in Equation (\ref{f:arv}) in order to determine solution quality.
\cite{Singh2007} propose a formulation based on the concept of mutual information, see Equation (\ref{f:mi}).
Note that both variants are monotonic, i.e., increasing the number of observations cannot decrease the total informativeness \citep{Krause2008,Binney2012}.
This corresponds to the intuitive notion that making more observations cannot decrease the overall information gain.

\subsection{Objective functions for approximating spatial interdependencies}\label{sec:ipp_discretized_models}

The stochastic informativeness measures discussed above provide sophisticated models for spatial interdependencies but are computationally expensive.
In the following, we derive objective functions that show similar characteristics regarding the spatial aspects of a solution while avoiding the computational overhead of a GP regression.

\paragraph{Approximating spatial correlations}

In statistical interpolation approaches, interdependencies between pairs of locations are described using a kernel function $k$.
Using the correlation rather than the covariance as a normalized measure of spatial similarity, a kernel is a function $k: \mathbb{R}^2 \rightarrow [0, 1]$ that decreases with distance and asymptotically converges to 0 \citep{Stachniss2009}.
In our approach, we represent these interdependencies using discrete weights $w$ defined over the set of candidate locations $\setTargets$ such that $w: \setTargets \times \setTargets \rightarrow [0, 1]$.
A weight of $w_{ij} = 1$ for any $i,j \in \setTargets$ indicates perfect correlation, whereas weights close to 0 mean that the observations at $i$ and $j$ are independent, i.e., no information can be inferred about $i$ upon visiting $j$. 
Observations at larger distances can generally be assumed to be independent of one another.
We can, therefore, make use of local neighborhoods including only pairs of locations between which there is a significant correlation.
To this end, we define a covering neighborhood $\coveringNH_i$ such that
$j \in \coveringNH_i \Leftrightarrow w_{ji} \gg 0$ for $j \neq i$.

\paragraph{Informativeness measures}

\begin{figure}[tb]
\centering
  \includegraphics[width=0.6\textwidth]{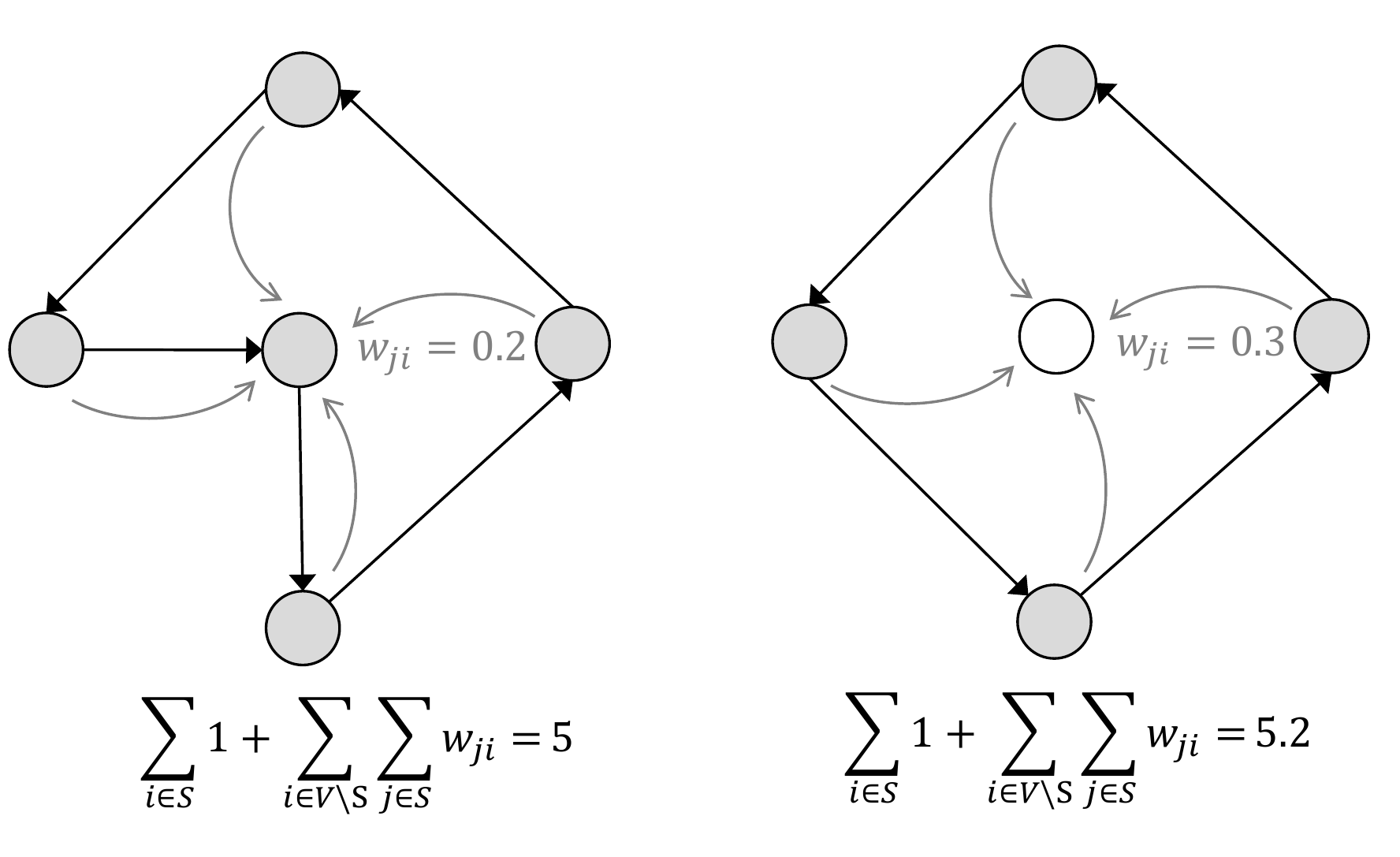}
    \caption[Limitations of the approximation by \cite{Yu2014}]{Illustration of the limitations of the approximation by \cite{Yu2014}. Given are five target locations and weights $w_{ji}$ with respect to the center $i$. The black arrows indicate the optimal route with the objective value given underneath the graph. Visited targets are indicated in grey. The right-hand side illustrates an improving solution obtained by visiting fewer targets.}
    \label{fig:problem_cortop_model}
\end{figure}

The discretized weights represent the similarity between different targets.
Using these weights, we are able to estimate the overall information obtained by a number of sampling locations.
A first model for estimating the overall informativeness based on these discretized weights has been proposed by \cite{Yu2014}.
The weights used in this approximation can be interpreted as the proportion of the information available at $i$ that is obtained upon visiting $j$.
Then, the overall informativeness is modeled as follows:
\begin{align}
& \informativeness^{YU}(\setSelected) = \underbrace{\sum_{i \in \setSelected}  ~1}_{\substack{\text{full information at}\\ \text{visited locations}}} + \underbrace{\sum_{i \in \setTargets\setminus\setSelected} \sum_{j \in \setSelected \cap \coveringNH_i} w_{ji}}_{\substack{\text{estimated proportion}\\ \text{of unvisited targets}}} \label{f:yu_approximation}
\end{align} 
In the following, we refer to the corresponding optimization model as IPP-YU.
This approximation, however, has drawbacks in practice.
One problem is illustrated in Figure \ref{fig:problem_cortop_model}, which gives the optimal routes and objective values on two small graphs that only differ in the degree of similarity between a center and the surrounding targets.
On the left-hand side, the corresponding weights are set to $w_{ji} = 0.2$, which means that full information about the graph is only obtained when the center is included in the vehicle tour.
On the right hand side, weights are increased to $w_{ji} = 0.3$.
In this case, the estimated information gain can be increased by leaving the center unvisited.
From a practical point of view and considering the monotonicity of the stochastic models in Section \ref{sec:ipp_probabilistic_models}, this effect is undesirable:
In any realistic setting, it is not possible to increase the information available about a location by not performing an observation.
Consequently, this model is not a valid approximation for the spatial interdependencies that govern the surveyed area.

\cite{Yu2014} avoid this problem by requiring that 
\begin{align}
\sum\limits_{j \in \coveringNH_i} w_{ji} u_i \leq u_i \label{f:dependency_weight_nh}
\end{align}
holds for all $i \in \setTargets$. 
In their model, this is achieved by explicitly setting weights $w_{ji}$ such that
\begin{align}
w_{ji} = \frac{1}{|\coveringNH_i|}, ~~ j \in \coveringNH_i. \label{f:yu_weights}
\end{align}
This model, however, introduces a counter-intuitive dependency between the size of the covering neighborhood and the weights.
In particular, spatial processes with strong correlations and observations that are similar even at larger distances should be associated with an increased size of the covering neighborhood and increased weights for all $i, j \in \coveringNH_i$.
This cannot be represented accurately in the approach by \cite{Yu2014}:
If the neighborhood size $|\coveringNH_i|$ increases as more locations are positively correlated, the weights $w_{ji}$ have to be reduced in order not to violate Equation (\ref{f:dependency_weight_nh}).
This means that information about the stronger spatial interdependencies is lost.

We propose to relax the condition stated in Equation (\ref{f:dependency_weight_nh}).
To avoid that the estimated information gain increases by leaving out target locations (see Figure \ref{fig:problem_cortop_model}), we limit the maximum information gain in a generalized objective function:
\begin{align}
&\informativeness^{GEN}(\setSelected) = {\sum_{i \in \setSelected} ~1} + {\sum_{i \in \setTargets\setminus\setSelected} \min\{1, \sum_{j \in \setSelected \cap \coveringNH_i} w_{ji}\}} \label{f:own_approximation}
\end{align}
We refer to the corresponding planning problem as IPP-GEN in the remainder of this paper.
We model weights using a simple approximation scheme based on inverse distance weighting.
We assign a weight $\bar{w} < 1$ to locations at distance $d^{min}$ and specify all other weights relative to $\bar{w}$ as follows:
\begin{align}
&w_{ji} = \begin{cases}
\bar{w} \cdot \frac{d^{min}}{d_{ji}}, &j \in \coveringNH_i\\
0, &\textup{otherwise.}
\end{cases}\label{f:idw_weights}
\end{align}
Other approaches based on different kernel functions $k$ are equally possible.

\begin{figure}[tbp]
    \centering
    \captionsetup[subfigure]{justification=centering}    
    \begin{subfigure}[b]{0.45\textwidth}
        \includegraphics[width=\textwidth]{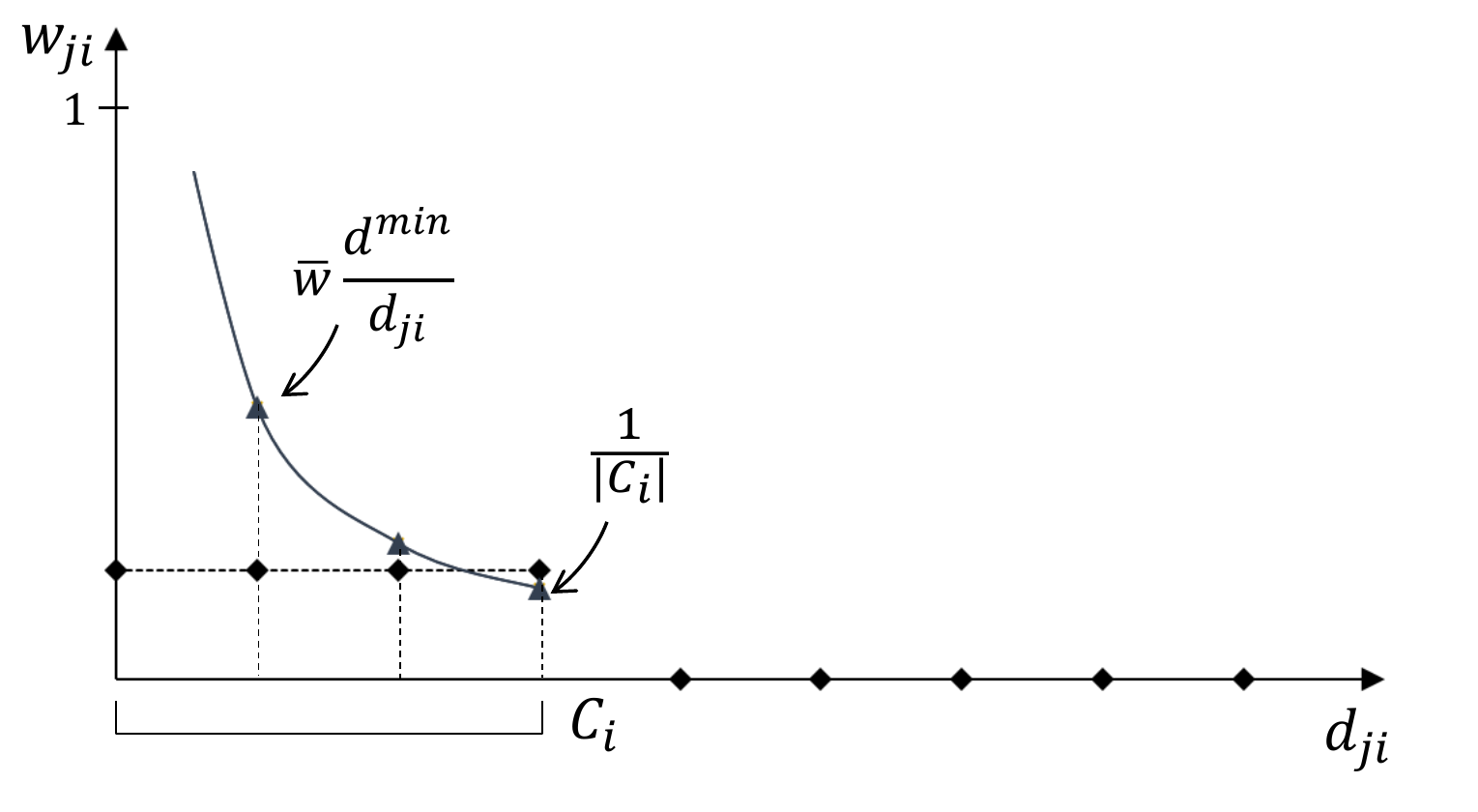}
        \caption{Initial neighborhood size}
        \label{fig:comparison_approximations_0}
    \end{subfigure}
    \begin{subfigure}[b]{0.45\textwidth}
        \includegraphics[width=\textwidth]{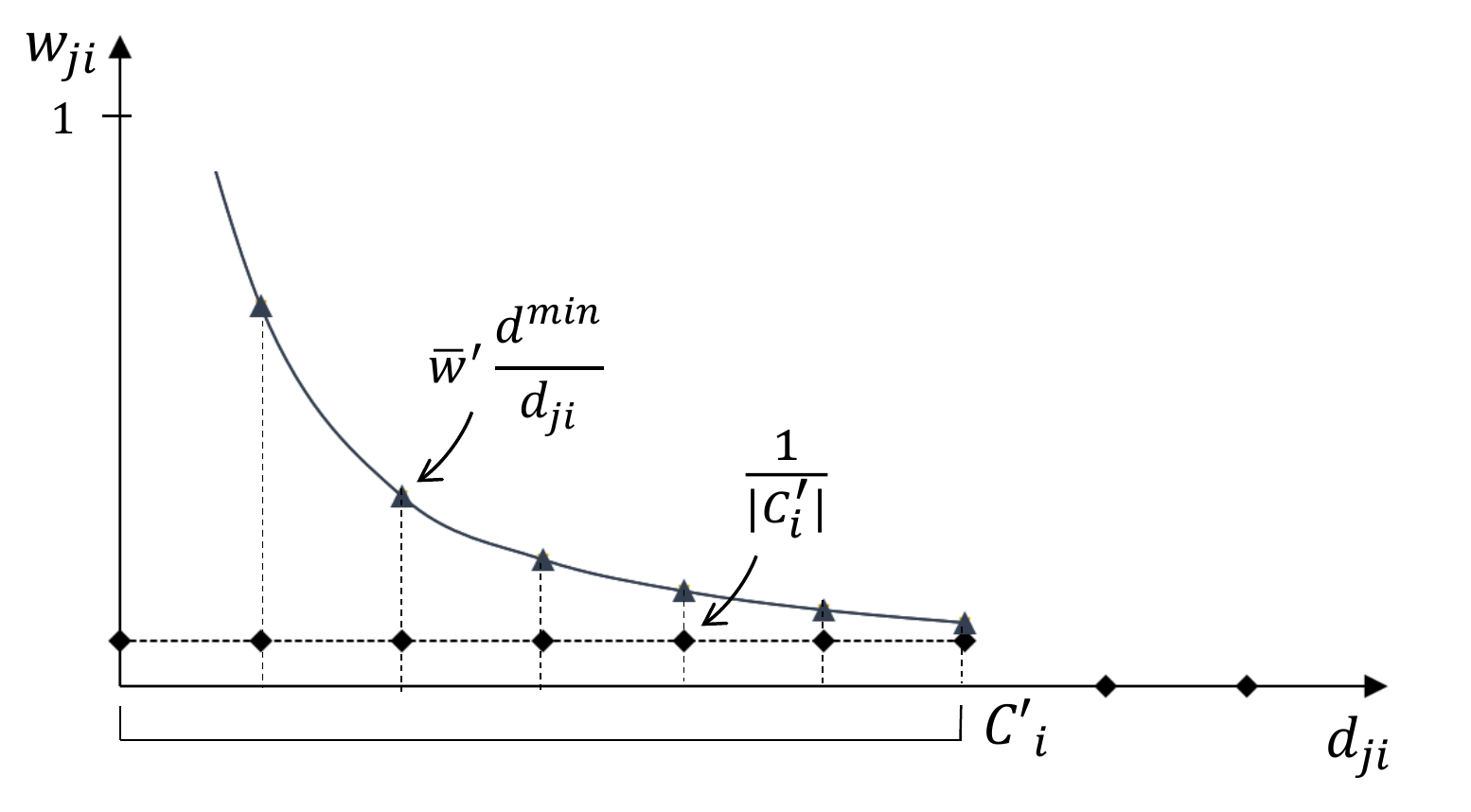}
        \caption{Doubled radius of $\coveringNH_i$}
        \label{fig:comparison_approximations_1}
    \end{subfigure}
    \caption{Impact of increased neighborhood size on approximated weights}\label{fig:comparison_approximations}
\end{figure}

Using such an approximation, we can derive weights $w_{ji}$ that can represent interdependencies of varying strength more accurately.
This is illustrated in Figure \ref{fig:comparison_approximations}:
On the left-hand side, we give approximated weights for our approach and for the model by \cite{Yu2014} for a comparatively small neighborhood, depending on the distance $d_{ji}$ to an observed location $i$.
The right-hand side gives weights for a situation with stronger correlations, represented by an increase in the size of the covering neighborhood.
In the case of the model by \cite{Yu2014}, increasing neighborhood size reduces the weights for all locations. 
In our model, the stronger correlation can be reflected by increasing $\bar{w}$ without side effects. 
Note that similar to the stochastic variants, both models are monotonic in the number of observations.

\subsection{Integrating priorities}\label{sec:ipp_gcortop_models}

Apart from the spatial interdependencies discussed above, UAV missions should also be based on priorities within the target region that characterize how valuable a sample is at a particular location.
As the specification of these priorities depends on the user, they do not necessarily have to be spatially correlated.
The observations made at these locations, in contrast, are always characterized by spatial correlations in our use case.
Hence, some information about a highly prioritized location can also be obtained by making measurements nearby.
The objective function needs to adequately account for this effect.

In this work, we combine these aspects: We model relations between target locations based on the approximated weights introduced in the previous section, and use these weights in the objective function to partially account for the priority of locations that are not visited themselves, but are in the proximity of targets included in a UAV mission.
This way, the fact that some information can be inferred based on observations nearby reduces the relative importance of additional observations at these locations.
\cite{Yu2014} address this by combining the classical TOP, which maximizes direct coverage of priorities, with the approximative model (IPP-YU) in Equation (\ref{f:yu_approximation}).
This yields a mixed integer quadratic programming (MIQP) formulation, denoted as the correlated team orienteering problem (CorTOP) by the authors, which is based on the following objective:
\begin{align}
& \informativeness^{CorTOP}(\setSelected) = \sum_{i \in \setSelected} u_i + \sum_{i \in \setTargets\setminus\setSelected} \sum_{j \in \setSelected \cap \coveringNH_i} w_{ji}u_i  \label{f:cortop_objective}
\end{align}
Equation (\ref{f:cortop_objective}) maximizes the sum of the priorities associated with selected sensing locations $\sum_{i \in \setSelected} u_i$.
Unobserved locations are partially taken into account, with their total contribution to the objective depending on the selected targets within their covering neighborhoods.
For an unobserved location $i$, the value of the indirectly inferred information is approximated as $\sum_{j \in \setSelected \cap \coveringNH_i} w_{ji}u_i$.
Please note that we adapted the formulation of \cite{Yu2014} for reasons of consistency to our notation. 
The original MIQP formulation corresponds to 
$\informativeness^{CorTOP}(\setSelected) = \sum_{i \in \setTargets}x_i u_i + \sum_{i \in \setTargets}\sum_{j \in \coveringNH_i} x_j(x_j - x_i) w_{ji} u_i $ 
with binary decision variable $x_i = 1$ if $i \in \setSelected$ and 0 otherwise.

This model suffers from the same problems as IPP-YU, see Section \ref{sec:ipp_discretized_models}.
We derive a new model based on the relaxed variant IPP-GEN.
We refer to the corresponding model as the generalized correlated team orienteering problem (GCorTOP) based on the following objective function: 
\begin{align}
& \informativeness^{GCorTOP}(\setSelected) = \sum_{i \in \setSelected} u_i + \sum_{i \in \setTargets\setminus\setSelected}\min\{u_i, \sum_{j \in \setSelected\cap\coveringNH_i}w_{ji}u_i\} \label{f:gcortop_a_objective}
\end{align}
Equation (\ref{f:gcortop_a_objective}) ensures that indirectly collected benefit never exceeds a location's priority. 
Hence, the benefit of each additional observation $j \in \coveringNH_i$ with respect to $i \not \in \setSelected$ is determined as 
\begin{equation}
\min \lbrace w_{ji}u_i, \max \lbrace 0, (1 - \sum_{k \in \setSelected\cap\coveringNH_i}w_{ki})\cdot u_i \rbrace\rbrace, \label{f:gcortop_a_objective_help}
\end{equation}
i.e., the benefit is at most $w_{ji}u_i$, thus imposing a threshold that prevents overestimating a mission's informativeness.
This combines the consideration of priorities as proposed by \cite{Yu2014} with the advantages of the relaxed discretization in Equation (\ref{f:own_approximation}).

\section{Solution approach}\label{sec:solution_approach}

We propose two solution approaches addressing the MPPES: 
an exact algorithm and a heuristic approach.
The exact approach based on bidirectional dynamic programming is introduced in Section \ref{sec:dynamic_programming_approach}.
We then discuss a two-phase multi-start adaptive large neighborhood search (2MLS) that is able to quickly provide solutions to realistic problem instances in Section \ref{sec:alns_overview}. 

\subsection{Exact solution approach}\label{sec:dynamic_programming_approach}

We use bi-directional dynamic programming for solving the MPPES exactly. 
Our algorithm assumes that the objective function $\informativeness(\setSelected)$ is monotonic in the number of selected samples.
We do not impose further requirements on the objective function, which makes this approach applicable to all informativeness measures that we have discussed in the previous section.
However, this means that all UAV routes have to be defined completely in order to determine their quality.

\subsubsection{Generation of vehicle routes} 

As the benefit of individual vehicle missions cannot be evaluated in isolation, i.e., without considering the routes of all other vehicles, we use a giant-tour representation for modeling the routes of all UAVs at once. 
Preliminary tests have indicated that this version is faster than a variant where individual missions are created first for each vehicle and the optimal combination of these tours is determined second.
The solution approach is based on the concepts introduced by \cite{Righini2008}.
The algorithm maintains data structure $\dplabel^{fw}$ and $\dplabel^{bw}$  comprising the generated labels in the forward and backward direction, respectively.
Each label corresponds to a tuple $(\setSelected, m, T_m, i)$ that represents a giant tour through the target locations in $\setSelected \subset \setTargets$.
The routes corresponding to a forward label start at $s_0$, i.e. at the starting node of the first UAV, and end at location $i \in \setAll$.
In the backward direction, routes starting at the ending location of the last UAV $e_{|\setVehicles|-1}$ and continue in the reverse direction until $i$.
The current vehicle, i.e., the UAV visiting location $i$, is indicated as $m$ with $m \in \{0, \ldots, |\setVehicles| - 1\}$. 
$T_m$ stores the travel time associated with the current (partial) routes of vehicle $m$ from $s_m$ to $i$ or from backwards $e_m$ to $i$. 

A label $(\setSelected, m, T_m, i)$ is extended to a label $(\setSelected', m, T_m', j)$ by adding a previously unvisited location $j \in \setTargets \setminus \setSelected$ and updating the set of visited target locations $\setSelected'$ and route cost $T_m'$ accordingly, i.e,\begin{align*}
\setSelected' = \setSelected \cup \{j\}, ~~T'_m = T_m + \tau_{ij}.
\end{align*}
The extension is feasible as long as $T_m' + \tau_{je_{m}} \leq T_m^{\textit{max}}$.
In the backward direction, we extend labels along arcs $(j,i$) starting at possible predecessors $j$ of location $i$.
An extension is not feasible if $T_m' + \tau_{s_{m}j} > T_m^{\textit{max}}$. 
If the maximum route duration of vehicle $m$ would be exceeded for all $j \in \setTargets \setminus \setSelected$, a label is extended by closing the current vehicle route and opening a new one.
In this case, the new label is set to $(\setSelected, m + 1, 0, s_{m+1})$ in the forward direction and $(\setSelected, m - 1, 0, e_{m-1})$ in the backward direction.

Similar to \cite{Righini2008}, we apply resource-based bounding to avoid the generation of redundant labels.
In our case, the critical resource is the flight duration.
Forward labels are extended as long as 
$ m \leq \lceil|\setVehicles| / 2 \rceil - 1 \equiv m^{fw}$.
Correspondingly, backward labels are extended while
$ m \geq |\setVehicles| - m^{fw} - 1 \equiv m^{bw}
$.
If $m$ is not a multiple of 2, the same vehicle is included in forward and backward labels, as $m^{fw} = m^{bw}$.
In this case, extensions in any directions are only performed while $T_m \leq T_{m}^{\textit{max}} / 2$ for all labels with $m = m^{fw} = m^{bw}$.

\subsubsection{Dominance tests}

As stated in the introduction of this section, we do not assume specific properties of the objective function other than monotonicity.
Therefore, dominance checks can only be performed on the selected locations and routing costs when comparing labels. 
Consequently, a label $(\setSelected, m, T_m, i) \in \dplabel^{fw}$ is dominated by another label $(\setSelected', m', T_m', i)\in \dplabel^{fw}$ if (a) $m' < m$ and $\mathcal{S}' \supseteq \mathcal{S}$ or (b) $m' = m$ and either $T_m' = T_m$ and $\mathcal{S}' \supseteq \mathcal{S}$ or $T_m' < T_m$ and $\mathcal{S}' = \mathcal{S}$.
Similarly, in the backward direction, a label $(\setSelected, m, T_m, i) \in \dplabel^{bw}$ is dominated by $(\setSelected', m', T_m', i)\in \dplabel^{bw}$ if (a) $m' > m$ and $\mathcal{S}' \supseteq \mathcal{S}$ or (b) $m' = m$ and either $T_m' = T_m, \mathcal{S}' \supseteq \mathcal{S}$ or $T_m' < T_m, \mathcal{S}' = \mathcal{S}$ hold true.

\subsubsection{Optimal solution}

To compute a feasible solution, we join forward and backward labels $(\setSelected, m, T_m, i) \in \dplabel^{fw}$ and $(\setSelected', m', T_m', j) \in \dplabel^{bw}$ such that $\setSelected \cap \setSelected' = \emptyset$.
If there is an uneven number of vehicles, i.e., $m^{fw} = m^{bw}$, a join is feasible only if $T_m + \tau_{ij} + T_m'\leq T_m^{\textit{max}}$.
The objective value $\informativeness(\setSelected)$ only depends on the visited target locations.
As several solutions may result in the same set of samples, we store already computed objective values in a hash table. 
Among all solutions achieving the same informativeness $\informativeness$, we then select the one with the lowest associated cost.

\subsection{Two-phase adaptive large neighborhood search}\label{sec:alns_overview}

As a generalization of the TOP, the GCorTOP is NP-hard, which results in high computation times even for small instances when solving these problems exactly.
In the following, we propose a two-phase multi-start adaptive large neighborhood search (2MLS) approach for solving the MPPES.
We first give an overview and then describe the constituting elements in depth.

\subsubsection{Overview and algorithm design} \label{sec:search_design}

Existing local search approaches for related problems usually focus on the cost-efficient service of targets. This often results in compact routes with target locations in close proximity.
This is opposed to our goal of observing large areas. 

Our algorithm specifically promotes explorative UAV routes.
The solution scheme is summarized in Algorithm \ref{alg:2mls}.
Phase 1 consists of a multi-start approach for constructing diverse solutions $\solution^{red}_i$ based on aggregated or decomposed problem representations $P^{red}$ (see Section \ref{sec:aggregation_strategies}). 
In each iteration, routes are initialized with seed nodes generated by a k-means++ algorithm (Section \ref{sec:algo_seeds}) and further are completed iteratively using an insertion strategy $\insertStrat$. 
The solution is improved using an adaptive large neighborhood search (ALNS) approach and finally transformed to a solution $\solution_i$ to the initial problem representation $P$.
In Phase 2, an ALNS is applied to the best solution obtained in Phase 1.

\begin{algorithm}
\footnotesize
\caption{Two-phase multi-start adaptive large neighborhood search (2MLS)}\label{alg:2mls}
\KwIn{Problem $P$}   
\KwOut{Solution $\solution^{final}$} 
\tcc{Phase 1: multi-start search on simplified problem representations} 
$\solution^{init} = \emptyset$, $\informativeness(\solution^{init}) = 0$\; 
\For{$i = 1~\textup{until the maximum number of multi-starts}$} {
    determine problem representation $P^{red}$\;
    initialize $\solution^{red}_i$ using a k-means++ algorithm\;
    $\solution^{red}_i \leftarrow \insertStrat(\solution^{red}, P^{red})$\;
    $\solution^{red}_i \leftarrow \mathtt{ALNS}(\solution^{red}_i)$\;
    determine solution $\solution_i$ to initial problem representation $P$ based on $\solution^{red}_i$\;
    \textbf{if} $\informativeness(\solution_i) > \informativeness(\solution^{init})$ \textbf{then} $\solution^{init} = \solution_i$\;
}
\tcc{Phase 2: ALNS on complete problem representation}
$\solution^{final} \leftarrow \mathtt{ALNS}(\solution^{init})$\;
\Return $\solution^{final}$\;
\end{algorithm}

\subsubsection{Adaptive large neighborhood search }

The ALNS used in both phases is an adaptation of the framework proposed by \cite{Pisinger2007}.
In each iteration, one removal strategy $\removalStrat$ and one insertion strategy $\insertStrat$ are selected from the sets of strategies based on their previous successes and failures.
The removal strategy selects $nh$ locations and removes them from the current solution (Section \ref{sec:removal_strategies}).
The removed locations are re-inserted by applying $\insertStrat$ to the partial solution (Section \ref{sec:insertion_strategies}).
The search is guided by a reheating-based acceptance criterion that accepts a non-improving move with a probability depending on the number of iterations since the last improvement (Section \ref{sec:acceptance}).
The ALNS stops if a convergence criterion is met or $\textit{iter}^{\textit{max}}$ is reached.

\subsubsection{Seed nodes} \label{sec:algo_seeds}

In each iteration of the multi-start approach, a seed for every vehicle route is determined using a k-means++ variant \citep{Arthur2006} that can account for route duration restrictions and starting and ending locations.
For each UAV, the algorithm determines the set of reachable locations $\setTargets_m = \{i \in \setTargets: \tau_{s_m i}+\tau_{ie_m} \leq T_m^{max}\}$ 
and then selects a seed location $i \in \setTargets_m$ with a probability proportional to the minimum squared distance between $i$ and any $j \in \setAll$ 
that is either a starting or ending location or has already been selected as seed in another route.
This ensures that the initial seed locations are spread across the area, thereby facilitating the construction of vehicle routes covering large parts of the target region. 

\subsubsection{Aggregation and decomposition strategies}\label{sec:aggregation_strategies}

For emphasizing spatial exploration, we propose two different schemes for providing reduced problem representations: 

\paragraph{Grid-based priority aggregation}

In this aggregation scheme, the target area is divided into evenly sized grid cells $C$. Each cell comprises a number of locations $\setTargets_c \subset \setTargets$ such that $\bigcup_{c \in C}\setTargets_c = \setTargets$.
For each cell $c \in C$, we determine the priority-weighted center $(\xCoord_{c}, \yCoord_{c})$ 
 based on the coordinates $(\xCoord_{i}, \yCoord_{i})$ for $i \in \setTargets_c$ such that $\xCoord_{c} = \frac{\sum_{i \in \setTargets_c}u_i \xCoord_{i}}{\sum_{i \in \setTargets_c}u_i}$ and $\yCoord_{c} = \frac{\sum_{i \in \setTargets_c}u_i \yCoord_{i}}{\sum_{i \in \setTargets_c}u_i}$.
We chose a representative $i_c^\star \in \setTargets_c$ that is closest to this point. 
The sum of priorities within a cell $\sum_{i \in \setTargets_c}u_i$ is assigned to its corresponding representative. 
This results in a set of target locations
\begin{equation}
\setTargets^{red} = \lbrace i_c^\star ~ | ~ c \in C \rbrace.
\end{equation}
In the multi-start approach, we use different levels of aggregation in each iteration to obtain a larger range of candidate solutions.
Specifically, we determine the size $|\setTargets_c|$ of the grid cells $c \in C$ such that either 4, 6 or 9 locations are grouped together, making exceptions for grid cells located at the border of the target region.
This approach allows to quickly identify highly relevant regions within the target area and to compute provisional routes for several vehicles. 
Furthermore, any solution to the reduced problem represents a feasible solution with respect to the initial problem representation.

\paragraph{Vehicle-oriented spatial decomposition}

The spatial decomposition scheme separates the target area into independent sections, which are assigned to one vehicle each, and solved as a series of single-vehicle problems. 
Given the seed routes $r_m = (s_m, c_m, e_m)$ (see Section \ref{sec:algo_seeds}), we determine subsets $\setTargets_m^{red} \subset \setTargets$ for all vehicles such that each target location is assigned to the vehicle whose seed tour it is closest to. 
Ties are broken arbitrarily. 
This makes it possible to apply the decomposition scheme to instances with multiple starting or ending locations.
As every single-vehicle problem contains a non-overlapping subset of candidate locations, 
the combination of routes represents a feasible solution for the original problem.

\subsubsection{Weighted objective function}

During the ALNS, we employ a hierarchical objective function with the distance traveled by the UAVs as a secondary objective.
This ensures that among solutions with similar sensing locations $\setSelected$, the ones with lower resource consumption are preferred.
This is particularly relevant whenever it is possible to achieve complete or near-complete coverage of a region, in which case distance minimization becomes more relevant. 
However, distance minimization is detrimental to our focus on explorative missions, as these typically require long travel times for traversing the entire target area.
Therefore, total route length is included in the objective function by multiplying it with a sufficiently small factor of $\varrho > 0$ to adjust its relative weight.
For each problem instance, we heuristically set $\varrho$ based on the sum of priorities $\sum_{i \in \setTargets} u_i$ and the average distance $\bar{d} = \frac{\sum_{i \in \setTargets}\sum_{j \in \setTargets}d_{ij}}{|\setTargets|^2}$ such that
\begin{equation}
\varrho = 0.01 \frac{\sum_{i \in \setTargets} u_i}{|\setTargets| \bar{d}}.
\end{equation}

\subsubsection{Removal strategies}  \label{sec:removal_strategies}

The basic removal strategy of our 2MLS is depicted in Algorithm \ref{alg:removal}.
The set of \emph{planned} locations $\setIncluded$ contains locations included in a route, 
while the set of \emph{unvisited} locations $\setNonIncluded$ contains locations not selected to be visited in the current solution. 
Note that we explicitly select locations among the set $\setNonIncluded$ for removal instead of considering all unvisited locations as candidates for re-insertion.
Firstly, this reduces the overall neighborhood size and ensures that candidates that are unlikely to be inserted in a tour do not need to be evaluated in the subsequent insertion strategy.
Secondly, this allows us to explicitly create different neighborhoods, e.g., for emphasizing either diversification or more local repair steps.

The number of locations to be removed, i.e. neighborhood size $nh$, is divided among the planned and the unvisited locations according to their proportion in the initial solution.
The planned locations are selected first following a removal heuristic $\delta^{removal-routed}$, resulting in relaxed routes with fewer targets.
Then, the free capacity $T_m^{\textit{free}}$ of each UAV is computed as the difference between the vehicle's maximum route duration $T_m^{\textit{max}}$ and the duration of the relaxed route $T_m'$. 
Afterward, the unvisited locations are selected for removal using $\delta^{removal-unvisited}$.
A feasibility check ensures that for each removed unvisited location, there is at least one vehicle in the relaxed solution that can reach this location. 
The removed locations are stored in the set of \emph{open} locations $\setOpen$.

\begin{algorithm}
\footnotesize
\caption{Basic removal strategy}\label{alg:removal}
\KwIn{Solution $\solution^{init}$ with selected locations $\setSelected$, neighborhood size $nh$
}
\KwOut{Open targets $\setOpen$, remaining planned locations $\setIncluded$, unvisited locations $\setNonIncluded$}
set $\setIncluded = \setSelected, \setNonIncluded = \setTargets \setminus \setSelected, \setOpen = \emptyset$, $nh^{p} = \lceil\frac{|\setSelected|}{|\setTargets|} \cdot nh\rceil$\;
\While{$|\setOpen| < nh^{p}$} {
    select $i \in \setIncluded$ following removal heuristic $\delta^{removal-routed}$\;
    $\setOpen = \setOpen \cup \lbrace i \rbrace$, $\setIncluded = \setIncluded \setminus \lbrace i \rbrace$\;
}
\For{$m \in \setVehicles$}
{
    compute route length $T'_m$ of the relaxed solution\;
    free capacity  $T^{\textit{free}}_m = T_m^{\textit{max}} - T'_m$\;
}
\While{$|\setOpen| < nh$} {
    select $i \in \setNonIncluded$ following removal heuristic $\delta^{removal-unvisited}$\;
    \If{$\exists m \in \setVehicles~ \textup{with minimum insertion cost of}~i~\textup{no larger than}~T^{\textit{free}}_m$}
    {
    $\setOpen = \setOpen \cup \lbrace i \rbrace$, $\setNonIncluded = \setNonIncluded \setminus  \lbrace i \rbrace$\;
    }
}
\end{algorithm}

\paragraph{Sequence-based nearest neighbors}
For this strategy, a starting point for each segment is selected randomly from $\setIncluded$.
The subsequent targets in the given vehicle route are removed until either the ending location is reached or the number of removed locations exceeds a given segment length $len^{segment}$.
The segment length limit depends on the average route size, i.e., $len^{segment} = \lfloor\upsilon \cdot \frac{|\setSelected|}{|\setVehicles|}\rfloor$ with a parameter $\upsilon < 1$.
For each segment, we select the closest remaining locations in $\setNonIncluded$ for removal, as these are the most likely candidates for re-insertion.

\paragraph{Route sparsification}
This strategy randomly selects one vehicle $m \in \setVehicles$ and removes a large number of target locations within this route in segments of length $len^{segment}$.
One location remains between any two consecutive segments.
This allows preserving the general shape of the route.
Unvisited locations are removed from a plan by choosing one location in each of the removed segment at random, and selecting the closest remaining locations among the set $\setNonIncluded$ for removal.

\paragraph{Priority-delta strategy}
This strategy selects the locations associated with the lowest contribution to the objective value among the targets in $\setIncluded$. 
Furthermore, it selects the locations with the highest priority among the reachable unvisited targets in $\setNonIncluded$.
As recommended by \cite{Pisinger2007}, both the selection of routed and unvisited locations are randomized.
To this end, locations in $\setIncluded$ are sorted by ascending priority, and locations in $\setNonIncluded$ by descending priority.
Given a randomization parameter $\textit{det}$ and a random number $p \in [0,1]$,  the $l$-th location is selected with $l = \lfloor p^{\textit{det}} |\setIncluded|\rfloor$ or $l = \lfloor p^{\textit{det}} |\setNonIncluded|\rfloor$, respectively.

\paragraph{Region-based selection}
This strategy accounts for priorities of other locations within a limited distance $d^{\textit{limit}}$ of a target instead of evaluating each location's priority in isolation. 
The value $d^{limit}$ is randomized in each iteration.
Based on this value, we define the cost-benefit ratio $\textit{ratio}_i$ of location $i$ as 
\begin{equation}
\textit{ratio}_i = \frac{\detour_{i,\textit{min}}}{\sum_{j \in \setTargets : d_{ij} < d^{limit}} u_j}.
\end{equation}
$\detour_{i,\textit{min}}$ corresponds to the minimum detour needed to insert a location $i$ at the best insertion position over all routes of the current solution. 
If a feasible insertion of location $i$ is not possible, $\detour_{i,\textit{min}}$ is set to $\infty$.
Locations in $\setIncluded$ are sorted by ascending $\textit{ratio}_i$, while those in $\setNonIncluded$ are sorted by descending $\textit{ratio}_i$. 
The selection is randomized using a determinism parameter $\textit{det}$ as above. 

\paragraph{Worst angle removal}
This strategy selects combinations of target locations that form sharp angles in the route. 
Sharp angles can be associated with considerable detours and imply that at least two targets in the route are located relatively close to one another, which is rarely beneficial in terms of information gain.
Therefore, in this strategy, we remove all three nodes related to the sharpest angles. 
Unvisited locations are selected randomly from $\setNonIncluded$.

\subsubsection{Insertion strategies} \label{sec:insertion_strategies}

The basic heuristic for constructing solutions is depicted in Algorithm \ref{alg:construction}.
The heuristic iteratively selects a location $i$ among the set of {open} locations $\setOpen$ following an insertion strategy $\insertStrat$.
Each selected location is inserted at its minimum cost position and added to the set of {planned} locations.
If no feasible insertion position is available, a location is left unvisited and added to the set of {unvisited} locations $\setNonIncluded$.
In order to prevent the repeated construction of similar solutions, the insertion heuristic is slightly randomized such that a random location is selected with probability $\insertRandFactor\in[0,1]$ instead of following $\insertStrat$.

\begin{algorithm}
\footnotesize
\caption{Basic insertion strategy}\label{alg:construction}
\KwIn{Partial solution $\solution^{init}$ with open targets $\setOpen$, planned locations $\setIncluded$, unvisited locations $\setNonIncluded$, randomization parameter $\insertRandFactor$
}
\KwOut{Completed solution $\solution^{final}$ with selected targets $\setSelected = \setIncluded$}
\While{$\setOpen \neq \emptyset$} {
    determine a random number $p \in [0,1]$\;
    \eIf{$p \leq \insertRandFactor$} {
        select $i \in \setOpen$ at random\;
    }{
        select $i \in \setOpen$ following construction heuristic $\insertStrat$\;
    }
    \eIf{$\exists \textup{ feasible insertion position in }\setIncluded$}{
        insert $i$ at its minimum cost position\;    
        $\setIncluded = \setIncluded \cup \lbrace i \rbrace$\;
    } {
        $\setNonIncluded = \setNonIncluded \cup \lbrace i \rbrace$\;
    }
    $\setOpen = \setOpen \setminus \lbrace i \rbrace$\;
}
\end{algorithm}

Visit insertion relies on the marginal benefit of inserting a target location, i.e., the incremental change of the objective caused by including a location in a route, given the current partial solution.
To this end, we distinguish between the remaining priority $\remainingUtility_{i}$ and marginal contribution $\marginalUtility_{i}$ of target $i$.
The remaining priority $\remainingUtility_{i}$ reflects the proportion of the location's priority value that has not yet been covered by visiting nearby target locations. 
The marginal contribution $\marginalUtility_{i}$ indicates the incremental change of the objective function when $i$ is included in a UAV route.
For the GCorTOP, the remaining priority of a location $i$ is: 
\begin{equation}
\remainingUtility_{i} = 
\max\lbrace 0, (1 - \sum_{j \in \setSelected\cap\coveringNH_i} w_{ji})\cdot u_i \rbrace.\label{f:remaining_utility_a}
\end{equation}
The marginal benefit of a location $i$ can then be determined as:
\begin{equation}
\marginalUtility_{i} = \begin{cases}
\remainingUtility_{i} + \sum_{j \in \setTargets \setminus \{i\}: i \in \coveringNH_j} \min \lbrace \remainingUtility_{j}, w_{ij}u_j \rbrace  & \text{ if } i \in \setOpen \\
0 &\text{ otherwise,}
\end{cases}
\end{equation}
i.e., it comprises the remaining priority of location $i$ plus the priorities of all locations that can be surveyed by $i$.

\paragraph{Maximum marginal priority insertion}
The {maximum marginal priority} insertion strategy greedily inserts the visit with the highest marginal contribution to the objective value.
In the case of a tie, the visit with the highest priority $u_i$ is inserted first. 

\paragraph{Best marginal priority ratio insertion}
The {best marginal priority ratio} strategy seeks to balance marginal benefit and resource utilization.
It selects a visit $i \in \setOpen$ such that
\begin{equation}
i = \arg\max_{\setOpen} \frac{(\marginalUtility_{i})^q}{\detour_{i,\textit{min}}}. 
\end{equation}
The factor $q > 0$ allows to adjust the relative importance of $\marginalUtility_i$ and $\detour_{i,\textit{min}}$.
Increasing $q$ leads to a higher focus on priorities during construction.
Higher factors would lead to results similar to the maximum priority strategy.
Again, ties are broken by the locations' direct priorities $u_i$.

\paragraph{Region-based insertion}
Similar to the corresponding removal strategy, the {region-based insertion strategy} accounts for the priorities of nearby locations. 
Locations are selected such that
\begin{equation}
i = \arg\max_{\setOpen}  \frac{\sum_{j \in \setTargets : d_{ij} < d^{limit}} \remainingUtility_{j}}{\detour_{i,\textit{min}}}.
\end{equation}
The objective is to prioritize highly relevant regions early during the \mbox{(re-)construction of routes}.
Note that we use remaining priorities $\remainingUtility_i$ instead of the marginal contributions $\marginalUtility_i$.
Otherwise, we would consider the locations' priorities several times 
as they are considered in the marginal priority values of all their respective neighbors.
The size $d^{limit}$ is randomized in order to insert locations in different orders during the search.

\paragraph{Cost-greedy insertion}
The {cost-oriented insertion strategy} greedily inserts visit $i$ with minimum insertion cost.
We use this strategy for diversification during the search, as most other strategies focus on the locations' priorities. 
Furthermore, in case of the TOP, it may help to construct and improve routes that include clustered locations.

\paragraph{Orienteering-regret insertion}

The {orienteering-regret} heuristic inserts visit locations first that may no longer be feasible if their insertion is delayed.
In order to use this strategy, we have to compute two values for each target remaining in $\setOpen$:
the change in the objective value at the current best position for visit $i$, denoted $\delta^{obj}_{i,1}$, and the change in the objective if $i$ is inserted in its $n$-th best position, denoted $\delta^{obj}_{i,n}$, with regret parameter $n \in \{2,\ldots,|\setIncluded|\}$.
If the $n$-th best position is not a feasible insertion position, we set $\delta^{obj}_{i,n} = 0$.
Then, the heuristic selects a visit $i$ such that
\begin{equation}
i = \arg\max_{\setOpen} |\delta^{obj}_{i,n} - \delta^{obj}_{i,1}|.
\end{equation}
If some locations only have few feasible insertion positions left, setting $\delta^{obj}_{i,n} = 0$ ensures that these are associated with high regret values.
In this case, the strategy prioritizes the locations that offer the highest marginal contribution to the objective, offering additional diversification as locations that are further away from a current route are inserted early.
If the $n$-th best position exists for all $i \in \setOpen$, the difference between $\delta^{obj}_{i,n}$ and $\delta^{obj}_{i,1}$ reflects the change in total route distance due to the weighted objective function. 
In this situation, the insertion strategy first selects locations where a delay may lead to a high detour. 
Similar to the cost-greedy strategy, this improves routes in terms of total duration.
If no feasible position remains, $\delta^{obj}_{i,1}$ is 0 and the location is added to set $\setNonIncluded$.

\subsubsection{Reheating-based acceptance criterion}\label{sec:acceptance}

As acceptance criterion, we employ a reheating-based scheme that increases the likelihood of accepting a move associated with a decreasing objective value with the number of moves since the last strict improvement of the objective function.
This way, subsequent search steps focus on intensification after an improving move. 
The acceptance threshold is only lowered if such improvements cannot be found, thereby accepting solutions with decreasing objective values in order to leave local minima.

Similar to simulated annealing, a new solution $\solution^{new}$ is accepted with a probability that depends on the best solution found so far $\solution^{best}$ and a current temperature $\temperature$.
The probability of accepting a non-improving move with an absolute decrease in solution quality of $\delta^{obj} = |\objVal^{best} - \objVal^{new}|$ is determined as:
\begin{equation}
P\left(\delta^{obj}\right) = \exp\left(\frac{-\delta^{obj}}{\temperature}\right).\label{f:sa}
\end{equation}
The temperature $\temperature_{i}$ in iteration $i$ is increased in non-improving iterations such that $\temperature_{i+1} = \temperature_{i} \cdot \heating$ with $\heating > 1$.
In the beginning and whenever a new best solution is found, $\temperature_i$ is set to a given minimum temperature $\temperature_0 = 1$.
We initialize heating parameter $\heating$ such that a decrease in solution quality of $\delta^{obj}$ will be accepted with probability $p = 0.5$ after $\kappa$ iterations.

As proposed by \cite{Kilby2011a}, we can use Equation (\ref{f:sa}) for computing an acceptance threshold $\currentObjLB$ prior to the construction of a solution.
An insertion step can be stopped preemptively when the maximum achievable priority value $\currentObjUB$ of a solution is less than $\currentObjLB$.
We therefore maintain an upper bound on the objective value of GCorTOP, which is computed as follows:
\begin{align}
\currentObjUB & = \sum_{i \in \setTargets} u_i - \sum_{i \in \setNonIncluded} ( u_i - \min\lbrace u_i, \sum_{j \in \coveringNH_i \setminus \setNonIncluded} w_{ji}u_i\rbrace) \label{f:ub_utility-a}
\end{align}
The upper bound in Equation (\ref{f:ub_utility-a}) corresponds to the sum of all priorities reduced by the sum of ``lost'' priority values, i.e., the proportion of priorities associated with locations in $\setNonIncluded$ that cannot be covered by any of the locations in $\setIncluded$ and $\setOpen$, respectively.

\section{Experimental study}\label{sec:cortop_study}

This section provides an in-depth evaluation of the proposed models and solution approaches.
It is dedicated to two main aspects: (1) The evaluation of the performance of the {solution approach} considering solution quality, robustness, and computation times, and (2) the assessment of the validity of the proposed {models} to the mission planning problem introduced in Section \ref{sec:problem_statement}.

\subsection{MPPES benchmark instances}\label{sec:new_benchmarks}

Our 2MLS heuristic addresses specific features of the GCorTOP, however, it is still applicable to TOP instances. 
For comparing our approach to state-of-the-art algorithms for related problems, we solve benchmark instances proposed by \cite{Chao1996top} and \cite{Dang2013pso} and compare them to the best approaches in terms of solution quality and runtime. 

We furthermore propose new benchmark instances for MPPES based on the practical use case introduced in Section \ref{sec:use_case}.
An overview of the parameters of the benchmark instances is given in Table  \ref{table:cop_benchmarks}.
In this study, we consider 5 different target area sizes.
For every target area size, we generate 90 instances by varying the assigned priorities, the number of UAVs and maximum flight time as indicated in the table. 
This leads to a total of 450 benchmark instances.

\begin{table}
\centering
\footnotesize
\begin{tabular}{ p{0.3\textwidth} p{0.4\textwidth} }
\hline\noalign{\smallskip}
Parameter & Values \\
\hline\noalign{\smallskip}
Cruise speed $v^{max}$ (in m/s) & 7 \\
Acceleration $a$ (in $\text{m}/\text{s}^2$) & 2 \\
Sensing time $\tau_i$ (in s) & 2 \\
Available UAVs & $\{1;~ 2;~ 3\}$\\
Mission duration $T^{max}$ (in s) & $\{600;~ 900;~ 1200;~ 1500; ~1800\}$ \\
Target area sizes ($\textup{km}^2$)& $\{1.5 \times 1.5; ~1.5 \times 2; ~2 \times 2; ~2.5 \times 2; ~2.5 \times 2.5\}$\\
\noalign{\smallskip}\hline\noalign{\smallskip} 
\end{tabular}
\caption{Parameter settings for MPPES benchmarks}
\label{table:cop_benchmarks}
\end{table}

\paragraph{Target areas and priorities}
We vary target area size between $1.5 \times 1.5 ~\textup{km}^2$ and $2.5 \times 2.5 ~\textup{km}^2$.
Smaller affected areas either do not require airborne surveillance or can be surveyed in their entirety within the available flight time limit.
We do not consider larger areas, as they are not within the UAVs' operation range.
For the MPPES benchmarks, we select target areas within the German state of North Rhine-Westphalia at random.
For each area, we determine priorities using the corresponding data on population density published by the Federal Statistical Office based on data collected in the 2011 census \citep{destatis2018}.
This dataset provides georeferenced information on population density in a resolution of $100 \times 100$~m, which is sufficient for our use case.

\paragraph{Vehicles}

In our study, we consider UAV systems similar to those used in the BigGIS project (see Section \ref{sec:uav_system}).
We generate instances with 1 to 3 UAVs.
The starting and ending position for each UAV is selected at random among the locations at the border of the target area.
To obtain insights on the trade-off between flight time and information gain, we vary the maximum flight time $T^{max}$ between 10 and 30 minutes. 
We assume a maximum cruise speed $v^{max} = 7~\text{m}/\text{s}$ (approx. $25~\text{km}/\text{h}$), which is realistic when taking payload weight and environmental influences into account.
Once a target location is reached, the observations are made at a standstill.
Hence, the time required for accelerating and decelerating between targets needs to be considered when computing total travel time.
For simplicity, we assume that UAVs accelerate and decelerate with a constant horizontal acceleration rate $a =2~\text{m}/\text{s}^2$. 
Then, travel time $\tau_{ij}$ between any two points $i$ and $j$ can be determined depending on the traveled distance $d_{ij}$ as follows: 
\begin{equation}
\tau_{ij} =
\begin{cases}
 2 \sqrt{\frac{d_{ij}}{a}} &\text{ if } d_{ij} < \frac{(v^{max})^2}{a} \\
\frac{d_{ij}}{v^{max}} + \frac{v^{max}}{a} &\text{ otherwise}
\end{cases}
\end{equation}
We furthermore require a fixed sensing time $\tau_i = 2$~s.
This is necessary as the UAVs need to verify their position via GPS, their flight needs to be stabilized at the target location, and several observations are performed and processed before traveling further.

\paragraph{Sensor equipment}

In our experiments, we assume that UAVs are equipped with a hyperspectral sensor with a resolution of 1000 $\times$ 1000 pixels and a focal length of 12 mm.    
Based on experiments in the BigGIS project, we consider a ground sampling resolution of $0.01~\text{m}^2/\text{pixel}$ to be acceptable. 
Given this information, we can determine the covered ground area of $100 \times 100$~m per image at a flight altitude of approximately 120~m.
Candidate sampling locations, which represent the center of an image taken by the UAVs, are located at 100~m distance to one another.
Given the target area sizes discussed above, this results in $15 \times 15 = 225$ to $25 \times 25 = 625$ candidate target locations.

\paragraph{Spatial distribution of contaminants}

The applicability of the proposed model largely depends on the predictive quality that can be achieved for the distribution of hazardous substances.
In order to assess this aspect, we generate spatially autocorrelated random fields that represent distributions of contaminants across the target area.
For this purpose, we define a Gaussian process $Z_\setTargets$ and create artificial distributions by drawing samples from this distribution (see Section  \ref{sec:spatial_interpolation}).
For each instance, this yields simulated values $Z(i)$ for all $i \in \setTargets$ that follow the specified covariance matrix of the GP.
We normalize these values such that $Z(i) \in [0, 100]$ for all $i$ with a mean value of 50.
The covariance functions used in the GP are based on \cite{Stachniss2009} who use an exponential kernel as well as a Mat\'{e}rn covariance function for modeling and predicting gas distributions.
We alternate between these two kernel functions when generating random fields.
Exemplary distributions are depicted in the visual interpretation of results at the end of this section. 

\subsection{Model configurations}\label{sec:model_config}

We compare three modeling variants:
The TOP, which only considers priorities, CorTOP (Equation (\ref{f:cortop_objective})) as a predecessor to our model with a lower emphasis on covering aspects, and finally, the GCorTOP model (Equation (\ref{f:gcortop_a_objective})).

In case of the latter two models, we have to specify the weights that define the objective function.
For CorTOP, these are based on the approximation used by \cite{Yu2014}, see Equation (\ref{f:yu_weights}), with the covering neighborhood $\coveringNH_i$ of location $i$ made up of its immediate neighbors.
For GCorTOP, our model is based on Equation (\ref{f:idw_weights}) with $\bar{w}=0.5$ and $d^{min}=100$~m.
The radius of the covering neighborhood is limited to $400$~m, as the impact of including locations at larger distances in the covering neighborhood is negligible.

\subsection{Results}\label{sec:computational_study_results}

The 2MLS is implemented in C++.
The dynamic programming approach is implemented in Python 3.
All evaluations are performed on an Intel Xeon 2.6GHz processor machine with 119 GB RAM running a 64-bit version of Windows Server 2012.
Except for the parameter tuning, each instance was solved 10 times.

\subsubsection{Parameter Tuning}

To determine good parameter settings, we performed a series of preliminary experiments on a subset of $60$ randomly selected instances among the TOP and MPPES instances.
The idea behind this approach is to find reasonable parametrizations that perform well on a diverse set of instances.
For each instance and parameter configuration, we ran the 2MLS three times.
The performance of different configurations are compared based on the gap to the best known solution (BKS) averaged over all instances.

Based on the results, we propose two parametrizations of the 2MLS: 
a ``regular'' configuration achieving good results (2MLS), and a particularly fast parametrization denoted as 2MLS-f.
This distinction allows us to gain insights into the trade-off between computation time and solution quality.
Furthermore, using the fast version, we can evaluate how the algorithm performs in a practical setting where computation times are strictly limited.
Table \ref{table:search_configuration} gives the parameter settings for the two configurations.
For the ALNS procedure, with the exception of the parameters explicitly mentioned, we follow the settings proposed by \cite{Pisinger2007}. 

\begin{table}
\centering\footnotesize
\begin{tabular}{ p{0.3\textwidth}  p{0.28\textwidth}  p{0.28\textwidth}}
\hline\noalign{\smallskip}
Parameter  & 2MLS-f & 2MLS \\
\hline\noalign{\smallskip}
Strategy for Phase 1 & Vehicle decomposition & Aggregation \\
Number of multi-starts & 4 & 4 \\
ALNS limit in Phase 1  & 100 iterations & 100 iterations \\
ALNS limit in Phase 2  & 2000 iterations & 2000 iterations \\
ALNS segment size & 200 iterations & 600 iterations \\
Convergence limit & 100 iterations    & 300 iterations \\
Neighborhood size limits &$nh^- = \max\{0.05 |\setTargets|, 10\}$ \newline $nh^+ = \min\{0.2|\setTargets|, 100\}$&    $nh^- = \max\{0.1 |\setTargets|, 10\}$ \newline $nh^+ = \min\{0.3|\setTargets|, 100\}$ \\
Priority ratio factor $q$ & $\{1, 2\}$ & $\{1, 2\}$\\
Segment length $\upsilon$ & $[0.1 \frac{|\setSelected|}{|\setVehicles|}, 0.25 \frac{|\setSelected|}{|\setVehicles|}]$& $[0.15 \frac{|\setSelected|}{|\setVehicles|}, 0.35 \frac{|\setSelected|}{|\setVehicles|}]$ \\
Randomization factor $\insertRandFactor$    & 0.05    & 0.05 \\
Regret parameter $n$ & 2 & 2 \\
Determinism parameter $\textit{det}$ & 6 & 6 \\
Heating parameter $\kappa$    & 400 iterations & 400 iterations \\
\hline\noalign{\smallskip} 
\end{tabular}
\caption{Search configurations for computational experiments }
\label{table:search_configuration}
\end{table}

\subsubsection{Results on TOP benchmarks}

We compare results of the 2MLS with other approaches representing the state of the art with respect to computation time or solution quality on TOP benchmark instances.
We selected two approaches delivering acceptable solutions in relatively short run times, the fast path relinking (FPR), proposed by \cite{Souffriau2010} and the multi-start local search (MS-LS) presented by \cite{Vidal2015},
Furthermore, we show the results of three approaches aiming for very good solution quality at the cost of longer run times, the particle swarm optimization-inspired algorithm (PSOiA) proposed by \cite{Dang2013pso}, unified hybrid genetic search (UHGS) proposed by \cite{Vidal2015}, and pareto mimic algorithm (PMA) by \cite{Ke2016}.
With the exception of two newer machines used by \cite{Vidal2015} and \cite{Ke2016}, the processors used are of a similar generation as ours and can provide a rough indication of computation times. 

Table \ref{table:summary_chao_results} gives an overview of the results, showing the average and best deviation to the BKS over 10 runs and the average computation time. 
Please note that we selected the instances commonly used in literature\footnote{The publications by \cite{Dang2013pso} and \cite{Ke2016} that report on the newer instances limit their study to a subset of 82 ``difficult'' instances in the benchmark set.
The comparison of results on the entire benchmark set is based on the results reported on the website of the authors of the second journal paper (\url{http://gr.xjtu.edu.cn/web/keljxjtu/download}, accessed 23.09.2018)}.

\begin{table}[tbp]
  \centering \footnotesize
    \begin{tabular}{cclrrrrrrr}
    \toprule
& & & {FPR\textsuperscript{1}} & {PSOiA\textsuperscript{2}} & UHGS\textsuperscript{3} & {MS-LS\textsuperscript{3}} & PMA\textsuperscript{4} & 2MLS-f & 2MLS \\
    \midrule
 \multirow{15}{*}{Chao et al.}  &   & avg.  & -- & 0.11~\% & 0.05~\% & 1.52~\% & 0.32~\% & 3.18~\% & 1.75~\% \\
   & set 4      & best  & 0.73~\% & 0.00~\% & 0.01~\% & 0.21~\% & 0.00~\% & 0.90~\% & 0.36~\% \\
     &     & t(s)  & 8.6   & 226.7 & 236.3 & 15.9  & 113.1 & 4.5   & 22.6 \\
    \cmidrule{2-10} 
      &   & avg.  & -- & 0.03~\% & 0.01~\% & 0.48~\% & 0.09~\% & 1.67~\% & 0.64~\% \\
    & set 5     & best  & 0.23~\% & 0.00~\% & 0.00~\% & 0.06~\% & 0.00~\% & 0.14~\% & 0.09~\% \\
     &     & t(s)  & 2.9   & 73.9  & 138.0 & 3.4   & 32.3  & 1.3   & 12.0 \\
    \cmidrule{2-10} 
      &   & avg.  & -- & 0.00~\% & 0.00~\% & 0.34~\% & 0.29~\% & 1.10~\% & 0.10~\% \\
     & set 6     & best  & 0.11~\% & 0.00~\% & 0.00~\% & 0.03~\% & 0.00~\% & 0.00~\% & 0.00~\% \\
      &    & t(s)  & 2.1   & 37.3  & 91.0  & 2.0   & 24.8  & 0.8   & 7.3 \\
    \cmidrule{2-10} 
       &  & avg.  & -- & 0.03~\% & 0.02~\% & 1.11~\% & 0.13~\% & 1.29~\% & 0.88~\% \\
     & set 7     & best  & 0.54~\% & 0.00~\% & 0.00~\% & 0.14~\% & 0.00~\% & 0.28~\% & 0.05~\% \\
      &    & t(s)  & 6.3   & 130.3 & 228.0 & 9.8   & 68.6  & 2.7   & 22.1 \\
    \cmidrule{2-10} 
 & & avg.  & -- & 0.06~\% & 0.03~\% & 0.99~\% & 0.20~\% & 2.03~\% & 1.04~\% \\
  &   \multicolumn{1}{l}{ all}      & best  & 0.47~\% & 0.00~\% & 0.00~\% & 0.13~\% & 0.00~\% & 0.43~\% & 0.17~\% \\
       &   & t(s)  & 5.0   & 138.4 & 192.0 & 9.3   & 69.3  & 2.8   & 18.0 \\
    \midrule
 \multirow{6}{*}{Dang et al.}  & & avg. & --& 0.50~\% & --& --& 0.42~\% & 2.86~\% & 1.86~\% \\
  & \multicolumn{1}{l}{sel. 82}    & best & --& 0.04~\% & --& --& 0.00~\% & 1.13~\% & 0.55~\% \\
       &    & t(s) & --& 11031.0 & --& --& 999.2 & 25.8 & 70.5 \\
    \cmidrule{2-10} 
   & & avg. & -- & 0.13~\% & -- & -- & 0.13~\% & 1.62~\% & 1.03~\% \\
  &   \multicolumn{1}{l}{all}  & best & -- & 0.01~\% &-- & -- & 0.00~\% & 0.42~\% & 0.19~\% \\
        &  & t(s) & -- & 4379.4 & --& --& 343.9 & 10.7  & 32.8 \\
    \bottomrule
\noalign{\smallskip} \multicolumn{9}{l}{\textsuperscript{1} tested on an Intel Xeon 2.5GHz processor machine}\\
\multicolumn{9}{l}{\textsuperscript{2} tested on an AMD Opteron 2.6GHz CPU}\\
\multicolumn{9}{l}{\textsuperscript{3} tested on an Intel Xeon 3.07GHz CPU}\\
\multicolumn{9}{l}{\textsuperscript{4} tested on an Intel Core i5 3.2GHz CPU}\\
    \end{tabular}%
  \caption{Results on \cite{Chao1996top} and \cite{Dang2013pso} instances}
  \label{table:summary_chao_results}%
\end{table}

Compared to the approaches striving for quality, 2MLS comes with a slightly worse solution quality, achieving a best gap of $0.17~\%$ and an average gap of $1.04~\%$.
2MLS-f has a best gap of $0.43~\%$ and an average gap of $2.03~\%$.
However, with an average computation time of 2.8~s, it is faster than any of the other approaches.
The best gap slightly improves on FPR.
For the instances of \cite{Dang2013pso}, the best solutions found by 2MLS are close to the BKS, with an average gap of $1.03~\%$ and a best gap of $0.19~\%$, averaged over all instances.
These gaps are slightly higher when considering the 82 more difficult instances but still remains below $2~\%$ on average.
For 2MLS-f, the average gap of these instances is less than $3~\%$.
Compared to the two other published approaches, this increase in average gap comes with a considerable improvement in terms of computation time:
2MLS is more than ten times faster than PMA, and
2MLS-f reduces average computation time by a factor of 30.
Both variants use less than $1~\%$ of the computation time required for PSOiA.

\subsubsection{Optimality gap on small MPPES benchmark instances} \label{sec:optimality_gap}

The exact solution approach is only applicable for small problem instances.
We were unable to solve any of the large-scale MPPES instances to optimality within reasonable computation time.
We therefore generated a set of smaller instances with 16 to 49 target locations in a rectangular grid based on the same parameters as depicted in Table \ref{table:cop_benchmarks}, except for $T^{max}$, which varies between 100~s and 250~s due to the smaller target areas.
The maximum runtime for the exact approach has been limited to 10000 seconds. 

We briefly give the main results of this evaluation: 
In the single-vehicle case, we were able to find optimal solutions for instances that include around 40 locations or less. 
Instances with two or more UAVs quickly become intractable for around 20 target locations.
This is consistent with the performance of the MIQP model proposed by \cite{Yu2014} for solving the CorTOP.
Unfortunately, a direct comparison of these approaches is not possible, as \cite{Yu2014} have not published their instances.
Compared to the known optimal results, 2MLS-f has an average gap of $0.1~\%$ with an average computation time of $0.22$~s.
2MLS has an average gap of $0.0005~\%$ at a computation time of $1.98$~s.
Both variants are able to find the optimal solution in 53 out of the 55 cases for which we know the exact solution.
Furthermore, for 51 out of 55 instances, 2MLS finds the optimal solution in all ten runs.

\subsubsection{Solution quality and robustness}

Moving on to the large-scale instances proposed in Section \ref{sec:new_benchmarks}, we investigate the running time behavior of the proposed solution approach to verify its applicability in practice.
Our project partners estimate that approximately two minutes are available after the arrival of the response personnel for preparing the UAV missions in order not to delay the surveillance and rescue operation.
The results on the larger instances by \cite{Dang2013pso} indicate that even though 2MLS scales better than other solution approaches, it is still associated with high computational effort for large instances.
This is particularly obvious for the 82 difficult instances.
These instances comprise on average 240 candidate locations. 
The largest instances in our study, are almost three times as large.
Hence, in the following, we focus on the 2MLS-f.

As optimal solutions to the large MPPES instances are not available, Table \ref{table:avg_cortop_gap} reports the average gap to the best found solution.
The minimum average gap is achieved for the smallest instances with high resource availability, i.e., instances with 225 and 300 target locations involving multiple UAVs and long mission durations.
In these instances, the objective value is generally close to the total sum of priorities $\sum_{i \in \setTargets} u_i$.
This means that it is comparatively easy for the solution approach to determine a solution with high coverage, and the solutions found during the search are mostly distinguished by their total duration rather than overall coverage.
The effect is exactly opposite in case of limited resources, i.e., single-vehicle instances with tight duration constraints. 
In these situations, efficient use of the available resources is crucial, and exchanging some locations in a solution can have a significant impact on the objective value.
Consequently, the highest gap of $2.9~\%$ is observed in the instances with 625 candidate locations and a single vehicle.

\begin{table}[tbp]
  \centering \footnotesize
    \begin{tabular}{ccrrrrrr}
    \toprule
    \multirow{2}[4]{*}{$|\setTargets|$} & \multirow{2}[4]{*}{$|\setVehicles|$} & \multicolumn{5}{c}{$T^{\max}$ (in s)}                 & \multicolumn{1}{c}{\multirow{2}[4]{*}{all}} \\
\cmidrule{3-7}          &       & 600 & 900 & 1200 & 1500 & 1800 &  \\
    \midrule
    \multirow{4}[4]{*}{225} & 1     & 1.8~\% & 1.1~\% & 0.5~\% & 0.3~\% & 0.1~\% & 0.7~\% \\
          & 2     & 0.8~\% & 0.1~\% & 0.0~\% & 0.0~\% & 0.0~\% & 0.2~\% \\
          & 3     & 0.2~\% & 0.0~\% & 0.0~\% & 0.0~\% & 0.0~\% & 0.0~\% \\
\cmidrule{2-8}          &       & {0.9~\%} & {0.4~\%} & {0.2~\%} & {0.1~\%} & {0.0~\%} & {0.3~\%} \\
    \midrule
    \multirow{4}[4]{*}{300} & 1     & 2.6~\% & 2.7~\% & 1.5~\% & 0.9~\% & 0.6~\% & 1.7~\% \\
          & 2     & 1.6~\% & 0.5~\% & 0.1~\% & 0.0~\% & 0.0~\% & 0.5~\% \\
          & 3     & 1.1~\% & 0.1~\% & 0.0~\% & 0.0~\% & 0.0~\% & 0.2~\% \\
\cmidrule{2-8}          &       & {1.8~\%} & {1.1~\%} & {0.5~\%} & {0.3~\%} & {0.2~\%} & {0.8~\%} \\
    \midrule
    \multirow{4}[4]{*}{400} & 1     & 1.5~\% & 1.9~\% & 1.5~\% & 1.0~\% & 0.7~\% & 1.3~\% \\
          & 2     & 0.7~\% & 0.9~\% & 0.4~\% & 0.1~\% & 0.0~\% & 0.4~\% \\
          & 3     & 0.3~\% & 0.3~\% & 0.1~\% & 0.0~\% & 0.0~\% & 0.1~\% \\
\cmidrule{2-8}          &       & {0.8~\%} & {1.0~\%} & {0.6~\%} & {0.4~\%} & {0.2~\%} & {0.6~\%} \\
    \midrule
    \multirow{4}[4]{*}{500} & 1     & 1.4~\% & 1.7~\% & 1.6~\% & 1.5~\% & 1.1~\% & 1.5~\% \\
          & 2     & 0.8~\% & 1.0~\% & 0.7~\% & 0.4~\% & 0.2~\% & 0.6~\% \\
          & 3     & 0.4~\% & 0.6~\% & 0.2~\% & 0.0~\% & 0.0~\% & 0.3~\% \\
\cmidrule{2-8}          &       & {0.9~\%} & {1.1~\%} & {0.9~\%} & {0.6~\%} & {0.5~\%} & {0.8~\%} \\
    \midrule
    \multirow{4}[4]{*}{625} & 1     & 2.0~\% & 2.4~\% & 2.9~\% & 2.0~\% & 1.8~\% & 2.2~\% \\
          & 2     & 1.2~\% & 1.6~\% & 1.2~\% & 0.6~\% & 0.5~\% & 1.0~\% \\
          & 3     & 0.8~\% & 1.0~\% & 0.4~\% & 0.2~\% & 0.1~\% & 0.5~\% \\
\cmidrule{2-8}          &       & {1.4~\%} & {1.7~\%} & {1.5~\%} & {0.9~\%} & {0.8~\%} & {1.2~\%} \\
    \midrule
    \multicolumn{2}{c}{all} & {1.2~\%} & {1.1~\%} & {0.7~\%} & {0.5~\%} & {0.3~\%} & {0.8~\%} \\
    \bottomrule
    \end{tabular}
     \caption[Average gap of 2MLS-f]{Average gap to the best found solution on the MPPES benchmark instances}
  \label{table:avg_cortop_gap}%
\end{table}%

\subsubsection{Runtime analysis}

Table \ref{table:avg_computation_times} gives the results in terms of computation time for  2MLS-f. 
The average computation time over all instances is approximately 56~s, i.e., slightly less than a minute and well within the allowed range in our application.
Increasing the number of target locations is almost always associated with an increase in computational effort.
Whereas 24~s suffice for the smallest instances involving 225 targets, this is increased to 96~s on average for the largest set with 625 locations.
The convergence criterion means that search is stopped when the objective value is comparatively high, which leads to lower computation times in case of several UAVs and high flight times. 

\begin{table}[tbp]
  \centering\footnotesize
    \begin{tabular}{ccrrrrrr}
    \toprule
   \multirow{2}[4]{*}{$|\setTargets|$} & \multirow{2}[4]{*}{$|\setVehicles|$} & \multicolumn{5}{c}{$T^{\max}$ (in s)}     & \multicolumn{1}{c}{\multirow{2}[4]{*}{all}} \\
\cmidrule{3-7}          &       & 600   & 900   & 1200  & 1500  & 1800  &  \\
    \midrule
    \multirow{4}[4]{*}{225} & 1     & 30.2  & 36.7  & 33.5  & 27.1  & 21.8  & 29.8 \\
          & 2     & 39.3  & 28.4  & 16.6  & 16.3  & 17.3  & 23.6 \\
          & 3     & 33.1  & 15.1  & 14.9  & 15.0  & 14.4  & 18.5 \\
\cmidrule{2-8}          &       & {34.2} & {26.7} & {21.7} & {19.5} & {17.8} & {24.0} \\
    \midrule
    \multirow{4}[4]{*}{300} & 1     & 30.9  & 44.7  & 51.0  & 48.8  & 37.6  & 42.6 \\
          & 2     & 53.1  & 54.0  & 39.9  & 27.6  & 20.2  & 39.0 \\
          & 3     & 60.5  & 37.9  & 21.9  & 19.0  & 20.6  & 32.0 \\
\cmidrule{2-8}          &       & {48.1} & {45.5} & {37.6} & {31.8} & {26.1} & {37.8} \\
    \midrule
    \multirow{4}[4]{*}{400} & 1     & 38.3  & 47.9  & 55.6  & 59.7  & 57.6  & 51.8 \\
          & 2     & 54.8  & 68.6  & 58.7  & 50.3  & 37.3  & 53.9 \\
          & 3     & 47.6  & 59.6  & 43.9  & 29.4  & 26.6  & 41.4 \\
\cmidrule{2-8}          &       & {46.9} & {58.7} & {52.7} & {46.5} & {40.5} & {49.1} \\
    \midrule
    \multirow{4}[4]{*}{500} & 1     & 42.3  & 52.9  & 69.7  & 80.0  & 71.6  & 63.3 \\
          & 2     & 72.1  & 90.1  & 97.4  & 83.6  & 70.2  & 82.7 \\
          & 3     & 83.2  & 104.0 & 71.2  & 56.5  & 43.6  & 71.7 \\
\cmidrule{2-8}          &       & {65.9} & {82.3} & {79.4} &{73.4} & {61.8} & {72.6} \\
    \midrule
    \multirow{4}[4]{*}{625} & 1     & 46.2  & 57.8  & 77.3  & 92.6  & 99.2  & 74.6 \\
          & 2     & 81.4  & 101.8 & 134.1 & 115.2 & 106.2 & 107.7 \\
          & 3     & 107.7 & 134.7 & 121.4 & 92.9  & 72.1  & 105.7 \\
\cmidrule{2-8}          &       & {78.5} & {98.1} & {110.9} & {100.2} & {92.5} & {96.0} \\
    \midrule
    \multicolumn{2}{c}{all} & {54.7} & {62.3} & {60.5} & {54.3} & {47.8} & {55.9} \\
    \bottomrule
    \end{tabular}%
  \caption[Average computation time of 2MLS-f ]{Average computation time of 2MLS-f on the MPPES benchmark instances}
  \label{table:avg_computation_times}%
\end{table}%

\subsubsection{Model evaluation and comparison}\label{sec:model_evaluation_comparison}

In this section, we study the results that are obtained using this approach from a practical point of view, i.e., we seek to determine whether valuable information can be obtained during the UAV missions.

\paragraph{Covered priorities}

The relative importance of target locations is considered in form of the priorities $u_i$.
In order to account for information about neighboring locations, we measure direct and indirect coverage using a distance-dependent coverage measure $\textit{PCov}_{d}$
\begin{align}
\textit{PCov}_{d}(\setSelected) = \frac{\sum_{i \in \setTargets: \exists j \in \setSelected \text{ with } d_{ij} \leq d} u_i}{\sum_{i \in \setTargets} u_i},\label{f:pcov_d}
\end{align}
which gives the proportion of priorities that are within distance $d$ to a sampled location. 
This criterion is independent of $\informativeness(\setSelected)$, thus allowing a comparison of all discussed models.

In Table \ref{table:pcov_comparison}, we report the average percentage of priorities visited directly by the UAVs ($\textit{PCov}_{0}$) or within 100 m or 300 m of a visited location ($\textit{PCov}_{100}$ and $\textit{PCov}_{300}$).
Considering only direct coverage, we can see that TOP clearly performs best.
This meets our expectations, as this model does not involve any trade-off between spatial coverage and priorities of directly visited locations.
Results obtained using CorTOP are similar to these results, whereas the GCorTOP yields the overall lowest values for $\textit{PCov}_{0}$.
This changes when accounting for covered priorities in the vicinity of the UAV missions.
GCorTOP consistently outperforms the two other models with respect to the $\textit{PCov}_{100}$ and $\textit{PCov}_{300}$ measures.
Even though fewer high-priority target locations are directly visited, more of them are at least close to a UAV's sensing location.

\begin{table}[tbp]
  \centering\scriptsize
    \begin{tabular}{@{\extracolsep{4pt}}rrccccccccc@{}}
    \toprule
    \multicolumn{1}{c}{\multirow{2}[4]{*}{$|\setTargets|$}} & \multicolumn{1}{c}{\multirow{2}[4]{*}{$|\setVehicles|$}} & \multicolumn{3}{c}{$\textit{PCov}_{0}$}     & \multicolumn{3}{c}{$\textit{PCov}_{100}$} & \multicolumn{3}{c}{$\textit{PCov}_{300}$} \\  \cmidrule{3-5}   
\cmidrule{6-8}    \cmidrule{9-11}   
      &       & TOP   & CorTOP & GCorTOP & TOP   & CorTOP & GCorTOP & TOP   & CorTOP & GCorTOP \\
    \midrule
      & 1     & \textbf{58.1~\%} & 56.2~\% & 46.1~\% & 71.9~\% & 79.9~\% & \textbf{85.2~\%} & 83.5~\% & 89.7~\% & \textbf{95.5~\%} \\
   225       & 2     & \textbf{84.7~\%} & 79.0~\% & 62.9~\% & 92.3~\% & 94.8~\% & \textbf{96.7~\%} & 97.2~\% & 98.4~\% & \textbf{99.7~\%} \\
          & 3     & \textbf{93.4~\%} & 88.9~\% & 75.4~\% & 97.2~\% & 98.4~\% & \textbf{99.1~\%} & 99.4~\% & 99.7~\% & \textbf{100.0~\%} \\
    \midrule
       & 1     & \textbf{42.5~\%} & 40.3~\% & 34.8~\% &   56.8~\% & 63.1~\% & \textbf{71.4~\%} & 73.5~\% & 78.7~\% &  \textbf{87.8~\%} \\
    300      & 2     & \textbf{69.6~\%} & 65.9~\% & 42.7~\% & 83.2~\% & 87.2~\% & \textbf{90.8~\%} & 93.8~\% & 95.6~\% & \textbf{98.3~\%} \\
          & 3     & \textbf{83.5~\%} & 78.5~\% & 60.0~\% & 91.5~\% & 93.9~\% & \textbf{96.3~\%} & 96.8~\% & 98.1~\% & \textbf{99.7~\%} \\
    \midrule
       & 1     & \textbf{42.9~\%} & 41.4~\% & 33.2~\% & 55.9~\% & 61.6~\% & \textbf{70.0~\%} & 68.5~\% & 74.4~\% & \textbf{85.8~\%} \\
    400      & 2     & \textbf{68.4~\%} & 64.8~\% & 46.2~\% & 81.0~\% & 84.4~\% & \textbf{88.6~\%} & 89.3~\% & 91.4~\% & \textbf{94.7~\%} \\
          & 3     & \textbf{81.7~\%} & 76.6~\% & 53.9~\% & 89.2~\% & 91.7~\% & \textbf{93.5~\%} & 93.4~\% & 94.5~\% & \textbf{95.8~\%} \\
    \midrule
       & 1     & \textbf{37.4~\%} & 36.4~\% & 29.0~\% & 49.3~\% & 56.0~\% & \textbf{63.1~\%} & 62.3~\% & 69.1~\% & \textbf{80.0~\%} \\
      500    & 2     & \textbf{60.7~\%} & 57.4~\% & 45.7~\% & 75.3~\% & 79.3~\% & \textbf{84.2~\%} & 86.0~\% & 88.7~\% & \textbf{93.4~\%} \\
          & 3     & \textbf{75.1~\%} & 70.7~\% & 56.6~\% & 85.1~\% & 87.9~\% & \textbf{90.9~\%} & 91.3~\% & 92.6~\% & \textbf{95.1~\%} \\
    \midrule
       & 1     & \textbf{25.8~\%} & 25.0~\% & 19.9~\% & 35.0~\% & 40.2~\% & \textbf{48.4~\%} & 47.6~\% & 53.9~\% & \textbf{67.1~\%} \\
     625     & 2     & \textbf{46.5~\%} & 44.0~\% & 34.9~\% & 63.7~\% & 68.8~\% & \textbf{74.7~\%} & 78.7~\% & 83.7~\% & \textbf{89.4~\%} \\
          & 3     & \textbf{63.1~\%} & 59.8~\% & 43.4~\% & 77.6~\% & 81.7~\% & \textbf{86.4~\%} & 87.5~\% & 90.4~\% & \textbf{93.8~\%} \\
          \midrule
    \multicolumn{2}{c}{all} & \textbf{62.2\%} &    59.0\%&    40.2\%    &73.7\%    &77.9\%    &\textbf{82.6\%}    &83.2\%    &86.6\%    &\textbf{91.7\%} \\
    \bottomrule
    \end{tabular}%
  \caption[$\textit{PCov}_{0}$, $\textit{PCov}_{100}$ and $\textit{PCov}_{300}$ depending on modeling approach]{Average values for $\textit{PCov}_{0}$, $\textit{PCov}_{100}$ and $\textit{PCov}_{300}$ depending on target area size, number of vehicles, and modeling approach. Best values for each set are indicated in bold.}
  \label{table:pcov_comparison}%
\end{table}%

\paragraph{Prediction quality}
We now investigate the prediction quality that can be achieved using the samples taken during the UAV missions.
To this end, we apply GP regression approaches (see Section \ref{sec:spatial_models}, which are computationally too expensive for evaluating interim solutions during the search.  
The planned UAV mission determines the set of sampling locations $\setSelected$.
The observed values $Z(i)$ for all $i \in \setSelected$ are taken from the simulated spatial distribution of contaminants.
We condition a Gaussian process on these samples.
For fitting the covariance function of this GP to the sample data, we use the Python \texttt{scikit-learn} package \citep{Pedregosa2011}.
The GP posterior then yields predicted values $\widehat{Z}(i)$ for all $i \in \setTargets$.
To measure the quality of this prediction, we determine the deviation of the predicted values from the ``true'' distribution of contaminants, i.e., the initially generated distribution.
In this study, we use three measures:
The mean absolute error (MAE)
\begin{align}
\text{MAE}(\setSelected) = &  \frac{1}{|\setTargets|}\sum_{i \in \setTargets} |\widehat{Z}(i) - Z(i)|\label{f:mae}
\end{align}
allows to compare overall accuracy of the predictions.
The mean error (ME), which is defined as 
\begin{align}
\text{ME} = &  \frac{1}{|\setTargets|}\sum_{i \in \setTargets} (\widehat{Z}(i) - Z(i)),\label{f:me}
\end{align}
indicates whether or not the selected samples lead to a systematic deviation in the prediction, e.g., a systematic underestimation or overestimation of the contamination.
Finally, we also determine the weighted mean absolute error (WMAE)
\begin{align}
\text{WMAE} = &  \frac{1}{\sum_{i \in \setTargets} u_i}\sum_{i \in \setTargets} u_i \cdot |\widehat{Z}(i) - Z(i)|\label{f:wmae}
\end{align}
which also accounts for priorities. 
This way, we can evaluate the trade-off between overall prediction quality and the accuracy at highly prioritized locations.

Table \ref{table:error_comparison} gives the results with respect to all three measures.
We can see that with respect to MAE, the GCorTOP model invariably achieves the lowest values, i.e., it yields the most accurate predictions.
Compared to TOP, MAE is reduced by approximately $43~\%$ on average.
Even compared to the CorTOP model, which already incorporates aspects of spatial coverage, MAE is reduced by $33~\%$.
The prediction bias (ME) is within a reasonable range for all models, and largely negligible for the GCorTOP when several UAVs are available.

\begin{table}[tbp]
  \centering\scriptsize
    \begin{tabular}{@{\extracolsep{4pt}}rrrrrrrrrrr@{}}
    \toprule
   \multicolumn{1}{c}{\multirow{2}[4]{*}{$|\setTargets|$}} & \multicolumn{1}{c}{\multirow{2}[4]{*}{$|\setVehicles|$}} & \multicolumn{3}{c}{MAE} & \multicolumn{3}{c}{ME} & \multicolumn{3}{c}{WMAE} \\\cmidrule{3-5}   
\cmidrule{6-8}    \cmidrule{9-11}   
       &       & \multicolumn{1}{l}{TOP} & \multicolumn{1}{l}{CorTOP} & \multicolumn{1}{l}{GCorTOP} & \multicolumn{1}{l}{TOP} & \multicolumn{1}{l}{CorTOP} & \multicolumn{1}{l}{GCorTOP} & \multicolumn{1}{l}{TOP} & \multicolumn{1}{l}{CorTOP} & \multicolumn{1}{l}{GCorTOP} \\
    \midrule
    \multirow{3}[2]{*}{225} & 1  & 10.67  & 8.41  & \textbf{6.00} & 0.97  & 0.82  & \textbf{0.43} & 5.24  & \textbf{3.95} & 4.64 \\
          & 2  & 5.45  & 3.85  & \textbf{2.06} & 0.53  & 0.48  & \textbf{-0.03} & 1.40  & 1.87  & \textbf{1.19} \\
          & 3  & 3.12  & 1.83  & \textbf{1.16} & 0.49  & 0.36  & \textbf{-0.07} & 0.47  & \textbf{0.44} & 0.90 \\
    \midrule
    \multirow{3}[2]{*}{300} & 1  & 9.62  & 9.07  & \textbf{6.14} & 1.12  & 0.62  & \textbf{0.56} & 7.72  & 6.00  & \textbf{5.91} \\
          & 2  & 6.37  & 4.81  & \textbf{3.84} & -0.14 & 0.21  & \textbf{0.07} & 2.42  & \textbf{2.09} & 2.69 \\
          & 3  & 4.17  & 2.71  & \textbf{2.17} & -0.44 & -0.07 & \textbf{-0.02} & 1.16  & \textbf{0.96} & 1.46 \\
    \midrule
    \multirow{3}[2]{*}{400} & 1  & 12.56 & 11.22 & \textbf{7.83} & \textbf{1.15} & 1.36  & 1.41  & 6.37  & 5.69  & \textbf{4.92} \\
          & 2  & 6.83  & 6.44  & \textbf{3.80} & 0.34  & 0.23  & \textbf{0.18} & 2.24   & 2.10 & \textbf{1.93}\\
          & 3  & 4.26  & 3.71  & \textbf{2.07} & 0.27  & 0.19  & \textbf{-0.03} & 1.04  & 1.24 & \textbf{0.88} \\
    \midrule
    \multirow{3}[2]{*}{500} & 1  & 11.56 & 10.91 & \textbf{8.57} & 1.64  & 1.38  & \textbf{0.83} & 6.46  & 5.53  & \textbf{4.34} \\
          & 2  & 6.55  & 6.14  & \textbf{3.41} & 0.46  & 0.44  & \textbf{0.32} & 2.20  & 1.93  & \textbf{1.90} \\
          & 3  & 4.58  & 4.31  & \textbf{2.49} & 0.44  & 0.47  & \textbf{0.18} & 1.16  & \textbf{1.06} & 1.08 \\
    \midrule
    \multirow{3}[2]{*}{625} & 1  & 13.53 & 12.64 & \textbf{9.28} & \textbf{0.33} & 0.60  & 0.80  & 9.52  & 8.23  & \textbf{5.69} \\
          & 2  & 7.23  & 6.49  & \textbf{5.17} & \textbf{1.27} & 1.31  & 1.36  & 3.28  & 2.56  & \textbf{2.27} \\
          & 3  & 4.98  & 4.42  & \textbf{2.88} & 1.27  & 1.14  & \textbf{0.32} & 1.56  & \textbf{1.41} & 1.22 \\
     \midrule
              \multicolumn{2}{c}{all} & 7.03    &6.46    &\textbf{5.09}    &0.65    &0.64    &\textbf{0.49}    &3.42    &2.96    &\textbf{2.80}
 \\
    \bottomrule
    \end{tabular}%
  \caption[Prediction quality depending on modeling approach]{Average prediction quality measures depending on target area size, number of vehicles, and modeling approach. Best values for each set are indicated in bold.}
  \label{table:error_comparison}%
\end{table}%

The WMAE measure is less than MAE for all models.
This is a consequence of the integration of priorities, as the focus on samples at prioritized locations ensures that predictions are comparatively good in the corresponding areas.
Despite this, TOP does not yield the best WMAE.
This is due to two reasons:
First, the overall reduction in MAE that is achieved by incorporating correlations also helps to reduce WMAE.
Second, routes planned using TOP are often very narrow, leaving out some areas with high priorities that do not justify a detour when coverage is not considered.
This is also demonstrated in the visual comparison in the final part of this section.

From a practical point of view, these are promising results.
While predictions are by no means flawless, for most areas of the target region, predicted values are at least close to the true distribution.
As the WMAE indicates, high prediction accuracy can be achieved at least for crucial areas, even given comparatively limited flight times and large target areas. 

\subsubsection{Visual comparison}

\begin{figure}[tbp]
    \centering
    \captionsetup[subfigure]{justification=centering}
    \begin{subfigure}[b]{0.45\textwidth}
        \includegraphics[width=\textwidth]{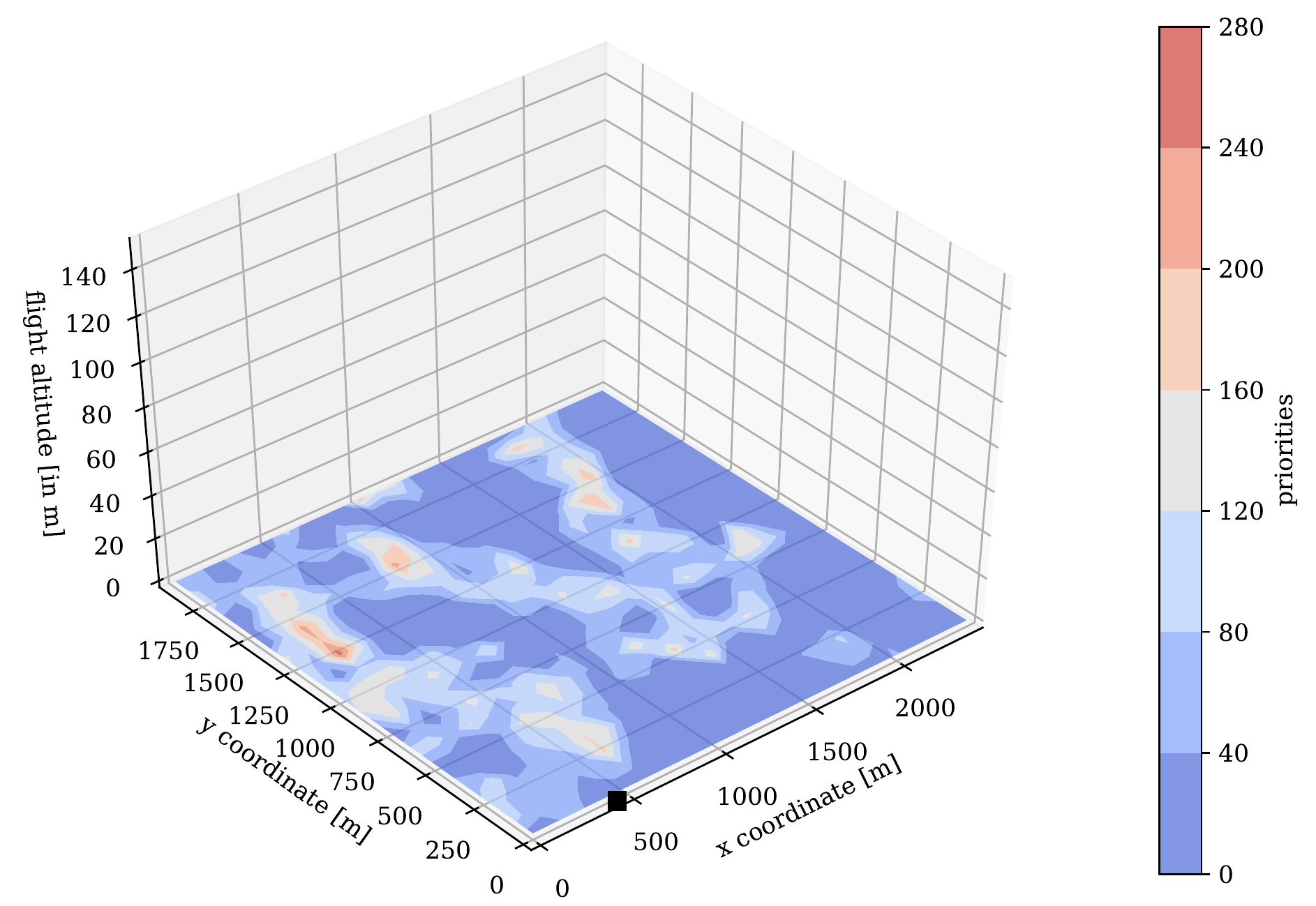}
        \caption{Priorities in the target area}
        \label{fig:sol_comp_ex_1_priorities}
    \end{subfigure}
    \begin{subfigure}[b]{0.45\textwidth}
        \includegraphics[width=\textwidth]{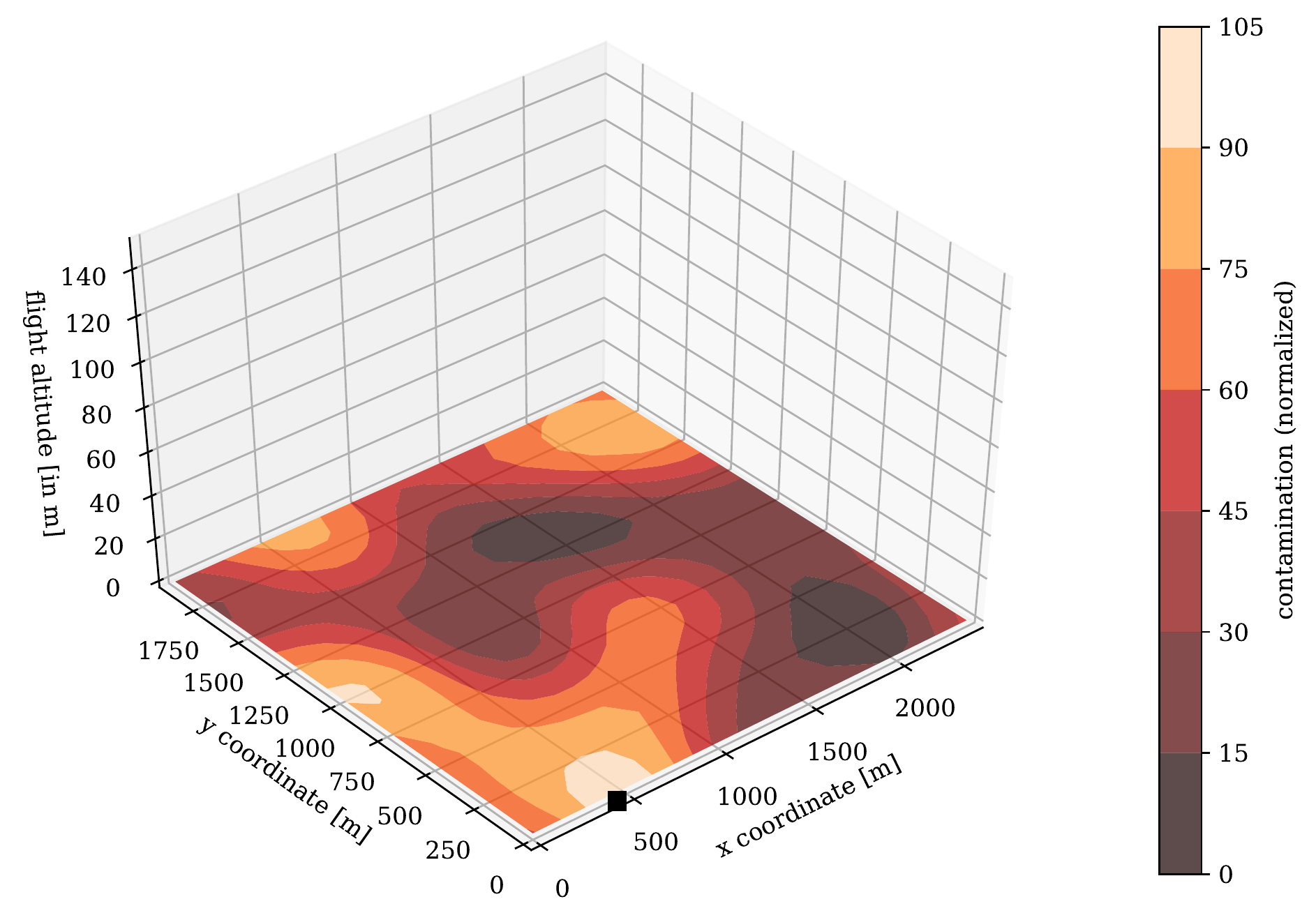}
        \caption{Simulated distribution of gases}
        \label{fig:sol_comp_ex_1_distribution}
    \end{subfigure}
    \caption{Base data of the benchmark instance for example 1}\label{fig:sol_comp_ex_1}
\end{figure}

In a last step, we give a visual indication of the impact that considering coverage aspects has on the solutions.
To this end, we compare results for two instances.
The first example is depicted in Figure \ref{fig:sol_comp_ex_1}.
We indicate priorities on the left-hand side and the simulated ``true'' distribution of contaminants on the right.

\begin{figure}[!tbp]
    \centering
    \captionsetup[subfigure]{justification=centering}
    \begin{subfigure}[b]{0.45\textwidth}
        \includegraphics[width=\textwidth]{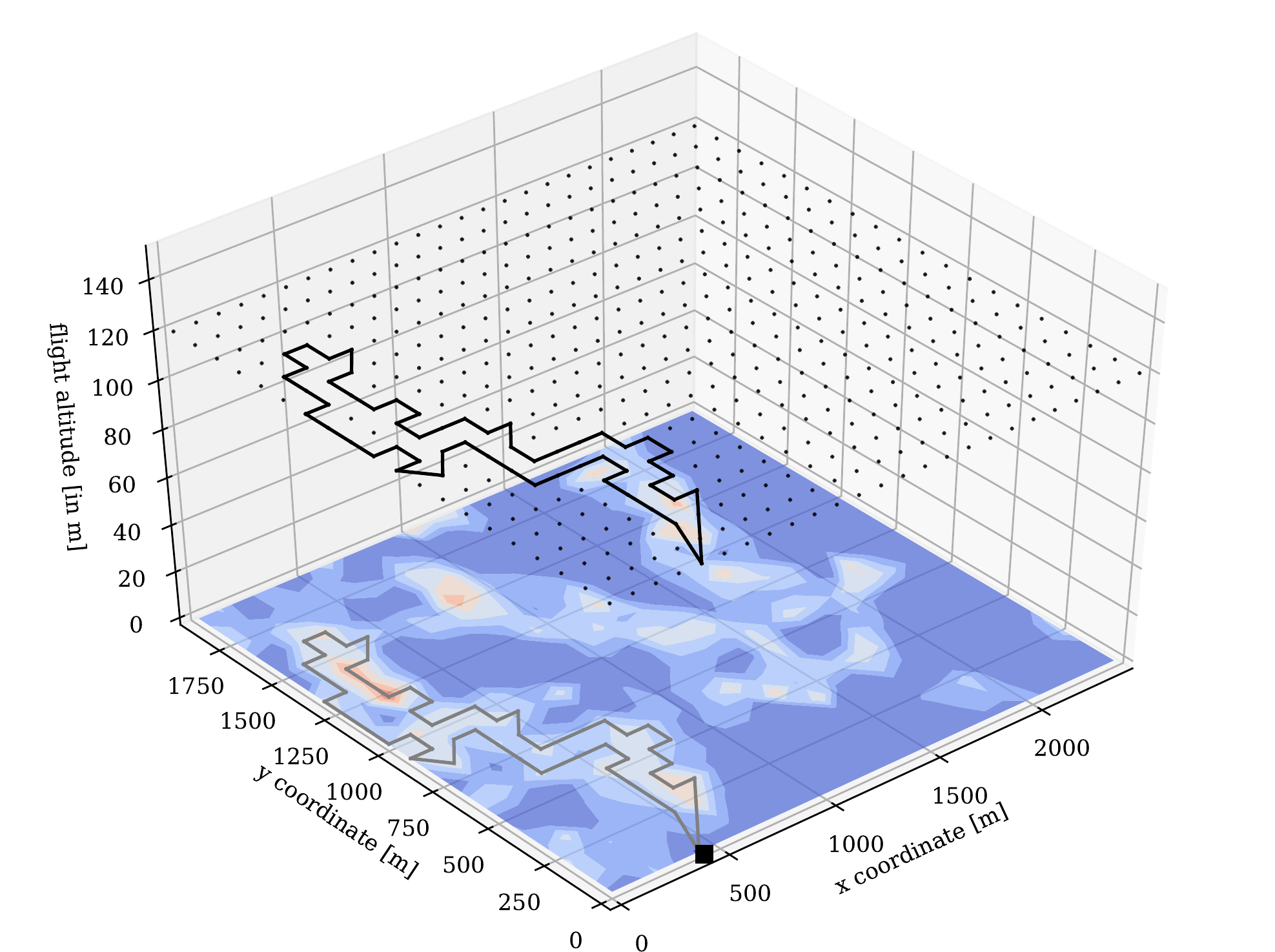}
        \caption{Planned mission}
        \label{fig:sol_comp_ex_1_top_plan}
    \end{subfigure}
        \begin{subfigure}[b]{0.45\textwidth}
        \includegraphics[width=\textwidth]{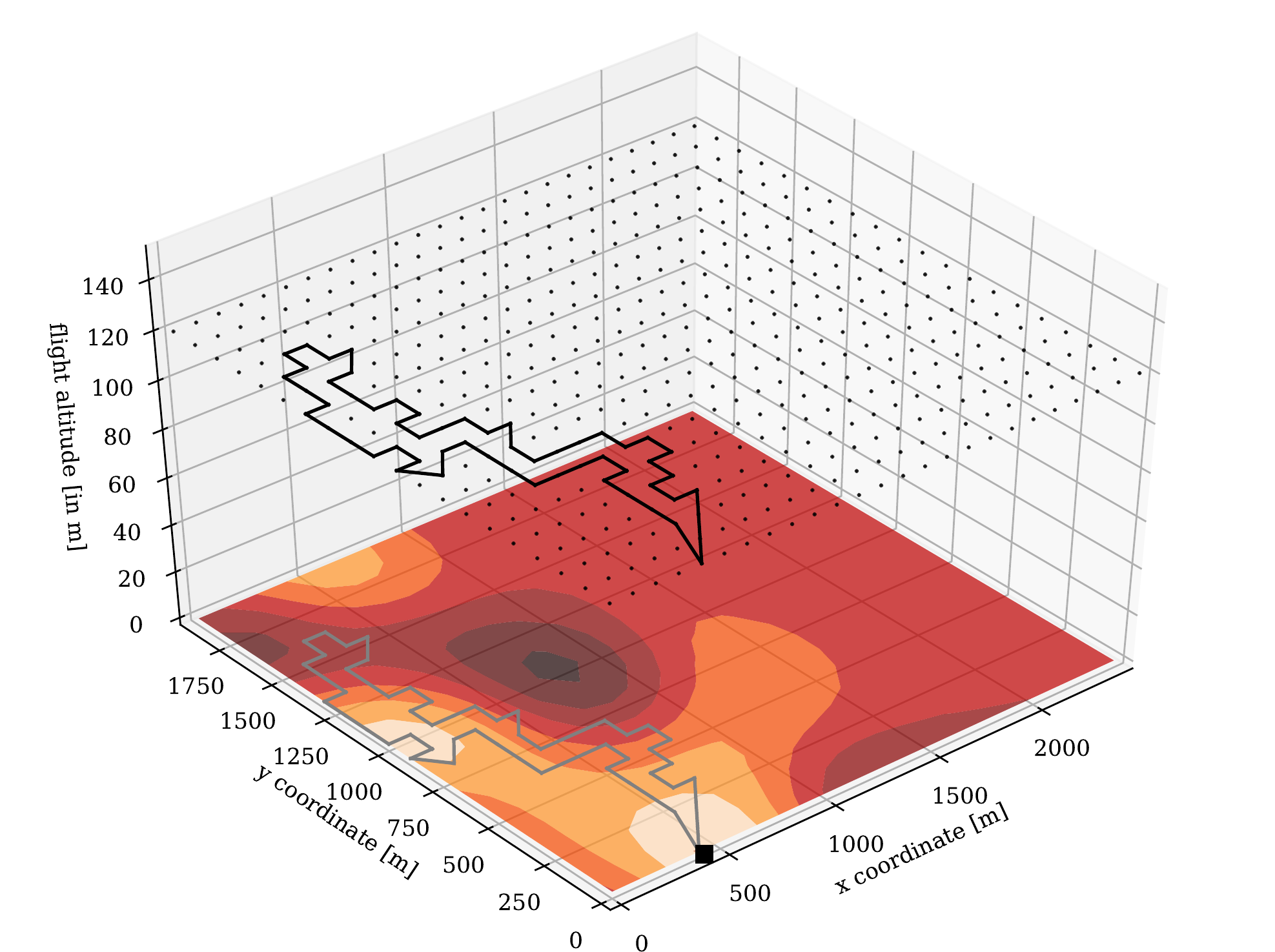}
        \caption{Predicted distribution}
        \label{fig:sol_comp_ex_1_top_pred}
    \end{subfigure}
    \caption{TOP results for example 1}\label{fig:sol_comp_ex_1_top}
        \vspace{0.5cm}
    \begin{subfigure}[b]{0.45\textwidth}
        \includegraphics[width=\textwidth]{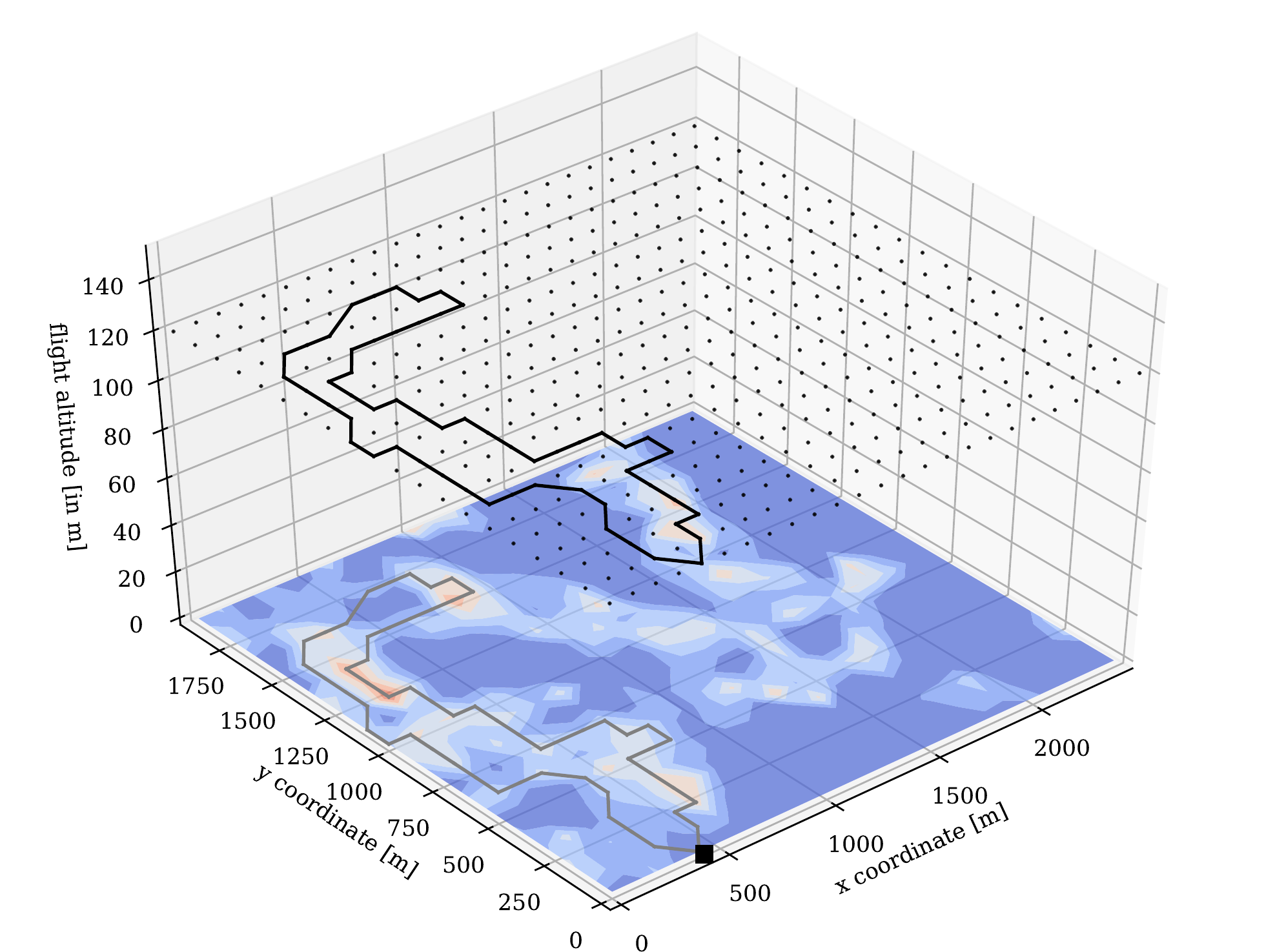}
        \caption{Planned mission}
        \label{fig:sol_comp_ex_1_cortop_plan}
    \end{subfigure}
        \begin{subfigure}[b]{0.45\textwidth}
        \includegraphics[width=\textwidth]{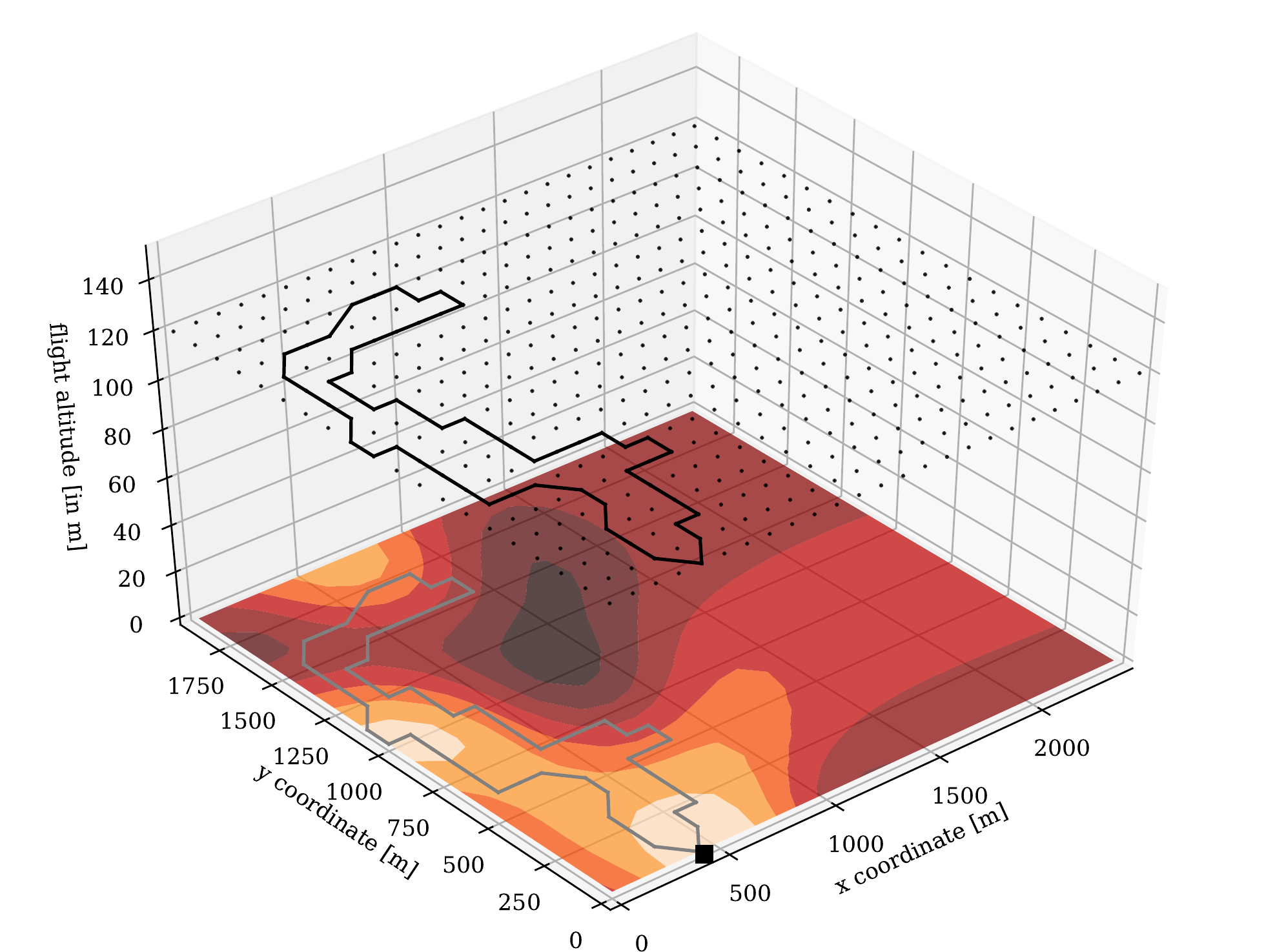}
        \caption{Predicted distribution}
        \label{fig:sol_comp_ex_1_cortop_pred}
    \end{subfigure}
    \caption{CorTOP results for example 1}\label{fig:sol_comp_ex_1_cortop}
    \vspace{0.5cm}
    \begin{subfigure}[b]{0.45\textwidth}
        \includegraphics[width=\textwidth]{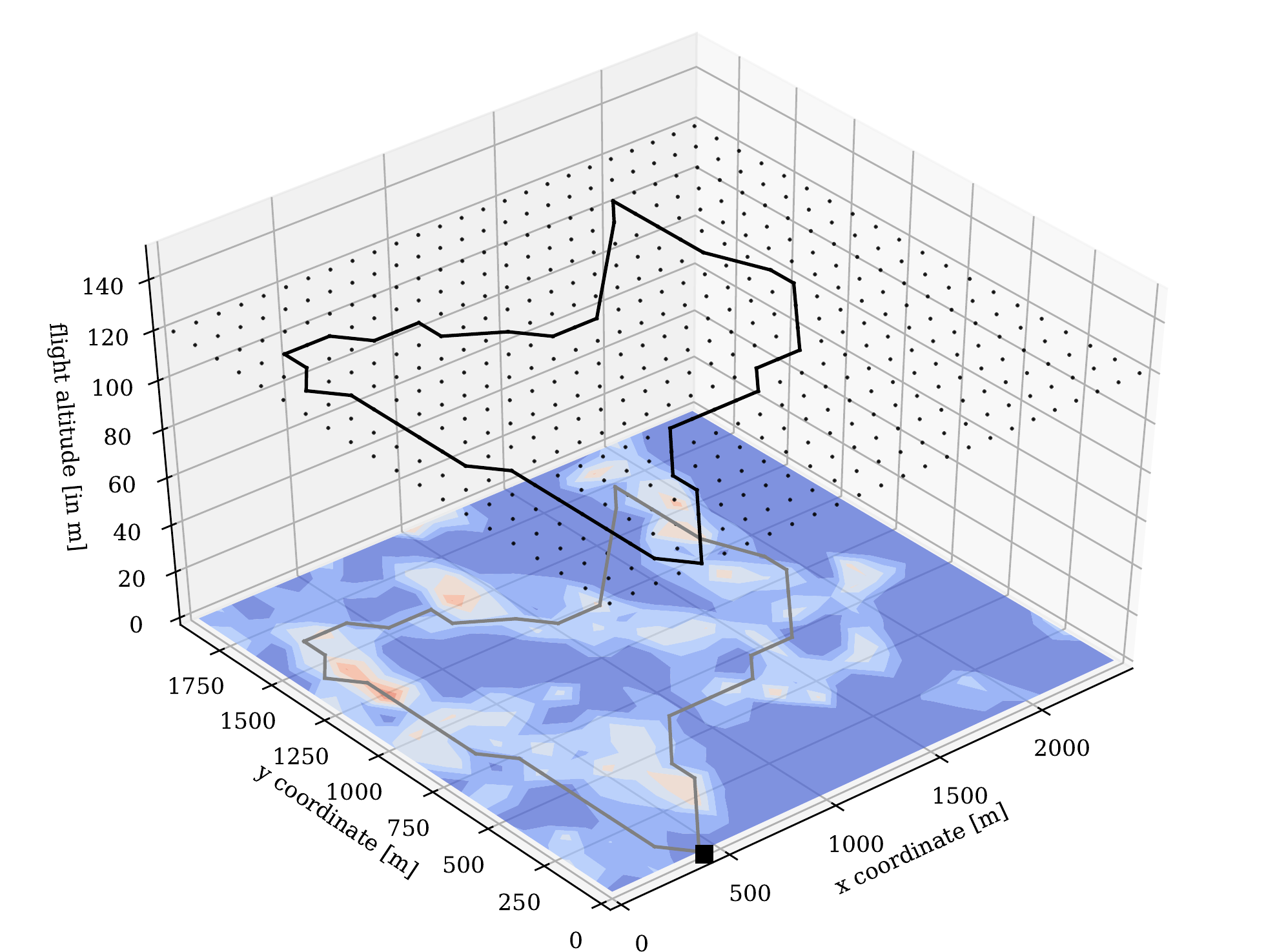}
        \caption{Planned mission}
        \label{fig:sol_comp_ex_1_gcortop_plan}
    \end{subfigure}
        \begin{subfigure}[b]{0.45\textwidth}
        \includegraphics[width=\textwidth]{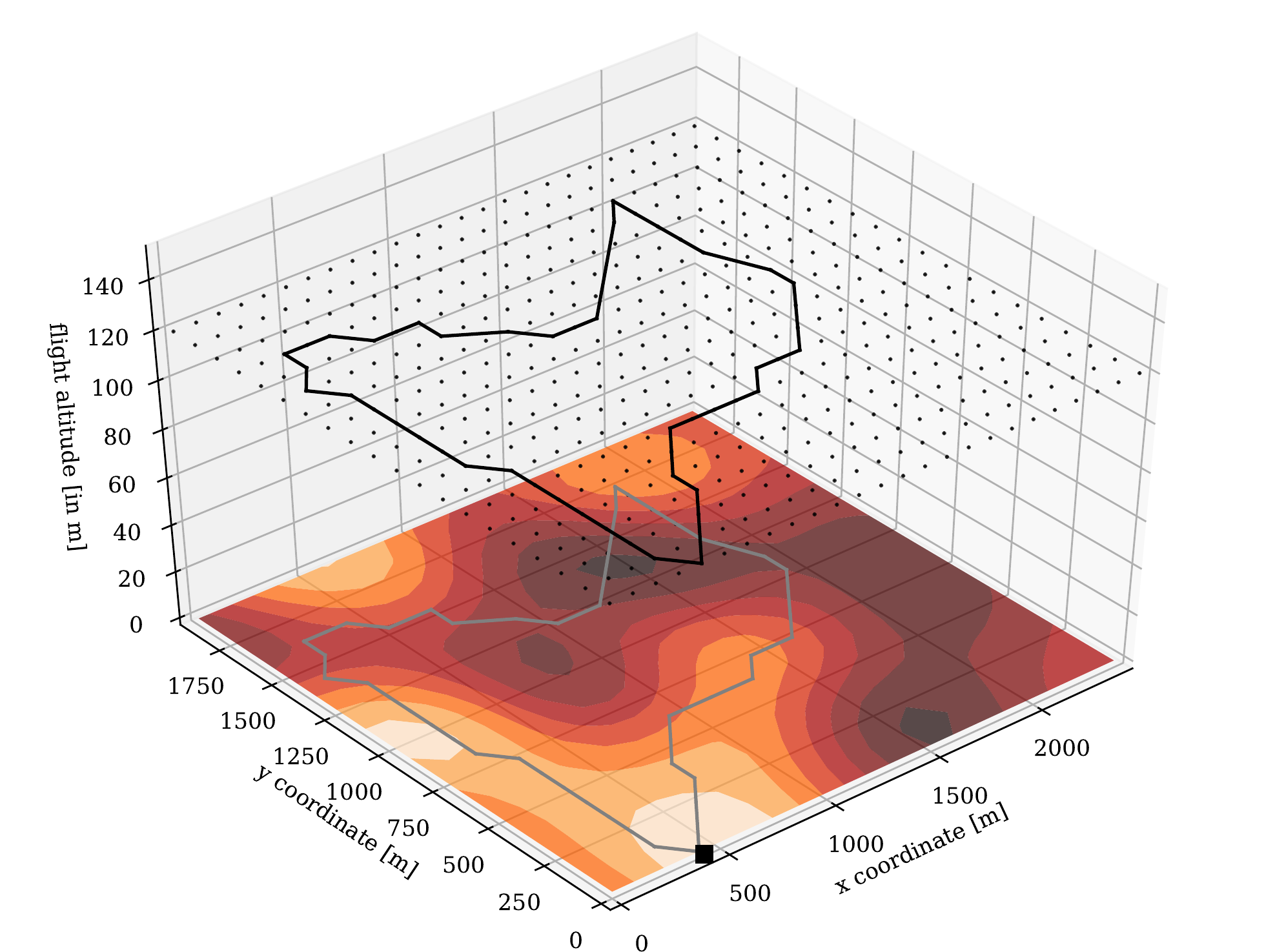}
        \caption{Predicted distribution}
        \label{fig:sol_comp_ex_1_gcortop_pred}
    \end{subfigure}
    \caption{GCorTOP results for example 1}\label{fig:sol_comp_ex_1_gcortop}
\end{figure}

Missions obtained using the three modeling variants are given in Figures \ref{fig:sol_comp_ex_1_top} to \ref{fig:sol_comp_ex_1_gcortop}.
These routes are planned for a single UAV with a maximum flight time of $T^{\max} = 1500$.
Flights start and end at position $(0, 400)$.
For each solution, we indicate the planned mission on the left-hand side, superposed on the priorities that are the basis for this plan.
On the right-hand side, we give the predicted distribution.
Note that the grey areas indicate that the associated predicted value is equal or close to the mean value of the samples that are taken.
Predictions converge to this value as the distance to the next sampling location increases and no local information that can indicate a deviation from the mean is available.

The TOP results in the narrow route given in Figure \ref{fig:sol_comp_ex_1_top}.
The UAV traverses a region with high utilities but leaves major parts of the target region entirely unexplored.
The mission planned using the CorTOP (Figure \ref{fig:sol_comp_ex_1_cortop}) is broader in comparison with the TOP, but follows a similar trajectory.
Finally, the GCorTOP determines a mission that differs strongly from the two other models.
The UAV travels further to gather more information at the regions with higher priorities that are indicated in the right part of the figure.
This significantly improves prediction quality.

\begin{figure}[tbp]
    \centering
    \captionsetup[subfigure]{justification=centering}
    \begin{subfigure}[b]{0.45\textwidth}
        \includegraphics[width=\textwidth]{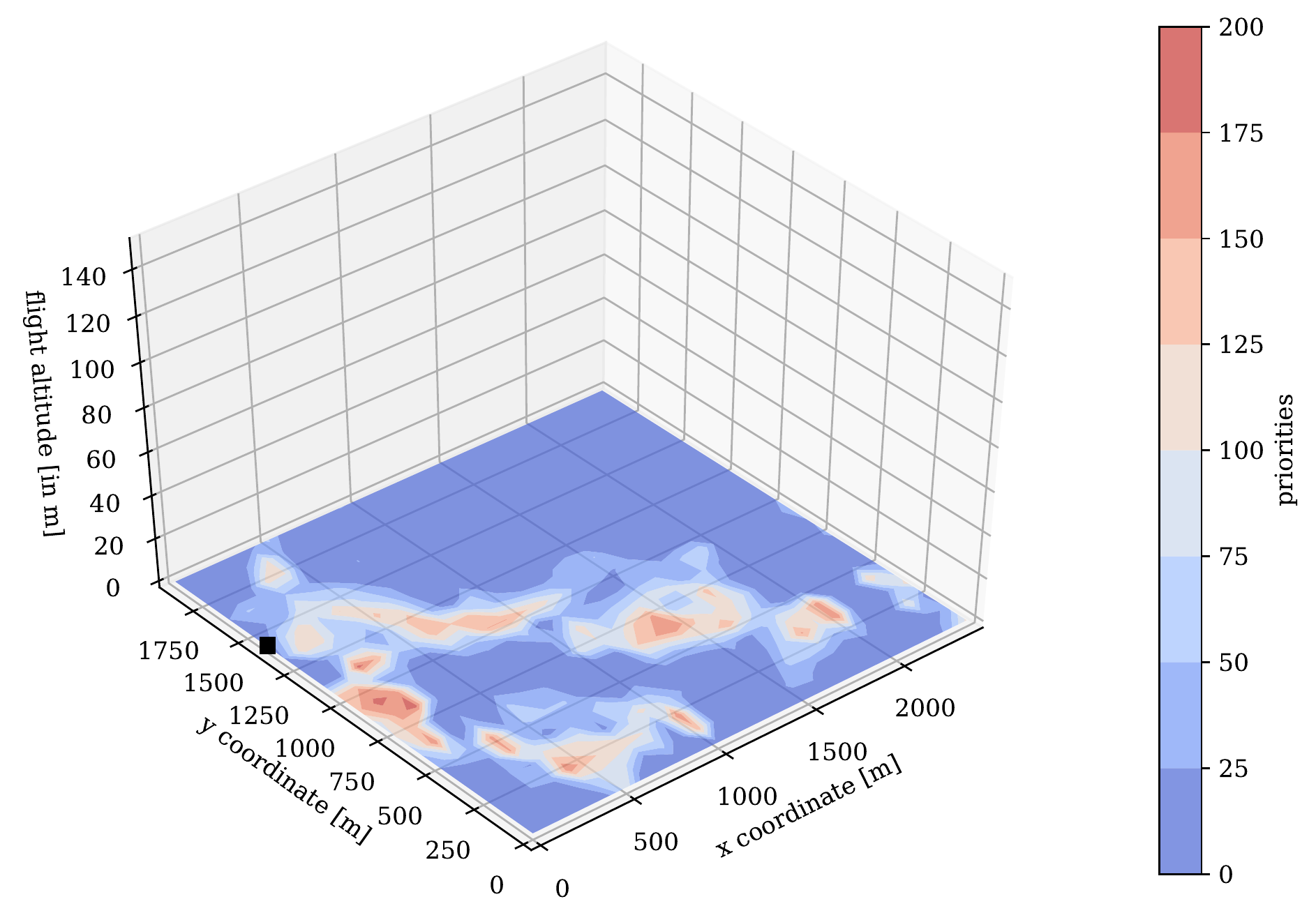}
        \caption{Priorities in the target area}
        \label{fig:sol_comp_ex_2_priorities}
    \end{subfigure}
    \begin{subfigure}[b]{0.45\textwidth}
        \includegraphics[width=\textwidth]{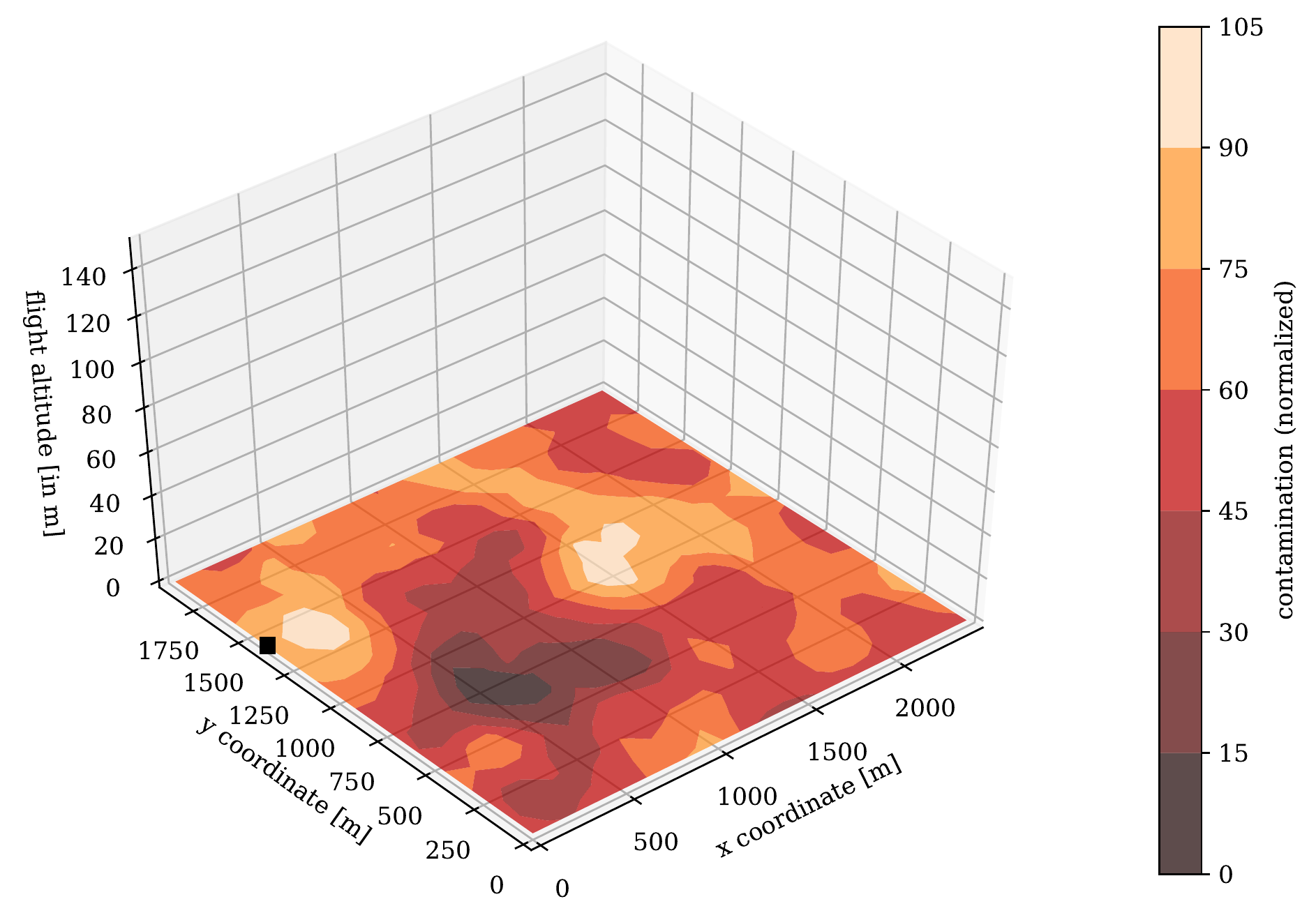}
        \caption{Simulated distribution of gases}
        \label{fig:sol_comp_ex_2_distribution}
    \end{subfigure}
    \caption{Base data of the benchmark instance for example 2}\label{fig:sol_comp_ex_2}
\end{figure}

The second instance is indicated in Figure \ref{fig:sol_comp_ex_2}.
Compared to the previous example, we can see that the simulated distribution of values is more coarse, i.e., it changes more quickly with distance, and is therefore more challenging to predict.
UAV missions are planned for one vehicle with a flight duration limit of $T^{\max} = 1200$~s, departing from position (0, 1400).

\begin{figure}[!tbp]
    \centering
    \captionsetup[subfigure]{justification=centering}
    \begin{subfigure}[b]{0.45\textwidth}
        \includegraphics[width=\textwidth]{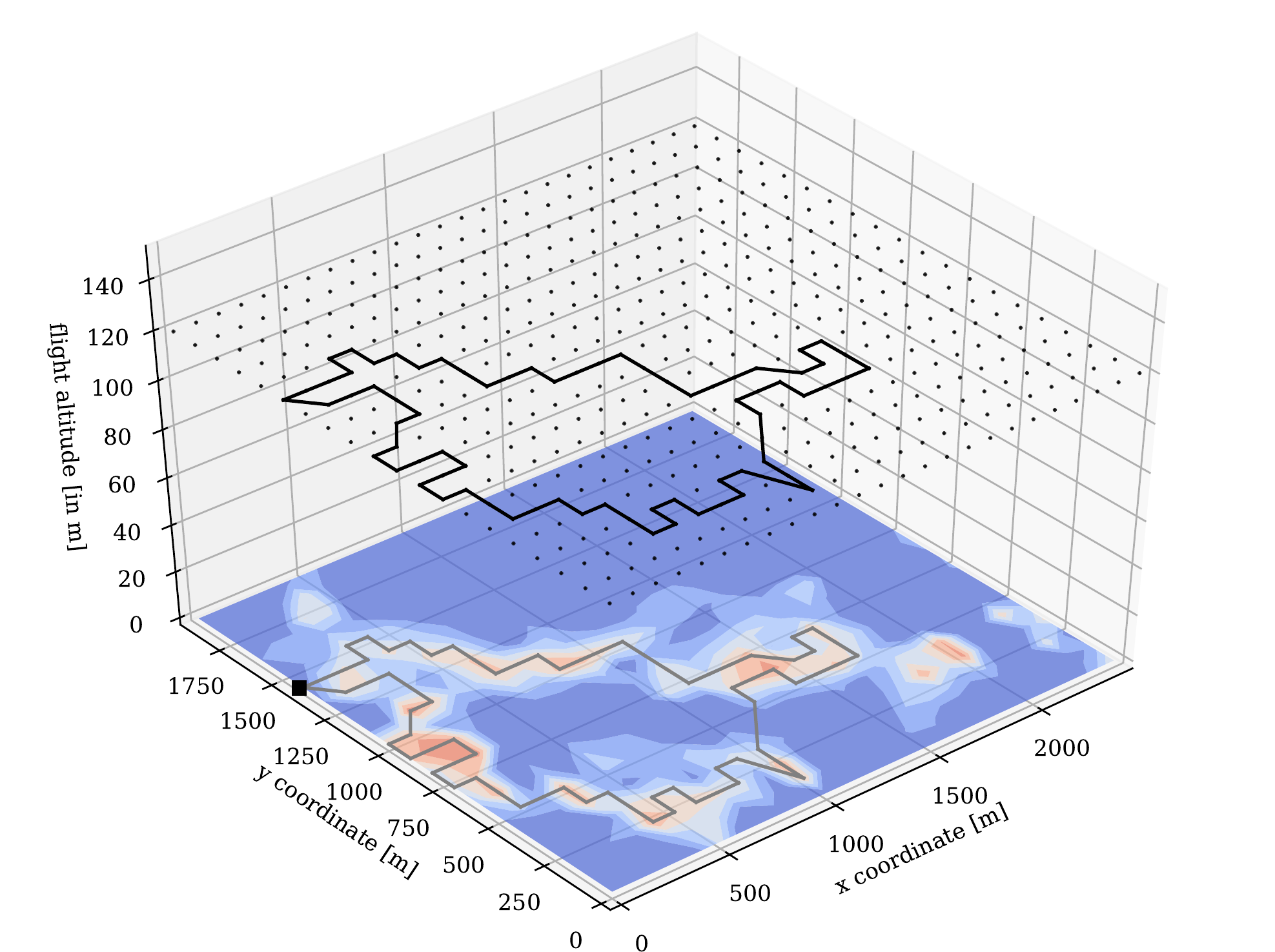}
        \caption{Planned mission}
        \label{fig:sol_comp_ex_2_top_plan}
    \end{subfigure}
        \begin{subfigure}[b]{0.45\textwidth}
        \includegraphics[width=\textwidth]{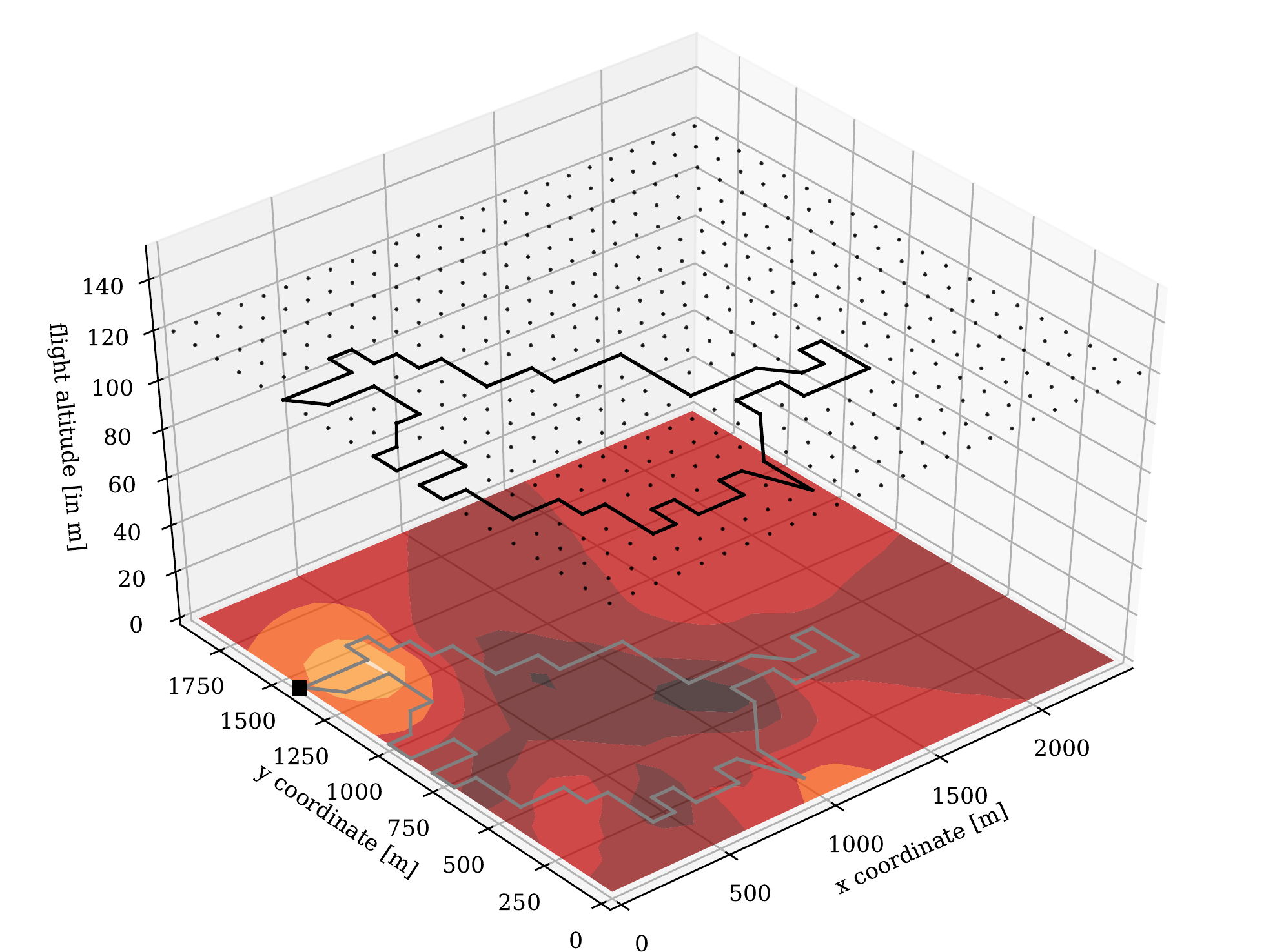}
        \caption{Predicted distribution}
        \label{fig:sol_comp_ex_2_top_pred}
    \end{subfigure}
    \caption{TOP results for example 2}\label{fig:sol_comp_ex_2_top}    \vspace{0.5cm}
    \begin{subfigure}[b]{0.45\textwidth}
        \includegraphics[width=\textwidth]{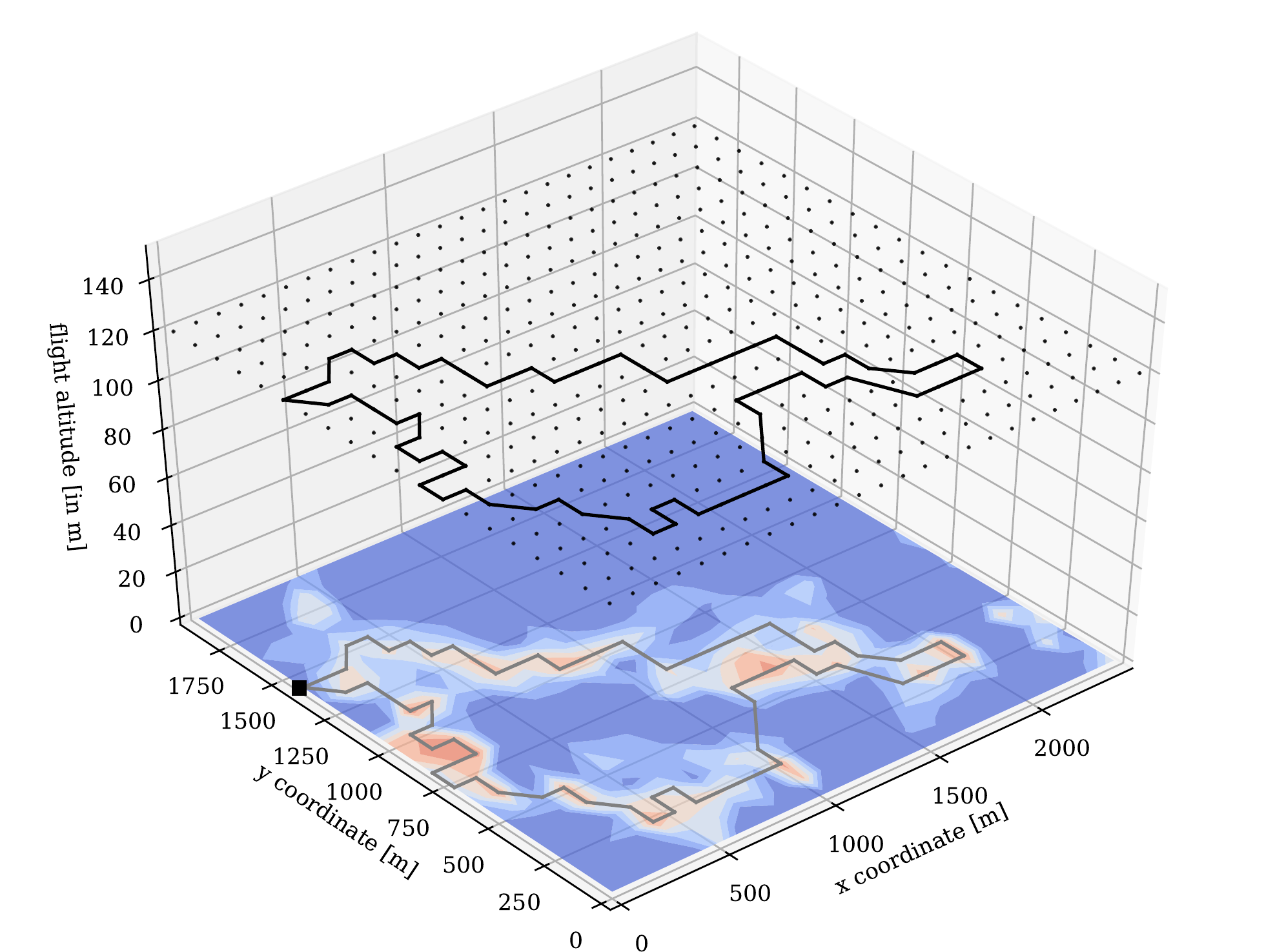}
        \caption{Planned mission}
        \label{fig:sol_comp_ex_2_cortop_plan}
    \end{subfigure}
        \begin{subfigure}[b]{0.45\textwidth}
        \includegraphics[width=\textwidth]{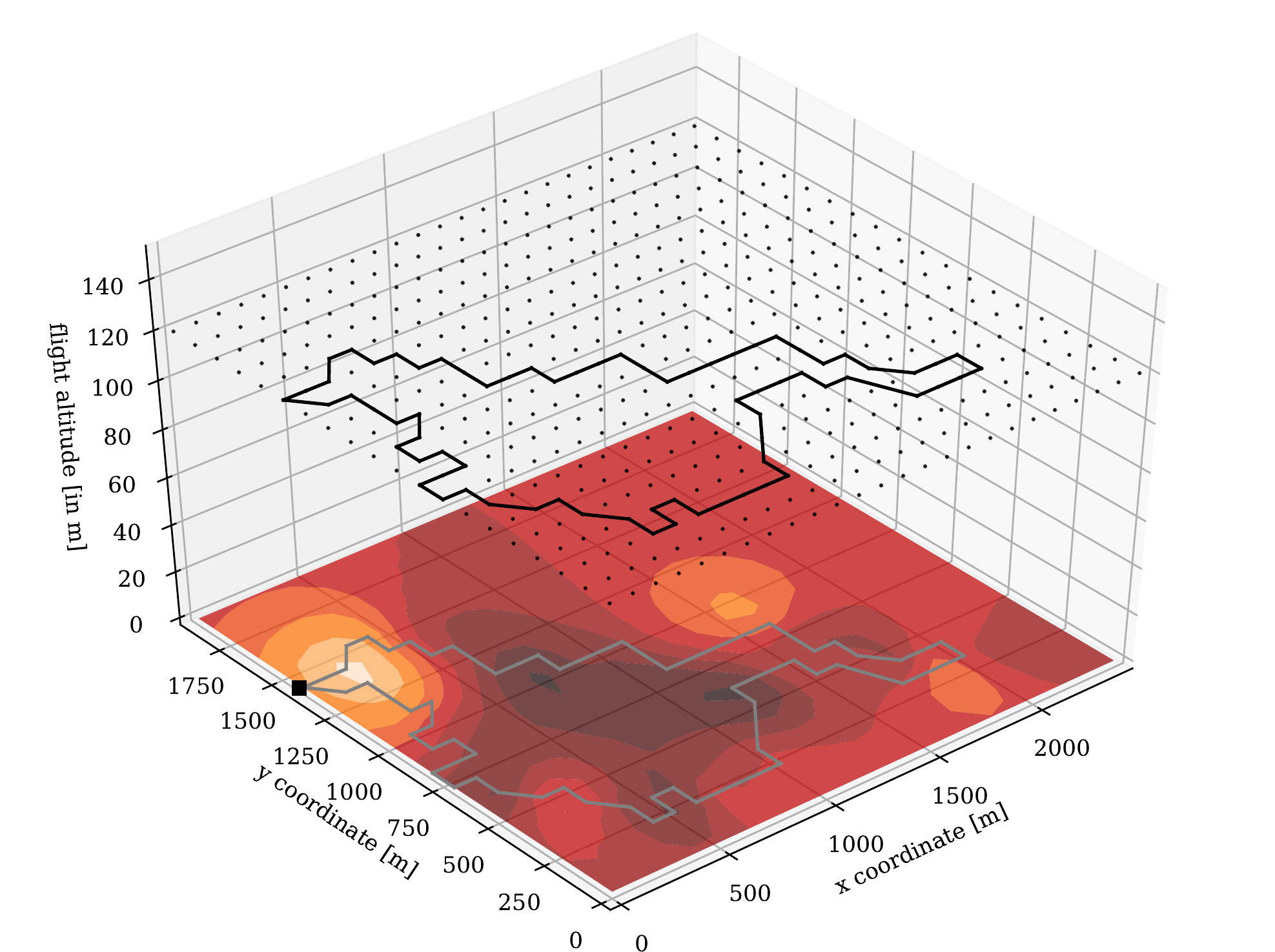}
        \caption{Predicted distribution}
        \label{fig:sol_comp_ex_2_cortop_pred}
    \end{subfigure}
    \caption{CorTOP results for example 2}\label{fig:sol_comp_ex_2_cortop}    \vspace{0.5cm}
    \begin{subfigure}[b]{0.45\textwidth}
        \includegraphics[width=\textwidth]{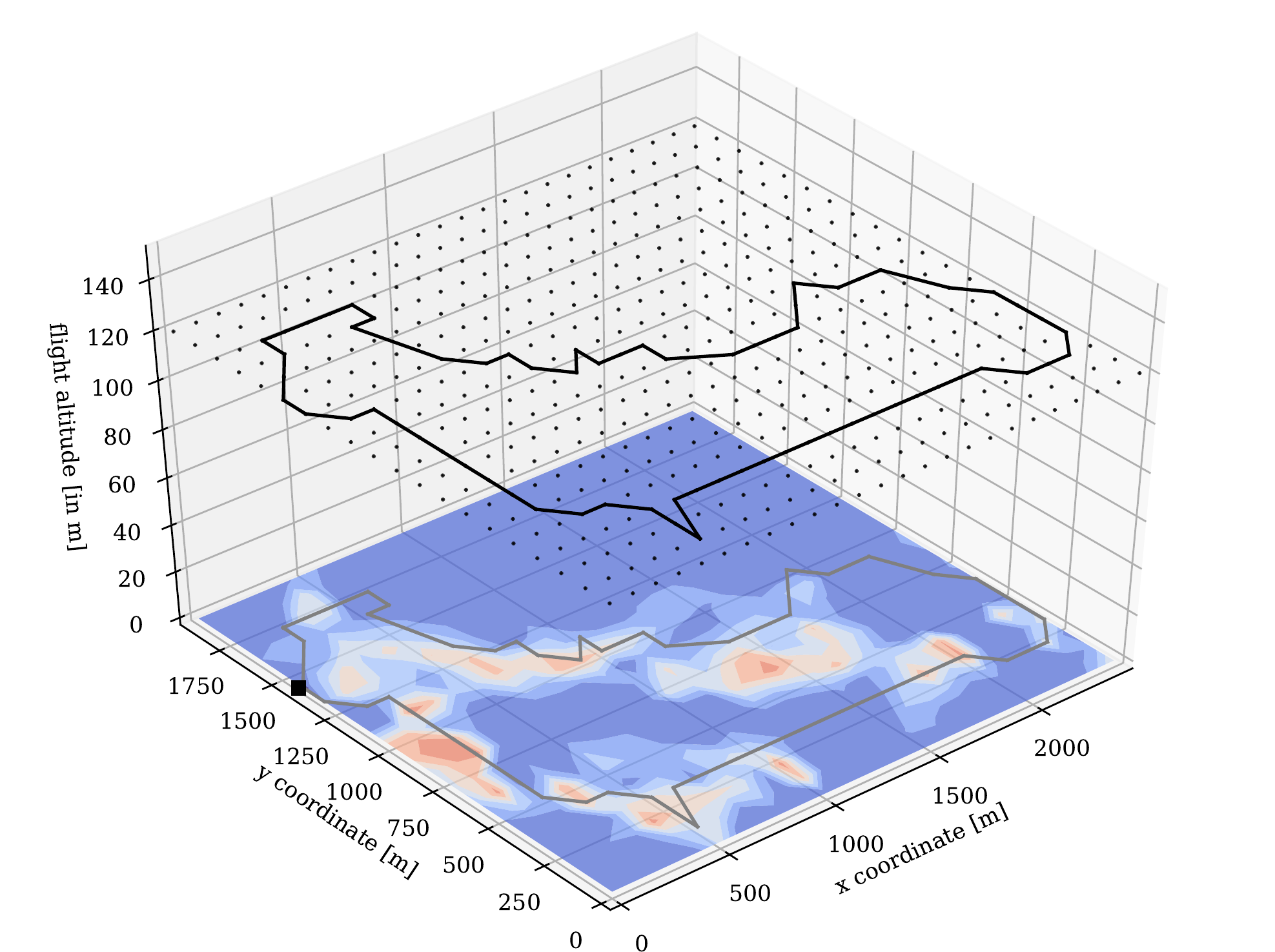}
        \caption{Planned mission}
        \label{fig:sol_comp_ex_2_gcortop_plan}
    \end{subfigure}
        \begin{subfigure}[b]{0.45\textwidth}
        \includegraphics[width=\textwidth]{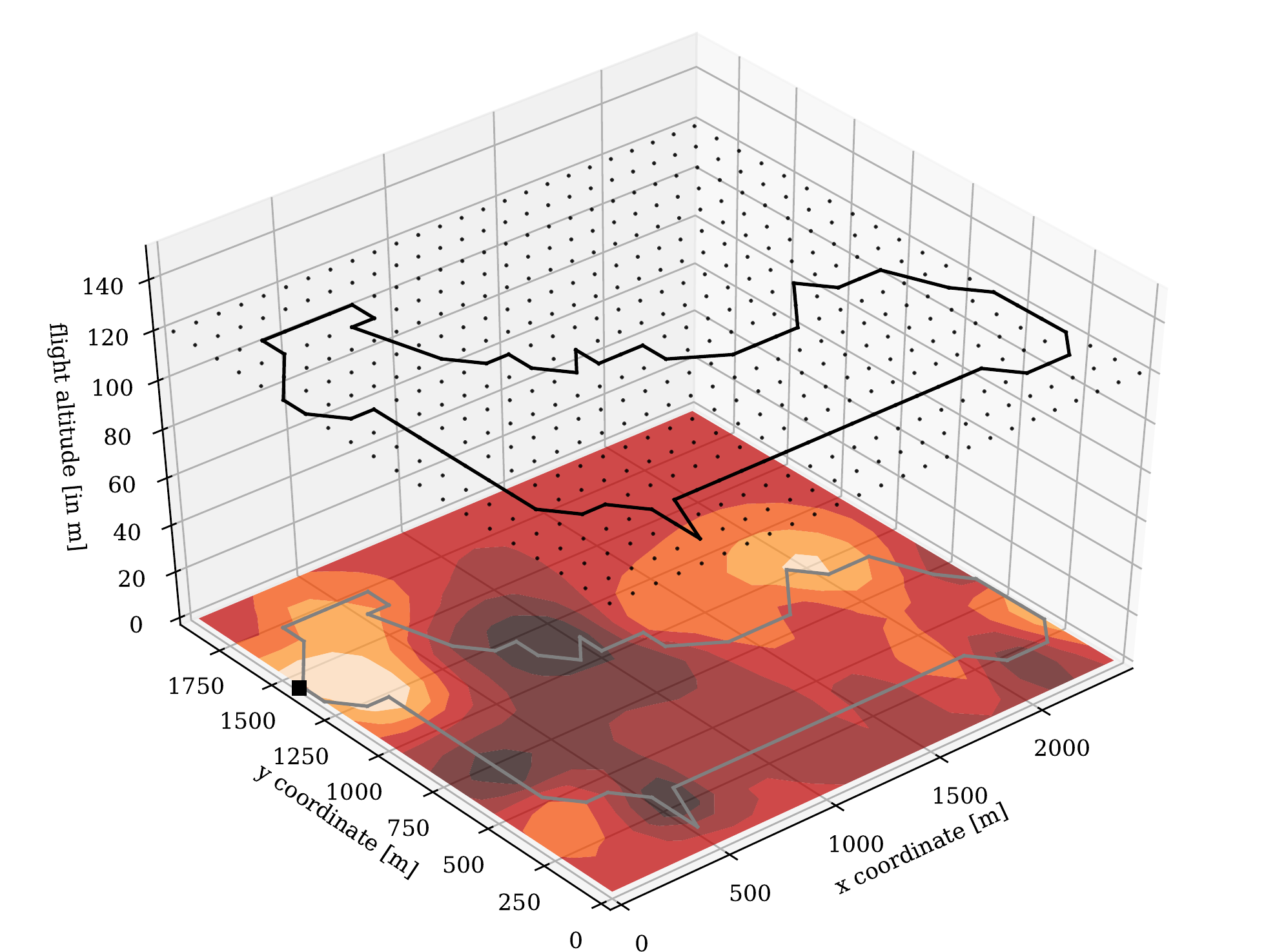}
        \caption{Predicted distribution}
        \label{fig:sol_comp_ex_2_gcortop_pred}
    \end{subfigure}
    \caption{GCorTOP results for example 2}\label{fig:sol_comp_ex_2_gcortop}
\end{figure}

The TOP solution in Figure \ref{fig:sol_comp_ex_2_top} clearly mirrors the shape determined by the priorities and successfully covers almost all of the highly prioritized target locations.
Even though, the prediction is comparatively imprecise due to the coarse spatial distribution, which means that prediction quality deteriorates quickly even in short distance to a sampling location.
The CorTOP model leads to a very similar result, both in terms of vehicle trajectory and prediction quality.
The GCorTOP further emphasizes spatial interdependencies, yielding a much broader route compared to the two other solutions (Figure \ref{fig:sol_comp_ex_2_gcortop}).
While prediction quality still suffers due to the coarse distribution, this is the only model to indicate higher contamination values at the right-hand side of the target area.
These are severely underestimated using the two other modeling approaches.

\section{Conclusion}\label{sec:conclusion}

This paper presents a novel approach for modeling spatial correlations when planning informative missions for UAVs.
These missions include a number of sampling locations, based on which we can predict the distribution of these contaminants across the affected area.
We formalize the mission planning problem for emergency surveillance (MPPES) and introduce a new model that considers both the information on spatial interdependencies as well as priorities within the target region in order to provide accurate data at locations that are particularly relevant for the emergency services.
For solving this planning problem, we propose a two-phase multi-start adaptive large neighborhood search.
In order to quickly determine good solutions, Phase 1 focuses on exploration by planning missions based on reduced problem representations.
In Phase 2, the missions are improved further using the complete problem instance.

We demonstrate the applicability of the proposed model and scalability of our solution approach in an extensive study based on MPPES benchmark instances derived from a real-world use case.
We furthermore propose several evaluation criteria specifically addressing the needs of this use case.
Using MPPES instances and performance measures, we can show that the proposed model can adequately capture spatial correlations.
The obtained routes are wider and explore larger areas, leading to major improvements in prediction quality. 
From a practical point of view, we can plan UAV missions that allow identifying affected areas in case of emergency, even in the face of large target areas of multiple square kilometers.
In future work, we investigate the MPPES in a dynamic setting, in which factors such as wind vary over time. 
This requires progress in two areas: 
Firstly, we need to incorporate process models that account for temporal interdependencies. 
This means that observations lose value over time, and additional measurements may have to be taken at or close to locations that have been surveyed earlier. 
Secondly, we derive adaptive methods that can incorporate this information to adjust the UAV paths based on updated process models while the mission is in progress.

%% file: main.bbl
\begin{thebibliography}{48}
\providecommand{\natexlab}[1]{#1}
\providecommand{\url}[1]{\texttt{#1}}
\providecommand{\urlprefix}{URL }

\bibitem[{Aasen et~al.(2015)Aasen, Burkart, Bolten, \protect\BIBand{}
  Bareth}]{Aasen2015}
Aasen H, Burkart A, Bolten A, Bareth G, 2015 \emph{{Generating 3D hyperspectral
  information with lightweight UAV snapshot cameras for vegetation monitoring:
  From camera calibration to quality assurance}}. \emph{{ISPRS Journal of
  Photogrammetry and Remote Sensing}} 108:245--259.

\bibitem[{Allahyari, Salari, \protect\BIBand{} Vigo(2015)}]{Allahyari2015}
Allahyari S, Salari M, Vigo D, 2015 \emph{{A hybrid metaheuristic algorithm for
  the multi-depot covering tour vehicle routing problem}}. \emph{European
  Journal of Operational Research} 242(3):756--768.

\bibitem[{Archetti, Carrabs, \protect\BIBand{} Cerulli(2018)}]{Archetti2018}
Archetti C, Carrabs F, Cerulli R, 2018 \emph{The set orienteering problem}.
  \emph{European Journal of Operational Research} 267(1):264--272.

\bibitem[{Arthur \protect\BIBand{} Vassilvitskii(2007)}]{Arthur2006}
Arthur D, Vassilvitskii S, 2007 \emph{k-means++: The advantages of careful
  seeding}. \emph{{Proceedings of the Eighteenth Annual ACM-SIAM Symposium on
  Discrete Algorithms}} 1027–--1035.

\bibitem[{{BBK}(2016)}]{BBK2016}
{BBK}, 2016 \emph{{Recommendations on sampling for hazard control in civil
  protection}}.

\bibitem[{{BBK}(2018)}]{BBK2018}
{BBK}, 2018 \emph{{Chemische Gefahren}}.
  \url{https://www.bbk.bund.de/DE/AufgabenundAusstattung/CBRNSchutz/Chemie/ChemGef/chemgef_node.html},
  online; accessed 14.05.2019.

\bibitem[{Binney, Krause, \protect\BIBand{} Sukhatme(2010)}]{Binney2010}
Binney J, Krause A, Sukhatme GS, 2010 \emph{Informative path planning for an
  autonomous underwater vehicle}. \emph{{IEEE International Conference on
  Robotics and Automation}}.

\bibitem[{Binney \protect\BIBand{} Sukhatme(2012)}]{Binney2012}
Binney J, Sukhatme GS, 2012 \emph{{Branch and bound for informative path
  planning}}. \emph{{IEEE International Conference on Robotics and Automation}}
  2147--2154.

\bibitem[{Boccardo et~al.(2015)Boccardo, Chiabrando, Dutto, Tonolo,
  \protect\BIBand{} Lingua}]{Boccardo2015}
Boccardo P, Chiabrando F, Dutto F, Tonolo FG, Lingua A, 2015 \emph{{UAV
  deployment exercise for mapping purposes: Evaluation of emergency response
  applications}}. \emph{Sensors} 15(7):15717--15737.

\bibitem[{Caselton \protect\BIBand{} Zidek(1984)}]{Caselton1984}
Caselton WF, Zidek JV, 1984 \emph{Optimal monitoring network designs}.
  \emph{Statistics \& Probability Letters} 2(4):223--227.

\bibitem[{Chao, Golden, \protect\BIBand{} Wasil(1996)}]{Chao1996top}
Chao IM, Golden B, Wasil E, 1996 \emph{The team orienteering problem}.
  \emph{{European Journal of Operational Research}} 88:464--474.

\bibitem[{{Copernicus EMS}(2018)}]{Copernicus}
{Copernicus EMS}, 2018 \emph{Copernicus emergency management service}.
  \url{http://emergency.copernicus.eu/}, online; accessed 15.05.2019.

\bibitem[{Cressie \protect\BIBand{} Wikle(2011)}]{Cressie2011}
Cressie NA, Wikle C, 2011 \emph{Statistics for Spatio-Temporal Data} (John
  Wiley \& Sons).

\bibitem[{Curran \protect\BIBand{} Atkinson(1998)}]{Curran1998}
Curran PJ, Atkinson PM, 1998 \emph{Geostatistics and remote sensing}.
  \emph{{Progress in Physical Geography}} 22(1):61--78.

\bibitem[{Current \protect\BIBand{} Schilling(1989)}]{Current1989}
Current JR, Schilling DA, 1989 \emph{The covering salesman problem}.
  \emph{{Transportation Science}} 23(3):208--213.

\bibitem[{Dang, Guibadj, \protect\BIBand{} Moukrim(2013)}]{Dang2013pso}
Dang DC, Guibadj RN, Moukrim A, 2013 \emph{{An effective PSO-inspired algorithm
  for the team orienteering problem}}. \emph{{European Journal of Operational
  Research}} 229(2):332--344.

\bibitem[{Das \protect\BIBand{} Kempe(2008)}]{Das2008}
Das A, Kempe D, 2008 \emph{Algorithms for subset selection in linear
  regression}. \emph{{Proceedings of the fortieth annual ACM symposium on
  Theory of computing}}, 45--54.

\bibitem[{Flanigan(1996)}]{Flanigan1996}
Flanigan DF, 1996 \emph{Short history of remote sensing of chemical agents}.
  \emph{Electro-Optical Technology for Remote Chemical Detection and
  Identification}, volume 2763, 2--18.

\bibitem[{{FwDV 100}(1999)}]{Feuerwehr1999}
{FwDV 100}, 1999 \emph{{Feuerwehr-Dienstvorschrift 100. F\"uhrung und Leitung
  im Einsatz. F\"uhrungssystem}}.
  \url{https://www.bbk.bund.de/SharedDocs/Downloads/BBK/DE/FIS/DownloadsRechtundVorschriften/Volltext_Fw_Dv/FwDV%20100.pdf?__blob=publicationFile},
  online; accessed 14.05.2019.

\bibitem[{Gendreau, Laporte, \protect\BIBand{} Semet(1997)}]{Gendreau1997}
Gendreau M, Laporte G, Semet F, 1997 \emph{The covering tour problem}.
  \emph{{Operations Research}} 45(4):568 -- 576.

\bibitem[{Golden et~al.(2012)Golden, Naji-Azimi, Raghavan, Salari,
  \protect\BIBand{} Toth}]{Golden2012}
Golden B, Naji-Azimi Z, Raghavan S, Salari M, Toth P, 2012 \emph{The
  generalized covering salesman problem}. \emph{{INFORMS Journal on Computing}}
  24(4):534 -- 553.

\bibitem[{Gunawan, Lau, \protect\BIBand{} Vansteenwegen(2016)}]{Gunawan2016}
Gunawan A, Lau HC, Vansteenwegen P, 2016 \emph{Orienteering problem: A survey
  of recent variants, solution approaches and applications}. \emph{{European
  Journal of Operational Research}} 255(2):315--332.

\bibitem[{H{\`{a}} et~al.(2013)H{\`{a}}, Bostel, Langevin, \protect\BIBand{}
  Rousseau}]{Ha2013}
H{\`{a}} MH, Bostel N, Langevin A, Rousseau LM, 2013 \emph{{An exact algorithm
  and a metaheuristic for the multi-vehicle covering tour problem with a
  constraint on the number of vertices}}. \emph{European Journal of Operational
  Research} 226(2):211--220.

\bibitem[{Hachicha et~al.(2000)Hachicha, Hodgson, Laporte, \protect\BIBand{}
  Semet}]{Hachicha2000}
Hachicha M, Hodgson MJ, Laporte G, Semet F, 2000 \emph{Heuristics for the
  multi-vehicle covering tour problem}. \emph{{Computers \& Operations
  Research}} 27(1):29--42.

\bibitem[{Harig \protect\BIBand{} Rusch(2011)}]{Harig2011}
Harig R, Rusch P, 2011 \emph{{Infrarot-Gefahrstoffkamera}}. \emph{{Forschung im
  Bevölkerungsschutz}} 14.

\bibitem[{Hollinger \protect\BIBand{} Sukhatme(2014)}]{Hollinger2014}
Hollinger GA, Sukhatme GS, 2014 \emph{Sampling-based robotic information
  gathering algorithms}. \emph{{The International Journal of Robotics
  Research}} 33(9):1271--1287.

\bibitem[{Irnich, Toth, \protect\BIBand{} Vigo(2014)}]{Irnich2014}
Irnich S, Toth P, Vigo D, 2014 \emph{The family of vehicle routing problems}.
  \emph{Vehicle Routing: Problems, Methods, and Applications, Second Edition},
  1--33 (SIAM).

\bibitem[{Ke et~al.(2016)Ke, Zhai, Li, \protect\BIBand{} Chan}]{Ke2016}
Ke L, Zhai L, Li J, Chan FTS, 2016 \emph{{Pareto mimic algorithm: An approach
  to the team orienteering problem}}. \emph{Omega} 61:155--166.

\bibitem[{Kilby(2013)}]{Kilby2011a}
Kilby P, 2013 \emph{Constraint programming for the vehicle routing problem}.
  Retrieved from: http://cp2013.a4cp.org/slides/t3.pdf.

\bibitem[{Krause et~al.(2008)Krause, McMahan, Guestrin, \protect\BIBand{}
  Gupta}]{Krause2008b}
Krause A, McMahan HB, Guestrin C, Gupta A, 2008 \emph{Robust submodular
  observation selection}. \emph{{Journal of Machine Learning Research}}
  9(Dec):2761--2801.

\bibitem[{Krause, Singh, \protect\BIBand{} Guestrin(2008)}]{Krause2008}
Krause A, Singh A, Guestrin C, 2008 \emph{{Near-optimal sensor placements in
  Gaussian processes: Theory, efficient algorithms and empirical studies}}.
  \emph{{Journal of Machine Learning Research}} 9(Feb):235--284.

\bibitem[{Mayfield, Eastwood, \protect\BIBand{} Burggraf(2000)}]{Mayfield2000}
Mayfield HT, Eastwood D, Burggraf LW, 2000 \emph{Infrared spectral
  classification with artificial neural networks and classical pattern
  recognition}. \emph{Chemical and Biological Sensing}, volume 4036, 54--66.

\bibitem[{Naji-Azimi et~al.(2012)Naji-Azimi, Renaud, Ruiz, \protect\BIBand{}
  Salari}]{Naji-Azimi2012}
Naji-Azimi Z, Renaud J, Ruiz A, Salari M, 2012 \emph{{A covering tour approach
  to the location of satellite distribution centers to supply humanitarian
  aid}}. \emph{{European Journal of Operational Research}} 222(3):596--605.

\bibitem[{Ozbaygin, Yaman, \protect\BIBand{} Karasan(2016)}]{Ozbaygin2016}
Ozbaygin G, Yaman H, Karasan OE, 2016 \emph{{Time constrained maximal covering
  salesman problem with weighted demands and partial coverage}}.
  \emph{{Computers \& Operations Research}} 76.

\bibitem[{Pedregosa et~al.(2011)Pedregosa, Varoquaux, Gramfort, Michel,
  Thirion, Grisel, Blondel, Prettenhofer, Weiss, Dubourg, Vanderplas, Passos,
  Cournapeau, Brucher, Perrot, \protect\BIBand{} Duchesnay}]{Pedregosa2011}
Pedregosa F, Varoquaux G, Gramfort A, Michel V, Thirion B, Grisel O, Blondel M,
  Prettenhofer P, Weiss R, Dubourg V, Vanderplas J, Passos A, Cournapeau D,
  Brucher M, Perrot M, Duchesnay E, 2011 \emph{Scikit-learn: Machine learning
  in {P}ython}. \emph{Journal of Machine Learning Research} 12:2825--2830.

\bibitem[{P{\v{e}}ni{\v{c}}ka, Faigl, \protect\BIBand{}
  Saska(2019)}]{Pvenivcka2019}
P{\v{e}}ni{\v{c}}ka R, Faigl J, Saska M, 2019 \emph{Variable neighborhood
  search for the set orienteering problem and its application to other
  orienteering problem variants}. \emph{European Journal of Operational
  Research} 276(3):816--825.

\bibitem[{Pisinger \protect\BIBand{} Ropke(2007)}]{Pisinger2007}
Pisinger D, Ropke S, 2007 \emph{A general heuristic for vehicle routing
  problems}. \emph{Computers \& Operations Research} 34(8):2403--2435.

\bibitem[{Rasmussen \protect\BIBand{} Williams(2006)}]{Rasmussen2006}
Rasmussen C, Williams C, 2006 \emph{Gaussian processes for machine learning},
  volume~1 ({MIT press, Cambridge}).

\bibitem[{Righini \protect\BIBand{} Salani(2008)}]{Righini2008}
Righini G, Salani M, 2008 \emph{New dynamic programming algorithms for the
  resource constrained elementary shortest path problem}. \emph{{Networks: An
  International Journal}} 51(3):155--170.

\bibitem[{Singh et~al.(2007)Singh, Kaiser, Batalin, Krause, \protect\BIBand{}
  Guestrin}]{Singh2007}
Singh A, Kaiser W, Batalin M, Krause A, Guestrin C, 2007 \emph{{Efficient
  planning of informative paths for multiple robots}}. \emph{{IJCAI
  International Joint Conference on Artificial Intelligence}} 2204--2211.

\bibitem[{Singh et~al.(2009)Singh, Krause, Guestrin, \protect\BIBand{}
  Kaiser}]{Singh2009}
Singh A, Krause A, Guestrin C, Kaiser WJ, 2009 \emph{Efficient informative
  sensing using multiple robots}. \emph{{Journal of Artificial Intelligence
  Research}} 34:707--755.

\bibitem[{Souffriau et~al.(2010)Souffriau, Vansteenwegen, Berghe,
  \protect\BIBand{} Van~Oudheusden}]{Souffriau2010}
Souffriau W, Vansteenwegen P, Berghe GV, Van~Oudheusden D, 2010 \emph{A path
  relinking approach for the team orienteering problem}. \emph{{Computers \&
  Operations Research}} 37(11):1853--1859.

\bibitem[{Stachniss, Plagemann, \protect\BIBand{}
  Lilienthal(2009)}]{Stachniss2009}
Stachniss C, Plagemann C, Lilienthal AJ, 2009 \emph{{Learning gas distribution
  models using sparse Gaussian process mixtures}}. \emph{{Autonomous Robots}}
  26(2-3):187--202.

\bibitem[{{Statistisches Bundesamt (Destatis)}(2018)}]{destatis2018}
{Statistisches Bundesamt (Destatis)}, 2018 \emph{{Ergebnisse des Zensus 2011}}.
  \url{https://www.zensus2011.de/DE/Home/Aktuelles/DemografischeGrunddaten.html?nn=3066576},
  online; accessed 14.05.2019.

\bibitem[{Tsiligirides(1984)}]{Tsiligirides1984}
Tsiligirides T, 1984 \emph{Heuristic methods applied to orienteering}.
  \emph{{Journal of the Operational Research Society}} 797--809.

\bibitem[{Vansteenwegen, Souffriau, \protect\BIBand{}
  Van~Oudheusden(2011)}]{Vansteenwegen2011}
Vansteenwegen P, Souffriau W, Van~Oudheusden D, 2011 \emph{{The orienteering
  problem: A survey}}. \emph{{European Journal of Operational Research}}
  209(1):1--10.

\bibitem[{Vidal et~al.(2015)Vidal, Maculan, Ochi, \protect\BIBand{} {Vaz
  Penna}}]{Vidal2015}
Vidal T, Maculan N, Ochi LS, {Vaz Penna} PH, 2015 \emph{{Large neighborhoods
  with implicit customer selection for vehicle routing problems with profits}}.
  \emph{{Transportation Science}} 50(2):720--734.

\bibitem[{Yu, Schwager, \protect\BIBand{} Rus(2014)}]{Yu2014}
Yu J, Schwager M, Rus D, 2014 \emph{Correlated orienteering problem and its
  application to informative path planning for persistent monitoring tasks}.
  \emph{{IEEE International Conference on Intelligent Robots and Systems}}
  342--349.

\end{thebibliography}
